\documentclass[11pt]{article}
\usepackage[linesnumbered,ruled,vlined]{algorithm2e}
\usepackage{amsfonts}
\usepackage{amsmath}
\usepackage{amssymb}
\usepackage{authblk}
\usepackage{bm}
\usepackage{booktabs}
\usepackage[margin=1in]{geometry}
\usepackage{graphicx}
\usepackage[separate-uncertainty=true]{siunitx}
\usepackage[colorlinks=true]{hyperref}
\usepackage{appendix}
\usepackage[]{mdframed}
\usepackage{float}

\title{GPT vs Human for Scientific Reviews: A Dual Source Review on Applications of ChatGPT in Science}
\author[1]{Chenxi Wu}
\author[1]{Alan John Varghese}
\author[1]{Vivek Oommen}
\author[1,2,*]{George Em Karniadakis}
\affil[1]{School of Engineering, Brown University, Providence, RI 02912, USA}
\affil[2]{Division of Applied Mathematics, Brown University, Providence, RI 02912, USA}
\affil[*]{Corresponding author. Email: george\_karniadakis@brown.edu}
\date{}

\begin{document}

\maketitle

%

\begin{abstract}
The new polymath Large Language Models (LLMs) can speed-up greatly scientific reviews, possibly using more unbiased quantitative metrics, 
facilitating cross-disciplinary connections, and identifying emerging trends and research gaps by analyzing large volumes of data.
However, at the present time, they lack the required deep  understanding of complex methodologies, they have difficulty in evaluating innovative claims, and they are unable to assess ethical issues and conflicts of interest. 
Herein, we consider 13 GPT-related papers across different scientific domains, reviewed by a human reviewer and SciSpace, a large language model,
with the reviews evaluated by three distinct types of evaluators, namely GPT-3.5, a crowd panel, and GPT-4. We found that 50\% of SciSpace’s responses to objective questions align with those of a human reviewer, with GPT-4 (informed evaluator) often rating the human reviewer higher in accuracy, and SciSpace higher in structure, clarity, and completeness. In subjective questions, the uninformed evaluators (GPT-3.5 and crowd panel) showed varying preferences between SciSpace and human responses, with the crowd panel showing a preference for the human responses. However, GPT-4 rated them equally in accuracy and structure but favored SciSpace for completeness.
\end{abstract}


\section{Introduction}

\indent 

The journal ``Philosophical Transactions of the Royal Society" is often credited as the first journal to establish a formal peer review process for scientific papers. While the journal dates back to 1665, a formal peer review process was first introduced in the mid-18th century, when the Royal Society began experimenting with a new form of peer review. Specifically, this early process involved distributing papers submitted to the Society to appropriate members (the reviewers) for their comments and recommendations while the final decision on publication was made by the Society's council. The peer review process evolved over time and became more systematic, rigorous and anonymous in the 19th and 20th centuries, aligning more closely with today's peer review process. 

Hence, for over a century, rigorous scientific literature reviews have been exclusively conducted by domain-specific human experts, disciplinary scientists who have spent decades on deepening their knowledge on a scientific topic by conducting experiments, deriving theory or performing computational simulations. However, as artificial intelligence (AI) advances, being at the dawn of artificial general intelligence (AGI) at the present time, we witness the emergence of advanced large language models (LLMs), bringing us to a crossroads in scientific literature review. In the following years, we will be confronted with the decision of whether LLMs will replace the traditional human experts in scientific literature reviews. As we are heading towards an era where LLMs will play a dominant role in the literature review, several key questions emerge at this crossroads. Firstly, what implications does this change hold for the quality and integrity of literature reviews in terms of rigor, fairness, depth, and context? Secondly, is it possible for LLMs to perform at the same level of expertise as domain experts who have dedicated decades to their fields? And finally, if LLMs can take over human roles in literature review, is there still a need for human involvement and oversight in this process? We have recently conducted a study that aims to partially address these questions by providing a dual-source literature review, where both humans and LLMs are simultaneously involved. Our goal is to initiate an in-depth discussion about the evolving role of LLMs in scientific literature reviews, examining how they can complement or potentially transform traditional methods, and offering insights into the future landscape of scholarly research. The papers we reviewed for this comparative analysis focus specifically on applications of ChatGPT within different scientific domains. These fields are characterized by a high volume of research and their dynamic nature, with numerous studies being published weekly. Besides exploring the role of LLMs in scientific literature reviews, this study also provides an overview of current trends and methodologies in the application of ChatGPT in science.

Researchers from various domains have conducted a multitude of studies to examine how large language models (LLMs) can be utilized to aid different tasks \cite{LLM_solveproblem1, LLM_solveproblem2, LLM_solveproblem3, LLM_solveproblem4, LLM_solveproblem5, LLM_solveproblem6, LLM_solveproblem7}, ranging from simple text generation to complex problem-solving. LLMs have also been studied in diverse areas such as speech recognition and synthesis \cite{LLM_speech1,LLM_speech2}, multimedia analysis \cite{LLM_media1,LLM_media2}, coding \cite{LLM_coding}, and even in taking exams \cite{LLM_exam1,LLM_exam2,LLM_exam3}. The rapid development in the field of AI has led to the creation of multiple advanced LLMs \cite{gpt-4inscience}. These include GPT-2, GPT-3, GPT-3.5, and GPT-4 from OpenAI \cite{gpt4_technical_report}, PaLM 2 from Google \cite{palm2}, Claude from Anthropic \cite{Anthropic}, and LLaMA 2 from Meta \cite{llama}. Among these LLMs, the GPT series from OpenAI has garnered significant attention due to its impressive capabilities. Over the past year, numerous studies have examined ChatGPT's capabilities and potential applications, particularly in public health \cite{potential_publichealth1,potential_publichealth2,potential_publichealth3,potential_publichealth4}, medicine \cite{potential_medicine1,potential_medicine2,potential_medicine2,potential_medicine3,potential_medicine4,potential_medicine5,potential_medicine6, potential_medicine7, potential_medicine8}, education \cite{potential_education1,potential_education2,potential_education3,potential_education4,potential_education5,potential_education6,potential_education7}, environment \cite{potential_environment1}, and mathematics  \cite{potential_math1,LLM_exam3}. 

Recently, the focus of academic inquiry has started to shift from just utilizing the model’s functionalities to investigating ways in which ChatGPT can be augmented and tailored to solve domain-specific problems in scientific disciplines, where there is a demand not only for speed but also for precision, robustness, accuracy, and reliability. An increasing number of papers goes beyond merely outlining what ChatGPT can do \cite{adautogpt,automlgpt,chatdrug,chemcrow,cancergpt,cohortgpt,genegpt,geogpt,geotechgpt,mycrunchgpt,pharmacygpt,roboticsgpt,surfacegpt,synergpt,chatmof}. They focus on exploring how the ChatGPT model can be extended or customized to meet specific needs, especially in scientific disciplines. Exploring how LLMs, like ChatGPT, can improve scientific research is crucial, with the potential to greatly enhance efficiency, productivity, and innovation. LLMs can process vast datasets quickly, offering insights and aiding hypothesis generation \cite{hypotheses,LLM_literature}, which is especially valuable in interdisciplinary fields \cite{LLM_interdisciplinary}. Additionally, domain-specific LLMs can lower barriers to entry, make complex information more accessible to non-experts, and encourage diverse participation from various backgrounds \cite{chemcrow,geogpt,geotechgpt,roboticsgpt}. This not only enriches the scientific discourse but also drives inclusive innovation, paving the way for a more collaborative and versatile research environment.

For example, in the field of medicine, the expansion of ChatGPT’s capabilities has proven to be particularly impactful ~\cite{LLM_solveproblem1}. Applications range from predicting drug interactions and synergies \cite{cancergpt,synergpt}, developing comprehensive medication plans \cite{cohortgpt}, categorizing and interpreting intricate medical reports, suggesting novel drugs \cite{chatdrug,chemcrow}, and summarizing health narratives to gain insights into conditions like Alzheimer’s disease \cite{adautogpt}. By integrating domain-specific databases and sophisticated algorithms, researchers have tailored ChatGPT to provide more accurate and relevant responses to queries related to gene sequences, protein structures, and associated biological functions \cite{genegpt}.

Similarly, in the realms of machine learning \cite{automlgpt}, engineering \cite{surfacegpt,roboticsgpt,chatmof}, and geography \cite{geogpt,geotechgpt}, scientists have been working on modifying ChatGPT to assist scientists and engineers in working more efficiently and productively. Specifically, in the domain of machine learning, ChatGPT is designed to conduct data preprocessing, fine-tuning hyperparameters, and aiding in model selections \cite{automlgpt}. Within the field of engineering, ChatGPT has been customized for specific tasks such as answering questions about surface engineering \cite{surfacegpt} and generating code for robotics design \cite{roboticsgpt}. Additionally, it has been utilized to perform comprehensive tasks in the field of geography \cite{geogpt,geotechgpt}.

In the present study, we present a partial overview of innovative frameworks and extensions applied to ChatGPT in scientific domains, presented in a question-answer format for clear and accessible communication. The responses are compiled from both SciSpace, a large language model (LLM), and one human reviewer, aiming to present a partial overview of ChatGPT’s developments in science. Additionally, this dual-source approach enables us to initiate a critical discussion on the performance of LLMs versus human reviewers in literature reviews, contributing valuable insights to the ongoing discourse on AI integration in academic research.

%

Scispace initially emerged as a typesetting tool designed to automate formatting for journal submissions. 
Over time, it evolved to proficiently leverage LLMs, aiming to enhance the accessibility of scientific information by reducing the comprehension barrier \cite{scispace_webpage}. While SciSpace has made significant advances in understanding and answering questions on scientific papers, it might lack the depth of understanding, critical thinking, logical analysis, and constructive expertise that human reviewers can provide. The combination of subjective judgment, expert knowledge, and logical reasoning might enable human reviewers to outperform SciSpace.  Despite this, SciSpace holds considerable value as a supportive tool, offering preliminary assessments that can streamline the review process and bolster efficiency. 

This work has multi-fold motivations and goals:

\begin{enumerate}
  \item To provide a comparative review from two perspectives (Scispace and the human reviewer) on the applications of ChatGPT in scientific fields.
  \item To present a representative sample of prevalent frameworks developed and implemented for various versions of ChatGPT across diverse scientific disciplines, such as medicine, machine learning, engineering, and geography.
  \item To demonstrate the effectiveness of SciSpace and the human reviewer in analyzing recent literature related to the applications of ChatGPT in scientific domains.
  \item To provide insights and statistical analyses that reflect the performance of SciSpace and the human reviewer, as evaluated from three different perspectives (GPT-3.5, GPT-4, and a crowd panel).
  \item To explore and articulate the potential limitations of LLMs-based models in performing and evaluating literature review.
\end{enumerate}

\section{Methodology}

The use of ChatGPT in science and engineering is a prominent and ongoing subject of intense research, leading to frequent publication of numerous scholarly papers on a daily basis. Capturing the current and innovative concepts becomes a critical component of research in the study of LLMs. Scispace, specifically developed for conducting literature reviews, can analyze papers at a significantly faster rate than humans. We aim to assess the performance of Scispace compared to humans in examining recent papers focused on the applications of ChatGPT in science. For our analysis, we have selected 13 papers and formulated five to six targeted questions for each. All questions and answers are included in the Appendix \ref{Appendix} for reference. The primary aim of these questions is to summarize the fundamental objectives, methodologies, and findings of each paper. It is important to note that these questions are not intended to appraise the novelty or quality of the papers. Instead, they are designed to provide the audience with a basic understanding of each paper's content without a comprehensive read-through.

\begin{figure}[h!]
    \centering
    \includegraphics[width=0.8\linewidth]{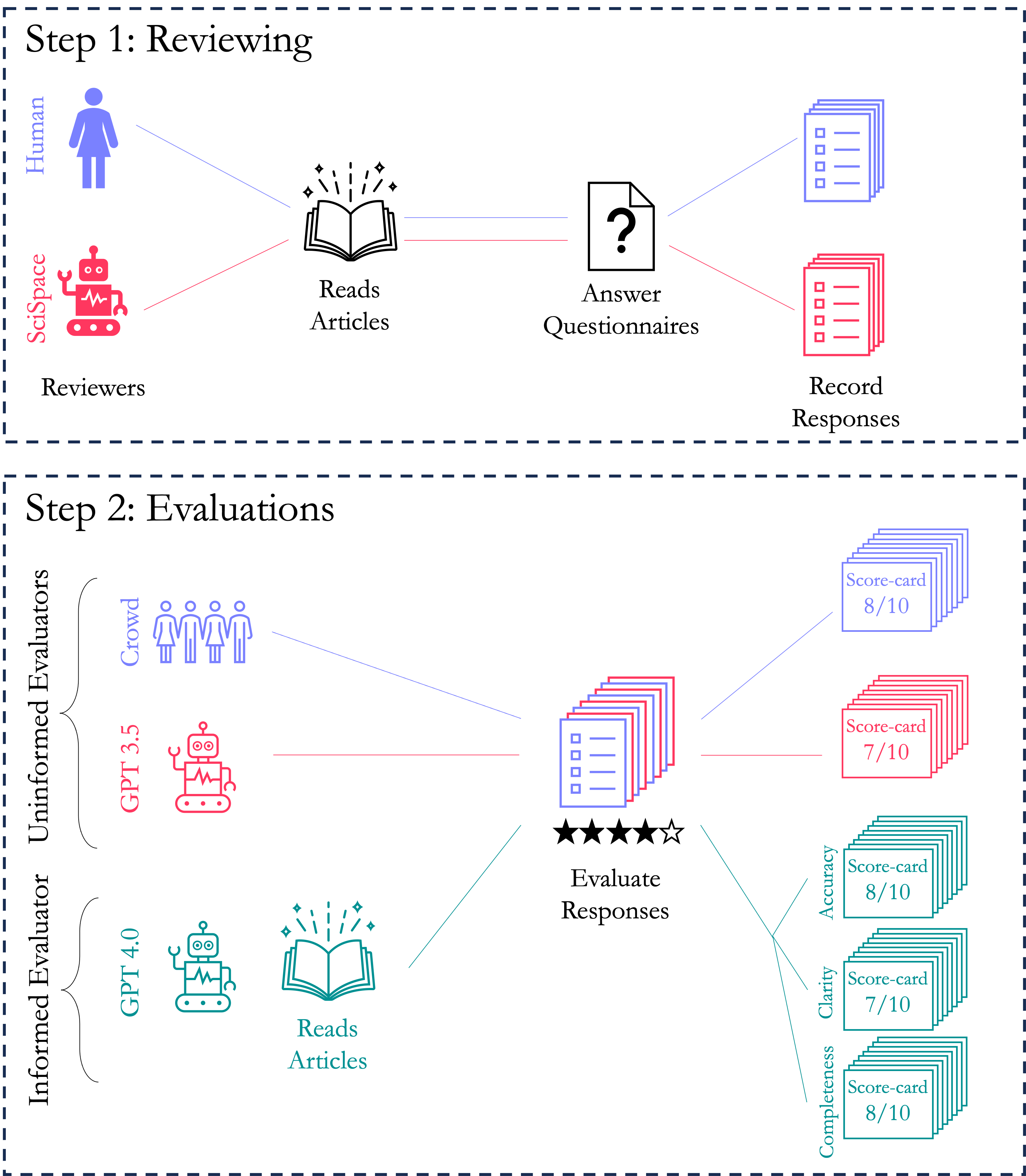}
    \caption{The methodology adopted in this study consists of two steps. First, a human reviewer and Scispace separately read the articles and answered a predefined questionnaire. In the second step, we evaluate the recorded responses. Specifically, we have three evaluators, 1) GPT 3.5, 2) crowd with 25 individuals, and 3) GPT 4.0. The crowd and GPT 3.5 directly evaluate the responses without reading the articles, while GPT 4.0 reads the articles before grading the responses.  }
    \label{fig:methodology}
\end{figure}

The designed questions are responded to by both Scispace and one human reviewer. The responses are subject to assessment by three distinct types of evaluators, namely GPT-3.5, a crowd panel, and GPT-4. These evaluators are categorized into two groups: uninformed evaluators (GPT-3.5 and the crowd panel) and informed evaluator (GPT-4). The uninformed evaluators provide their assessments without having access to the paper, whereas the informed evaluator conducts evaluations having reviewed the paper. In Figure \ref{fig:methodology}, a graphical representation of our methodology is depicted. Detailed information regarding the evaluators and their respective methodologies can be found in Section \ref{evaluator}.

\subsection{SciSpace}


%
SciSpace is a GPT-3 based AI-tool that aims to accelerate scientific discovery by assisting researchers to analyze and understand scientific literature faster. SciSpace offers the ability to read, summarize and provide in-depth explanations of scientific papers. Moreover, it enables users to interact with research papers through a chatbot. Users can upload research papers to SciSpace or access a repository with over 200 million papers from diverse disciplines \cite{scispace_webpage}.
SciSpace is equipped with the following utilities. 
\begin{itemize}
    \item \textbf{Copilot}: SciSpace has a utility called `Copilot - Read with AI' that reads and comprehends research articles and answers related queries in an interactive manner. Copilot supports follow-up questions and replies to both general and specific questions from within the article and highlights the corresponding location in the manuscript. SciSpace also provides a chrome extension of copilot and extends the utilities to anywhere on the web for a better user experience.

    \item \textbf{AI assisted literature survey}: SciSpace offers a literature review tool that can summarize and help gain insights into a given research topic from a collection of research papers uploaded by the user or from a repository of more than 200 million papers already available within SciSpace. Users also have the flexibility to customize the criteria that SciSpace uses to compare and contrast research articles.

    \item \textbf{Citation generator}: SciSpace has a citation generator that allows users to create citations in their preferred style from a selection of over 2500 citation styles.
    
    \item \textbf{Paraphraser}: SciSpace offers a paraphrasing tool that helps to change the structure of the sentence without changing its meaning. The paraphrasing tool can rephrase the provided sentences up to 500 words into a more academic tone. 

    \item \textbf{AI Detector}: SciSpace has an AI detector that quantifies the presence of AI in query sentences. 
\end{itemize}

In this study, we utilize the copilot tool provided by SciSpace for understanding the research article. 
We consider research articles that investigate the influence of ChatGPT in four scientific disciplines - Medicine and Health, Machine Learning, Engineering, and Geography, as mentioned earlier. 
We prepare a questionnaire for each article, and use these questions for prompting the SciSpace copilot, and record the response. 
Next, we compare and critically analyze the copilot's response with a human's response.

\subsection{Human reviewer}

To ensure an unbiased and fair comparison of the competencies between the human reviewer and SciSpace, the human reviewer conducts the review of scientific papers independently, without referring to the answers provided by SciSpace or any other external tools. We ensured that the conclusions and assessments of the human reviewer are derived exclusively from the reviewer's own expertise, knowledge, and critical analytical skills, devoid of any influence from SciSpace’s responses.

The primary objective of this study is to demonstrate the performance of Scispace in comparison to that of a human reviewer, specifically within the realm of literature focusing on the recent advancements in ChatGPT. This investigation is intentionally designed to be case-specific, concentrating on a niche area of research rather than covering a wide range of topics. This targeted approach allows for a more detailed and focused examination of Scispace's capabilities in handling specialized content, thereby providing valuable insights into its efficacy in facilitating scientific research around ChatGPT.

As a result of this specialized focus, the study does not incorporate a wide range of human reviewers from various academic or professional backgrounds. For the sake of consistency, the same human reviewer is utilized throughout the study. The human reviewer involved in this study possesses a moderate level of expertise in machine learning, deep learning, and large language models, ensuring that the reviewer has the requisite understanding necessary to critically evaluate and interpret the specialized literature in this field. Our intention is to maintain a consistent level of expertise and understanding of the content across all papers. This approach seeks to balance the depth of knowledge with a representative level of expertise that reflects a significant portion of the academic community engaged in this area of research.

\subsection{Evaluations} \label{evaluator}

To assess the responses provided by Scispace and the human reviewer, we utilize GPT-3.5, GPT-4, and a panel of crowd evaluators to systematically rate the answers from both Scispace and the human reviewer.

\subsubsection{Uninformed evaluators}

The uninformed evaluators, GPT-3.5 and the crowd panel, conduct their assessments solely based on the responses, without access to the papers. Thus, their evaluations rely exclusively on the information presented in the responses from Scispace and the human reviewer. This scenario mirrors the real-world situation with many readers, especially those who seek quick insights  or with limited time. They rely heavily on summaries, abstracts, or expert opinions to grasp the paper's content. In this context, they effectively act as uninformed evaluators. An ideal set of responses to the designed questions should equip the uninformed readers with a basic understanding of the paper's content, objectives, methodology, and results, thereby eliminating the need for them to read the full paper. The feedback from these uninformed evaluators provides critical insights into how effectively the responses from Scispace and the human reviewer can convey the essence of the research paper to someone who has not read it.

However, uninformed evaluators face inherent challenges in assessing the accuracy of answers due to their lack of access to the original source material. This limitation is particularly significant in the case of objective questions, which are based entirely on factual information from the paper. Without direct access to these details, uninformed evaluators are unable to verify the factual correctness of the responses. When the questions are not merely interpretive or subjective, but require specific information directly from the paper, ratings from uninformed evaluators become less effective. Therefore, in this study, we have oriented uninformed evaluators towards evaluating more subjective and interpretive questions.

Here is a prompt provided to both GPT-3.5 and the crowd panel for evaluation:

\begin{mdframed}
\textit{As an instructor, you assigned two students a literature review and posed several questions to assess their understanding. Your aim is to determine which student's responses are more informative. Without reading the paper yourself, assess which student provides a clearer and better explanation of the paper. Please assign a score from 1 to 10 to each student’s responses.}
\end{mdframed}

This prompt sets up a scenario where evaluators act as instructors and judge the quality of literature reviews based on the students' responses to specific questions. They are instructed not to read the papers themselves, focusing instead on assessing clarity and informativeness of each response, with a scoring system from 1 to 10 for comparison.
\begin{itemize}
    \item \textbf{GPT-3.5:} GPT-3.5, as developed by OpenAI, is designed with a wide range of capabilities, but it does not possess the ability to directly access or read external documents, such as academic papers. This limitation inherently positions GPT-3.5 as an uninformed evaluator. In this study, the prompt and the responses from both Scispace and the human reviewer are inputted into GPT-3.5 anonymously, labeled simply as `Student 1' and `Student 2'. To further mitigate any potential biases, the labels `Student 1' and `Student 2' are alternated between Scispace and the human reviewer in different instances. Such a methodological design ensures that GPT-3.5's assessments are based solely on the content and quality of the responses, independent of any preconceived notions about the source of the response.
    
    \item \textbf{Crowd panel:} The crowd panel for this study consists of 25 researchers affiliated with Brown University with diverse academic and professional backgrounds, including biomedical engineering, ocean engineering, chemical engineering, fluid mechanics, and mathematics. This diversity across disciplines allows the panel to provide a wide range of viewpoints in their evaluations. All members of the panel possess experience in machine learning and deep learning research, which is crucial for evaluating literature on ChatGPT and similar models. Their collective expertise enables them to critically analyze and provide insightful ratings on the effectiveness of the responses in conveying complex technical information. Physical copies of the questionnaires are provided to the crowd panel. In these questionnaires, the responses from Scispace and the human reviewer are anonymously labeled as `Student 1' and `Student 2'. To remove potential biases, the labels `Student 1' and `Student 2' are alternated in each questionnaire. Along with the questionnaires, the crowd panel is given the aforementioned prompt to guide their evaluation process. Each paper in the study has been assessed by four distinct individuals from the crowd panel. The average scores from these four separate assessments are calculated.
\end{itemize}

\subsubsection{Informed evaluators}
In this study, the role of the informed evaluator is fulfilled by GPT-4, which is equipped with a plugin feature that enables it to read and analyze academic papers. This functionality allows GPT-4 to access and process the papers' content, enhancing its ability to provide insightful evaluations of responses compared to GPT-3.5. GPT-4 can evaluate responses to questions and directly compare them against the actual content of the papers, thereby offering a more informed and accurate assessment. 

While innovative, the functionality of GPT-4, equipped with a plugin to read and analyze academic papers, does come with certain limitations. Its understanding relies on data patterns, differing from human-like comprehension. For instance, although GPT-4 can process and analyze text, its understanding is based on patterns in data rather than human-like comprehension. Moreover, GPT-4's interpretations are influenced by the data it has been trained on, which can introduce biases or a lack of understanding of context. The detailed discussion about its limitations can be found in section \ref{section:limitation}. Despite the limitations, GPT-4 can efficiently provide a high-level overview of the content, highlight key points in a paper, and verify the responses from both Scispace and the human reviewer for objective questions. Thus, GPT-4's contribution to analyzing responses in academic papers can still be both substantial and meaningful.

In this study, we instruct GPT-4 to evaluate the responses from Scispace and the human reviewer based on three criteria: accuracy, structure \& clarity, and completeness. Similarly, Scispace and the human reviewer are anonymously labeled as `Student 1' and `Student 2', with these labels being alternated for different cases. The following prompt is provided to GPT-4 for this purpose:

\begin{mdframed}
\textit{As a teacher, you are to evaluate the responses of two students to a paper. Please begin by reading the paper provided. Then, rate each student's answers according to the following criteria on a scale from 1 to 10:}

\begin{itemize}
    \item \textit{Accuracy: The correctness of the answers in relation to the paper.}
    \item \textit{Structure and Clarity: The logical organization, conciseness, and ease of understanding of the answers.}
    \item \textit{Completeness: The extent to which the answer fully addresses the question posed.}
\end{itemize}

\textit{Your evaluation should reflect the students' level of comprehension and their ability to effectively communicate their understanding.}

\end{mdframed}

This prompt directs GPT-4 to provide three separate scores based on the factual accuracy of the responses, their structure and clarity, as well as the completion with which they address the posed questions. The objective is to ensure that the evaluation captures both the depth of content and the effectiveness of communication. 

\section{Applications of ChatGPT in scientific fields}

During the preparation of this paper, Microsoft released a study titled `Impact of Large Language Models on Scientific Discovery: a Preliminary Study using GPT-4' \cite{gpt-4inscience}. Their research primarily explored the inherent abilities of Large Language Models (LLMs) in addressing scientific challenges. They explicitly stated that the integration of LLMs with other tools or models was beyond the scope of their study. 

In our research, we strive to provide a summary of various external tools and methods utilized for enhancing the performance of GPT models in scientific disciplines. Our study is not restricted to a specific version of GPT; it includes examples utilizing GPT-2, GPT-3, GPT-3.5, and GPT-4 across different fields such as medicine, machine learning, engineering, and geography. We summarize strategies that scientists employ to merge the capabilities of GPT models with their specific domains, aiming to enhance outcomes. The majority of the papers we reviewed adopted a combination of these methods.

\begin{itemize}
  \item \textbf{Iterative Questioning and Incorporating Evaluators' Feedback:} The effectiveness of GPT models can be enhanced through iterative questioning. When initial responses from the model are inadequate or incorrect, evaluators refine their prompts, highlight mistakes, or supply additional information. This process can guide the model towards providing more accurate and relevant answers. The ChatDrug \cite{chatdrug} integrated an evaluator module to validate responses and facilitate iterative questioning. In \cite{roboticsgpt}, non-technical users can stay in the loop to evaluate ChatGPT’s code output, either through direct inspection or by using a simulator. The non-technical users can provide feedback and help GPT models improve their coding for robotics in an iterative manner.
  
  \item \textbf{Chain-Of-Thought:} The success of Chain-of-Thought reasoning when combined with Large Language Models' (LLMs) use of external tools demonstrates a significant advancement in AI capabilities \cite{chain-of-thought1,chain-of-thought2}. The Chain-of-Thought model adheres to a structured format known as Thought, Action, Action Input, and Observation. The GPT model acts as organizer of information; it reasons and assesses the current status of a task, evaluates its relevance to the final objective, and plans subsequent steps to solve the task. In this study, we include six papers that applied the iterative Chain-of-Thought to enhance the performance in chemistry \cite{chemcrow}, biology \cite{genegpt}, health \cite{adautogpt,cohortgpt}, and geography \cite{geogpt,geotechgpt}. This iterative Chain-of-Thought process, as illustrated in studies such as \cite{chemcrow,geogpt,geotechgpt,adautogpt,genegpt}, is typically combined with domain-specific tools to boost performance. During the `Thought' phase, the GPT model reasons and then requests a domain-specific tool, indicated by the keyword “Action,” along with the necessary input for this tool, marked by “Action Input.” Following this, during the text generation pauses, an external tool executes the requested function using the given input. The outcome is returned back to the GPT models with the keyword “Observation,” prompting the models to return to the `Thought' phase. This cycle repeats iteratively until the final goal is achieved.
  
  \item \textbf{Integration with Domain-Specific Tools:} It is Often advantageous to blend GPT models's capabilities with tools and models specifically designed for scientific discovery. This synergy allows researchers to capitalize on the strengths of both GPT models and specialized tools for more reliable and precise outcomes. For scientists, considering the integration of existing domain-specific tools with GPT models can be highly beneficial. In our study, we summarize examples of how this can be effectively achieved \cite{chemcrow,geogpt,geotechgpt,adautogpt}.
  
  \item \textbf{External Database:} GPT models have a vast knowledge base, but they might not always be up-to-date or comprehensive in a specific domain. External databases can provide up-to-date and specific data, which significantly improve the accuracy of the model's outputs, especially in rapidly evolving fields like medicine or technology. This can be done by connecting the GPT model with external databases through APIs (Application Programming Interfaces) \cite{adautogpt}. Another method is to develop a hybrid system where the GPT model can query the external database as part of its processing \cite{synergpt, cancergpt,cohortgpt,pharmacygpt,surfacegpt,automlgpt}.
  
  \item \textbf{Existing Platforms:} Several platforms are currently available, which facilitate the development of custom GPT models for a diverse range of users, including AutoGPT, HuggingGPT \cite{LLM_media4}, LlamaIndex and AutoGen \cite{autogen}. AD-AutoGPT is a customized AutoGPT that aims to analyze complex health narratives of Alzheimer's Disease \cite{adautogpt}. GPT for Surface Engineering \cite{surfacegpt} utilizes LlamaIndex to index data in the field of surface engineering and also assess whether or not the sourced information given by the sourced model can adequately answer the query.

  \item \textbf{Few-shot In-context Learning:} Few-shot in-context learning involves providing a small number of examples within the prompt to GPT models. These examples can add context to the task at hand, helping to enhance the model's performance by guiding it toward the desired type of response or solution. This method leverages the model's pre-trained knowledge and ability to infer patterns from limited data. In this study, a few paper \cite{synergpt, cancergpt,cohortgpt} demonstrate the effectiveness of few-shot in-context learning in GPT models.
\end{itemize}

\section{Discussion}
\subsection{Statistics}

Ensuring the accuracy of the responses generated by Scispace is crucial, as it significantly impacts the tool's reliability and effectiveness, especially in academic and research contexts. To facilitate the evaluation of Scispace's responses, we have divided the questions into two categories: objective and subjective. In the Appendix \ref{Appendix}, we have set the background for objective questions in blue and for subjective questions in grey.

\begin{figure}[h!]
    \centering
    \includegraphics[scale=0.75]{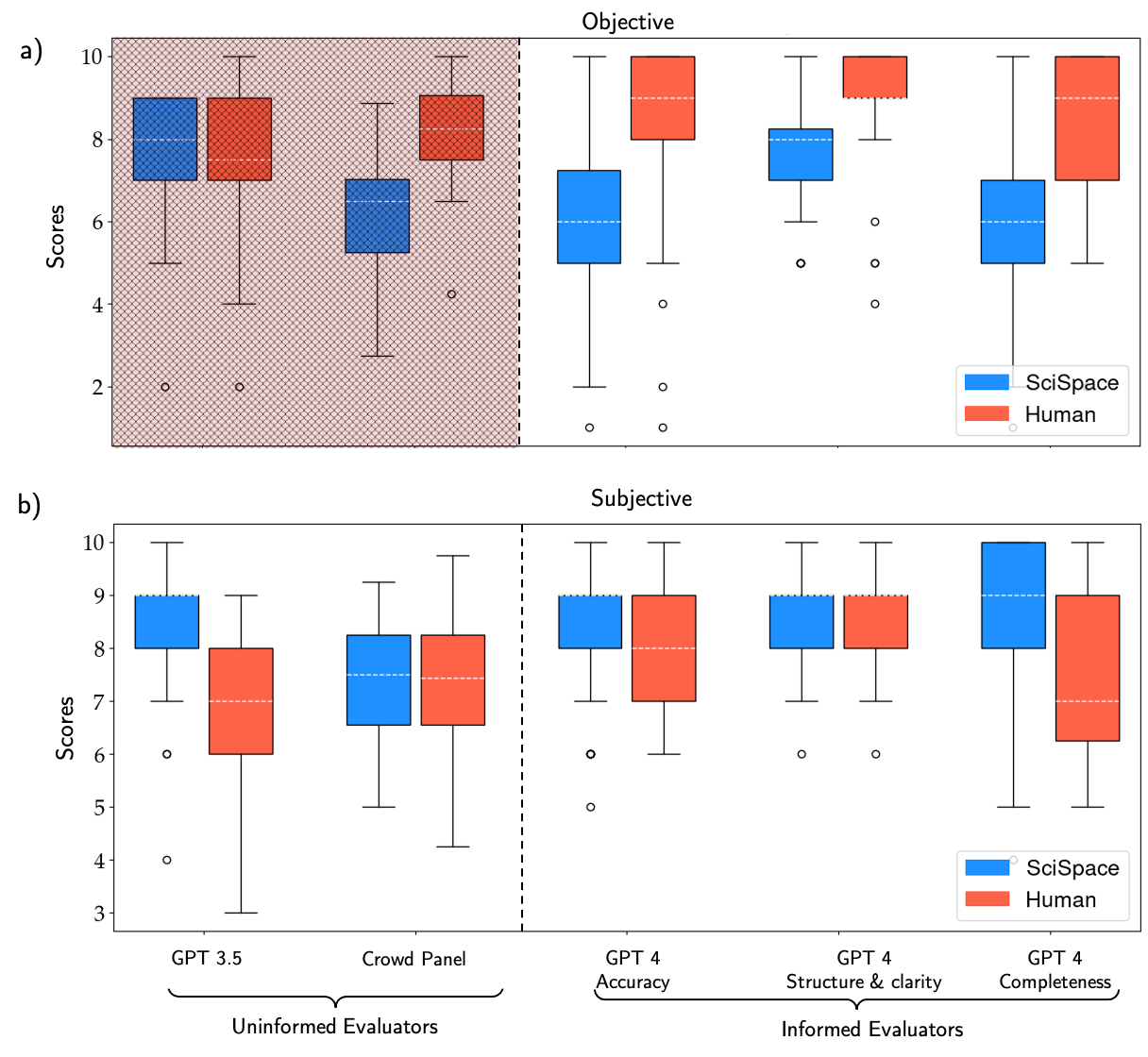}
    \caption{Comparitive evaluation of responses by SciSpace and human for a) objective and b) subjective questions. The box plots indicating the scores assigned by the different evaluators for the responses by SciSpace is shown in blue, and that for the responses by human is shown in red. We see that on objective questions an informed evaluator prefers the human response over the response by SciSpace. In the case of subjective questions, the different evaluators prefer the responses by SciSpace slightly more than, if not equally as, the human response.}
    \label{fig:comparison_fig}
\end{figure}

Objective questions are straightforward with definitive, factual answers that can be can be directly retrieved from the paper. In this study, 24 questions are classified as objective questions. We find that 50\% of SciSpace's responses to objective questions agree with the human reviewer's answers. Shown in Figure \ref{fig:comparison_fig} a), the informed evaluator GPT-4 also gives a higher score in accuracy for the human reviewer's answers. GPT-4 also gives a higher score for SciSpace's response for structure \& clarity and completeness.

While all the questions in our study are crafted to summarize the papers and are grounded in factual information, subjective questions are those that allow for a greater degree of personal interpretation. These interpretations play a crucial role in aiding uninformed readers to better understand the papers. To effectively compare the performance of Scispace against that of the human reviewer, we plot the ratings on these subjective questions in Figure \ref{fig:comparison_fig} b). In the case of uninformed evaluators, GPT 3.5 assigns higher scores to responses by SciSpace, whereas the crowd panel equally favors the responses by SciSpace and humans. The informed evaluator (GPT-4) equally prefers the responses by SciSpace and human in terms of accuracy and structure \& clarity. However, there is a clear preference for SciSpace responses in terms of completeness because GPT-4 often interprets a lengthier answer to be more complete.

\subsection{Advantages of SciSpace} 

Utilizing SciSpace as a tool for reviewing academic papers proves to be a significant time-saving strategy, as it can swiftly and efficiently respond to a wide array of questions. SciSpace excels at answering broad and overarching questions, such as delineating the paper’s main objectives and providing general summaries. As documented in the Appendix, SciSpace’s answers regarding the paper’s objectives are highly accurate, though they sometimes lack conciseness.

As our study demonstrates, SciSpace’s major advantage lies in its speed. Its ability to rapidly sift through extensive volumes of text and pinpoint relevant information enables reviewers to expedite their work, affording them more time and attention for the deeper, more nuanced aspects of the paper that may necessitate human expertise and critical analysis. Thus, SciSpace enhances the efficiency of the review process, ensuring a swift establishment of a foundational understanding of the paper’s content. Indeed, SciSpace can serve as a valuable supplementary tool in the paper review process, significantly reducing the time and effort required for completion.

\subsection{Clarity and Length} 

Out of a total of 66 questions, the human reviewer provided longer responses in 7 instances. This implies that in approximately 90\% of the cases, or 59 out of the 66 questions, Scispace provided longer responses than the human reviewer. While LLMs-based evaluators generally perceive SciSpace's responses as more informative, the crowd panel critiques these responses for being redundant and lacking a coherent structural organization. SciSpace's responses are typically presented in bullet points. In an effective list, bullet points are usually either parallel or progressive in nature. Parallel structure means each point focuses on explaining a different aspect of a matter, creating a consistent and easy-to-follow flow. Progressive structure, on the other hand, implies that each point builds upon the previous one, leading the reader through a logical progression of ideas. However, in SciSpace's responses, the bullet points often lack interconnection. They tend not to follow a parallel or progressive format, impacting the overall coherence of the response.

\begin{figure}[htbp]
    \centering
    \includegraphics[scale = 0.5]{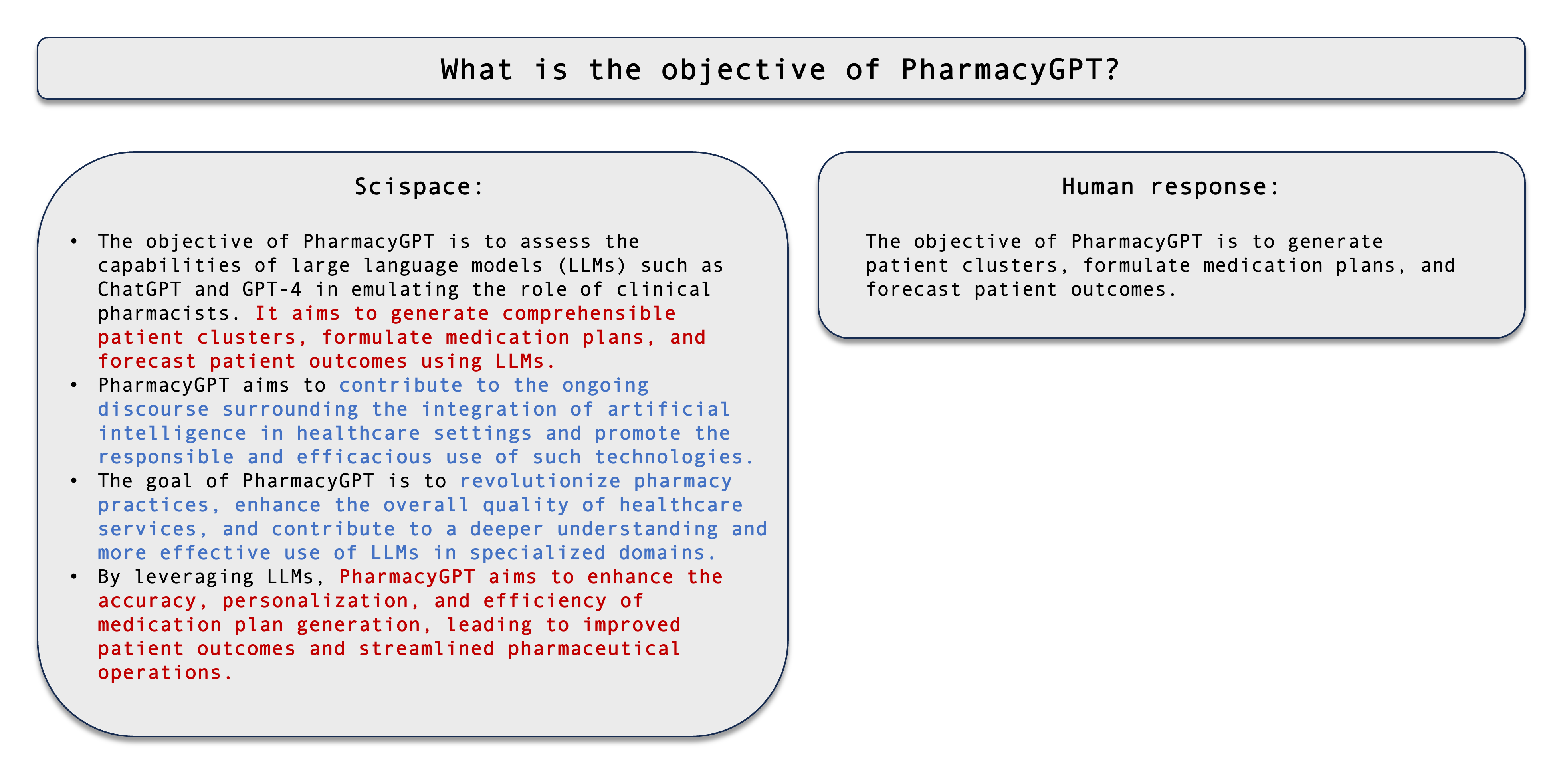}
    \caption{Comparison of Clarity and Length in Responses from SciSpace and Humans.}
    \label{fig:example1}
\end{figure}

Figure \ref{fig:example1} is an example, when we ask SciSpace what the objective of PharmacyGPT is, the bullet points given by SciSpace are not effective for several reasons:

\begin{itemize}
    \item \textbf{Repetitiveness and Overlapping Themes:} There is considerable overlap in the content of each bullet point. For example, all points repeatedly emphasize the use of LLMs in pharmacy practices, but they do not distinctly advance the topic or provide new information in each point. As highlighted in red and blue, the themes in each bullet point overlap significantly. The first and fourth points both discuss improving patient outcomes with LLMs, while the second and third points reiterate the goal of revolutionizing pharmacy practices.
    \item \textbf{Verbosity:} The bullet points are wordy and include more information than necessary to convey the core idea. Phrases such as ``to contribute to the ongoing discourse surrounding the integration of artificial intelligence in healthcare settings" could be more concise. The verbosity in SciSpace's responses makes the answers harder to read than the paper's abstract, counteracting SciSpace's goal of simplifying information comprehension.
    \item \textbf{Lack of Clear Logical Flow: } Each bullet point seems to restate the project's goals with slight variations, instead of introducing new aspects or details in a structured manner. This results in a lack of logical flow from one point to the next, hindering the reader's ability to comprehend distinct aspects or the full scope of the paper.
\end{itemize}

On the other hand, the human response is notably more succinct and straightforward. This brevity and directness can make it easier for readers to quickly grasp the primary intentions of paper. Despite being less detailed, the human response effectively conveys the essential information and usually demonstrates a clear and logical presentation of the main ideas.

\subsection{Structural Comprehension} 

Academic papers frequently employ the use of subtitles to clearly define and separate different sections. This structural element is crucial for readers to follow the argument, comprehend the methodology, and grasp the results presented. When a paper offers an analysis of a specific model from three distinct perspectives, a human reader can efficiently utilize the subtitles to navigate the sections and develop a comprehensive understanding of the content within each part. However, SciSpace, in its current form, faces challenges in accurately capturing and reflecting this cohesive structure in its responses.

For instance, in the experimental sections of the paper on ChatDrug \cite{chatdrug}, the authors verify the effectiveness of ChatDrug for drug editing on three types of drugs: small molecules, peptides, and proteins. The discussion of each drug type is allocated to separate sections — section 4.1 for small molecules, section 4.2 for peptides, and section 4.3 for proteins — allowing a reader to follow and comprehend each category in a structured and segmented manner.

\begin{figure}[htbp]
    \centering
    \includegraphics[scale = 0.5]{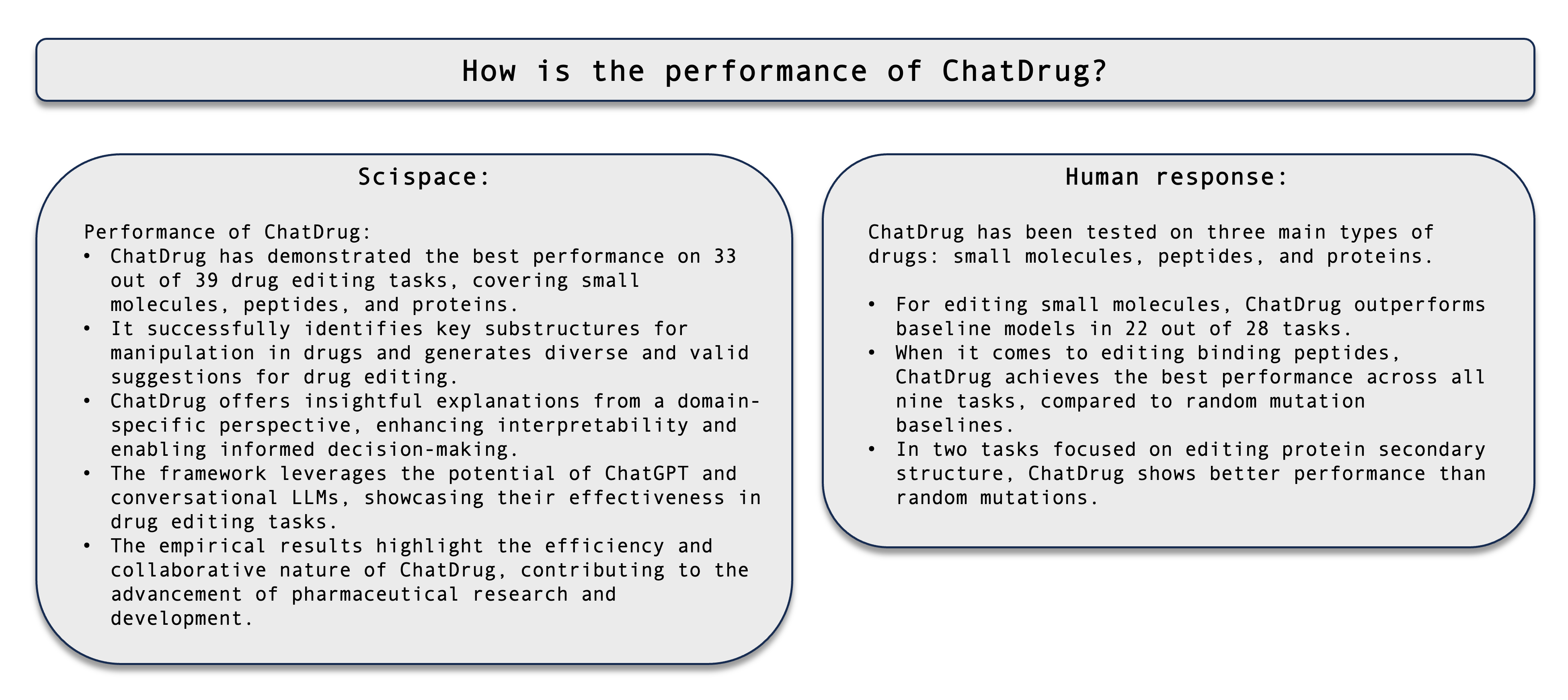}
    \caption{Comparison of Structural Comprehension in Responses from SciSpace and Humans.}
    \label{fig:example2}
\end{figure}

As shown in Figure \ref{fig:example2}, the response from the human reviewer successfully captures the structure of the paper as organized by its subtitles. SciSpace, on the other hand, tends to struggle with this task. SciSpace's response provides a general summary of ChatDrug's performance. However, it does this without following the structural divisions indicated by the paper's subtitles. The result is a comprehensive but vague overview. It merges the distinct analyses for different drugs into a single narrative. The answer from Scispace overlooks the subtitles of the paper uses, leading to a summary that lacks the detailed segmentation presented in the original paper.

SciSpace's responses might not clearly distinguish the unique contributions of each paper section, leading to a lack of depth and clarity. Improving its ability to use the structure indicated by subtitles would help produce more comprehensive and coherent summaries, offering users a better understanding of the paper's content.

\subsection{Interpreting Graphical Information} 

Scispace’s performance is hindered by its lack of capability to interpret graphical data, leading to significant oversights in its responses. Graphs, charts, and other visual representations in scientific papers are critical, as they often convey key examples and insights essential for understanding the paper. Neglecting this information means that users miss out on these essential parts of the paper, especially when trying to grasp complex topics or nuanced details.

In several papers analyzed in this study, model architectures are depicted through figures. While these figures are typically accompanied by explanatory subtitles, not including the visual content itself results in the loss of important details. This issue is particularly evident in responses to questions about how models function in papers like AutoML-APT \cite{automlgpt}, ChatDrug \cite{chatdrug}, and GPT for surface engineering \cite{surfacegpt}. When asked about the workflow of these models, the human reviewer provides answers that effectively translate the visual information from workflow diagrams into textual descriptions. Conversely, SciSpace's responses do not effectively convey the information presented in these graphical formats. It becomes clear that SciSpace's textual summaries alone are insufficient for capturing the intricate details and nuances often communicated through these visual representations, highlighting a significant gap in its ability to fully interpret and relay information from such crucial graphical data.

\subsection{Completeness and Relevance} 
SciSpace struggles to interpret complex model structures and theoretical frameworks, often providing overly broad or imprecise responses that don't fully address the question. Additionally, SciSpace sometimes avoids directly answering a question, which notably impacts the completeness and relevance of its responses. For example, when asked to explain a model's structure with an example, it typically focuses on the model's objectives instead. 

\begin{figure}[htbp]
    \centering
    \includegraphics[scale = 0.5]{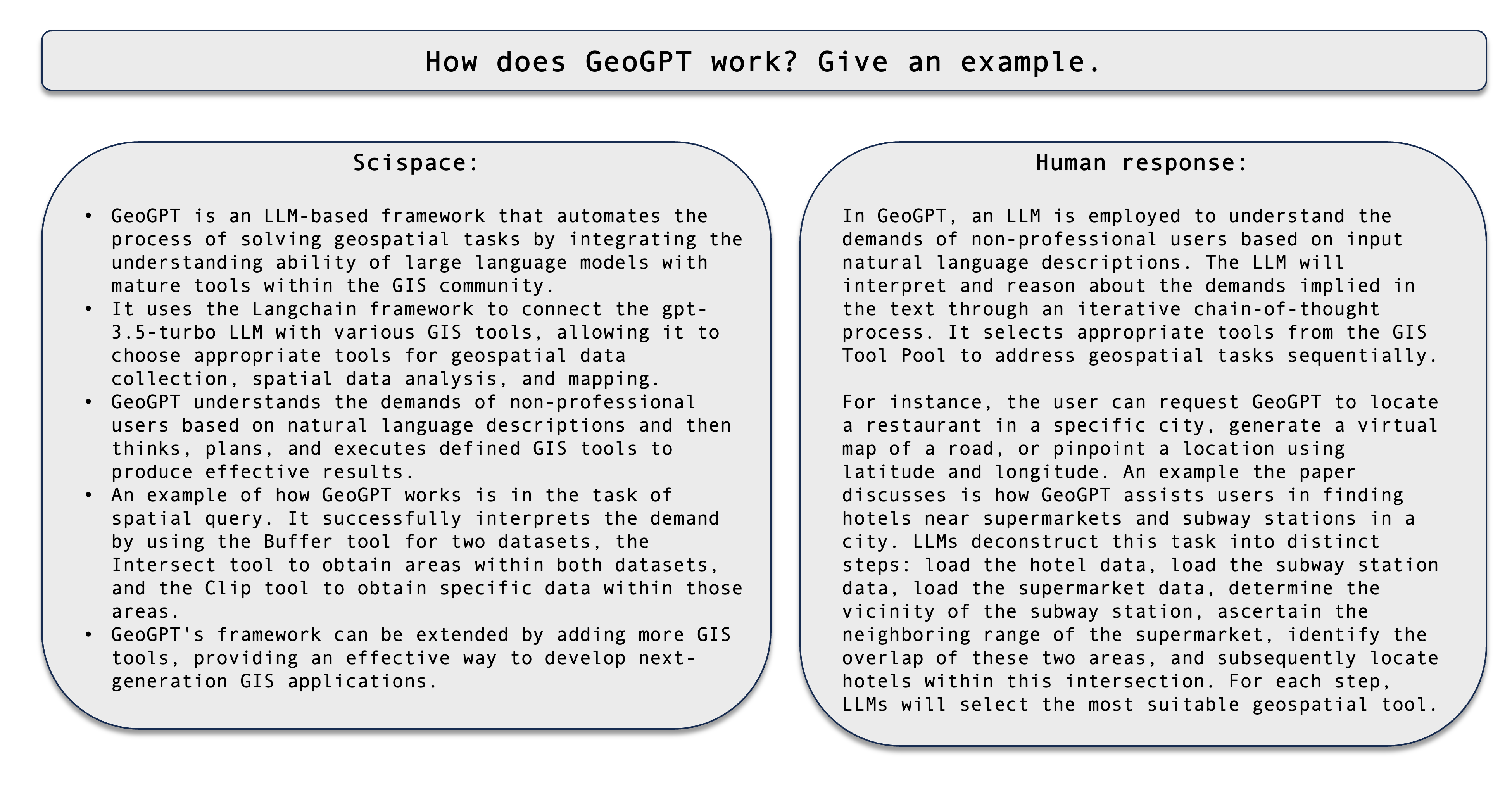}
    \caption{Comparison of Completeness and Relevance in Responses from SciSpace and Humans.}
    \label{fig:example3}
\end{figure}

In response to the question ``How does GeoGPT work? Give an example," SciSpace's answer illustrates its struggle to provide precise and direct responses to complex questions (Figure \ref{fig:example3}). While it offers a general overview of GeoGPT as an LLM-based framework that automates geospatial tasks, the response predominantly outlines the model's objectives. It touches upon the integration of large language models with GIS tools and the Langchain framework's role but does not delve deeply into specific model structures or processes. Furthermore, the provided example of how GeoGPT executes a spatial query, although somewhat relevant, still remains on a surface level. It mentions the use of various GIS tools like Buffer, Intersect, and Clip, but this description lacks the depth or detailed insight into the GeoGPT's intricate structure or the underlying theoretical framework. On the other hand, the human response delves deeper into the operational aspects of GeoGPT. It explains how the LLM interprets user demands through a chain-of-thought process to select appropriate tools from the GIS Tool Pool. The human answer also provides a more concrete and step-by-step example of how GeoGPT assists users in locating hotels near supermarkets and subway stations. This response breaks down the task into distinct steps, clearly illustrating how LLMs choose suitable geospatial tools for each phase.

\subsection{Perturbation Test}
\begin{figure}[htbp]
    \centering
    \includegraphics[scale = 0.5]{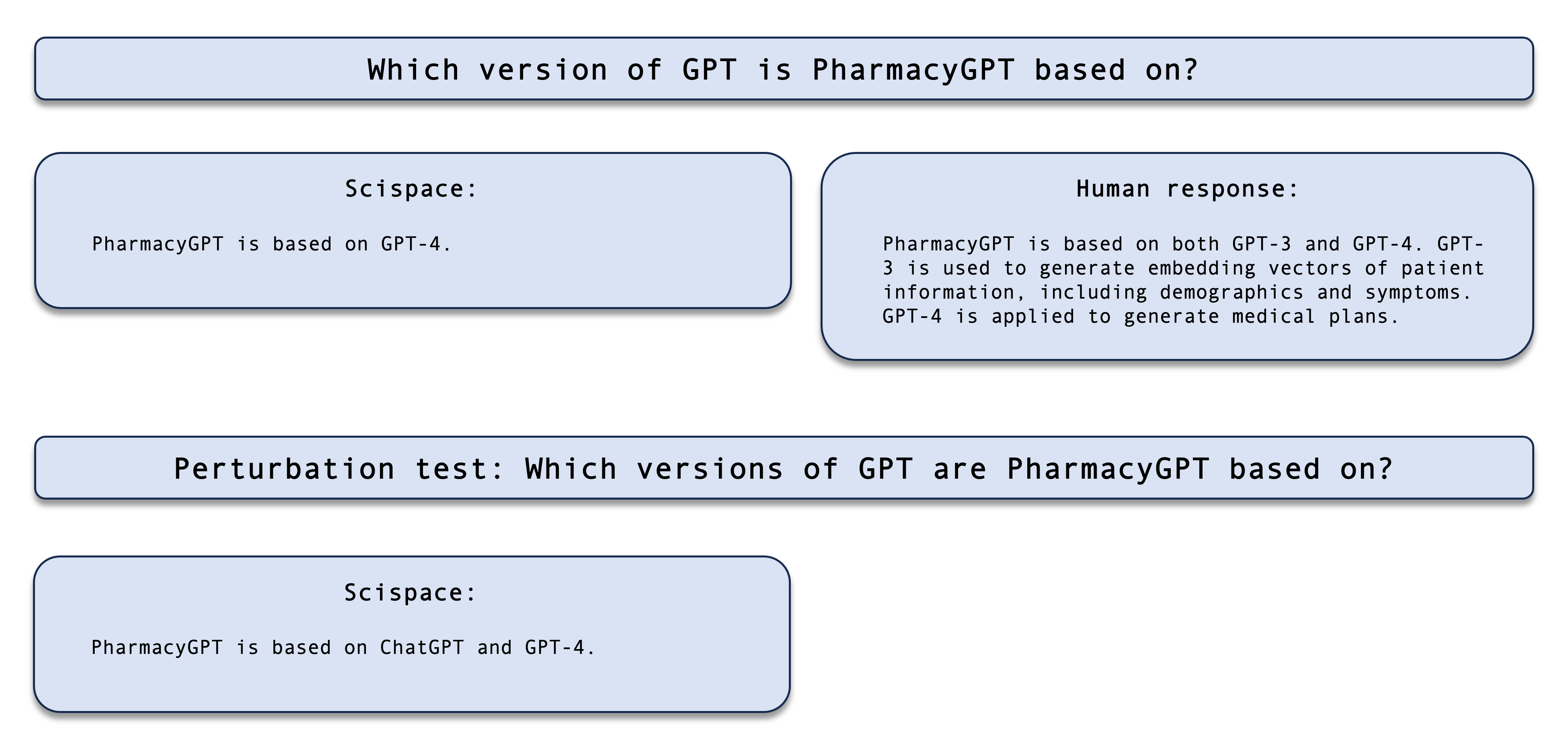}
    \caption{Perturbation Test on SciSpace's Responses.}
    \label{fig:example4}
\end{figure}

In our study, we also implemented a perturbation test (Figure \ref{fig:example4}). Considering that most papers in our dataset focus on enhancing or adapting one version of GPT for specific domains, our initial question was designed to identify which version of GPT each model was based on. However, for PharmacyGPT \cite{pharmacygpt}, which uniquely utilizes both GPT-3 and GPT-4 at different stages, this question required refinement. Initially, SciSpace identified PharmacyGPT as solely based on GPT-4. Recognizing a potential bias in our question phrasing, which might imply the use of only one GPT version, we modified the question from “Which version of GPT is PharmacyGPT based on?” to “Which versions of GPT are PharmacyGPT based on?”. After this adjustment, SciSpace's response changed to “PharmacyGPT is based on ChatGPT and GPT-4,” which, while still incorrect, showed an adaptive response to the revised question. This indicates that SciSpace's ability to interpret and respond to questions can be influenced by the specific phrasing used. This test revealed two key findings: first, SciSpace's responses can be sensitive to the phrasing of questions, indicating a responsiveness to linguistic nuances; second, despite this adaptability, SciSpace still showed limitations in accurately processing and conveying detailed, multifaceted information.

\section{Limitations}\label{section:limitation}
GPT-4 as an informed evaluator comes with certain limitations:

\begin{itemize}
    \item \textbf{Depth of Technical Analysis:} While GPT-4 with a plugin can process and analyze text, its understanding is based on patterns in data rather than human-like comprehension. Although GPT-4 can provide a general analysis of academic papers, it may not match the depth and detail that a subject matter expert in specialized fields can offer.
    \item \textbf{Data Currency:} GPT-4's knowledge is limited to the data on which it was trained, which may not include the most recent publications or the latest research developments. However, in this study, all the papers are published after 2023, potentially containing timely information. Papers introducing novel concepts that aren't well-represented in the training data may pose challenges for GPT-4's analysis. 
    \item \textbf{Subjectivity and Bias:} GPT-4's interpretations are influenced by the data it has been trained on, which can introduce biases or a lack of understanding of cultural or contextual subtleties.
\end{itemize}

\section{Summary}
We explored the utilization of large language models (LLMs), particularly the GPT series (GPT-2, GPT-3, GPT-3.5, and GPT-4) from OpenAI. We discussed  how integrating iterative questioning, chain-of-thought workflows, domain-specific databases, external tools, and few-shot in-context learning with ChatGPT has led to more accurate and relevant responses in scientific areas. Additionally, we introduced several platforms currently facilitating the development of customized GPT models for scientific research. 

Our study summarized academic papers aiming to enhance and tailor GPT models for specific scientific disciplines, including medicine, machine learning, engineering, and geography. These papers are presented in a question-answer format for clear and accessible communication, with responses compiled from SciSpace, a large language model, and a human reviewer. This dual-source approach allowed for a critical discussion on the performance of LLMs versus human reviewers in literature reviews, offering insights into the integration of AI in academic research. We found that 50\% of SciSpace's responses to objective questions align with those of a human reviewer, with GPT-4 often rating the human reviewer higher in accuracy and SciSpace higher in structure, clarity, and completeness. In subjective questions, uninformed evaluators and the crowd panel showed varying preferences between SciSpace and human responses, with the crowd panel showing a preference for the human responses. Meanwhile, GPT-4 rated them equally in accuracy and structure but favored SciSpace for completeness.

Our study noted that SciSpace's responses are often repetitive, verbose, and lack a coherent structure, with a tendency to miss the essence of the articles and evade challenging questions by reiterating the paper's objectives. Additionally, while SciSpace's answers can adjust to the question's wording, a notable limitation is its inability to interpret graphical data. Additionally, SciSpace's answers demonstrate adaptability to the specific wording of questions. However, a notable limitation of SciSpace is its inability to interpret graphical data, which restricts its effectiveness in contexts where visual information plays a crucial role. Our research provides both qualitative and quantitative analysis of the comprehension levels of SciSpace, an LLM-based system, and a human reviewer regarding contemporary scientific articles, incorporating assessments from evaluators with varying levels of information.

\section{Acknowledgments}
We sincerely thank the 25 members of the Crunch Group at Brown University for their vital role as the crowd panel in our study. Their contributions and insights have greatly enhanced our research, and we deeply appreciate their dedication and support. This work was supported by the ONR Vannevar Bush Faculty Fellowship (N00014-22-1-2795)

\bibliographystyle{unsrt}
\bibliography{main}

\begin{thebibliography}{10}

\bibitem{LLM_solveproblem1}
Arun~James Thirunavukarasu, Darren Shu~Jeng Ting, Kabilan Elangovan, Laura Gutierrez, Ting~Fang Tan, and Daniel Shu~Wei Ting.
\newblock Large language models in medicine.
\newblock {\em Nature medicine}, 29(8):1930--1940, 2023.

\bibitem{LLM_solveproblem2}
Yiheng Liu, Tianle Han, Siyuan Ma, Jiayue Zhang, Yuanyuan Yang, Jiaming Tian, Hao He, Antong Li, Mengshen He, Zhengliang Liu, et~al.
\newblock Summary of chatgpt-related research and perspective towards the future of large language models.
\newblock {\em Meta-Radiology}, page 100017, 2023.

\bibitem{LLM_solveproblem3}
Shijie Wu, Ozan Irsoy, Steven Lu, Vadim Dabravolski, Mark Dredze, Sebastian Gehrmann, Prabhanjan Kambadur, David Rosenberg, and Gideon Mann.
\newblock Bloomberggpt: A large language model for finance.
\newblock {\em arXiv preprint arXiv:2303.17564}, 2023.

\bibitem{LLM_solveproblem4}
Xinyi Hou, Yanjie Zhao, Yue Liu, Zhou Yang, Kailong Wang, Li~Li, Xiapu Luo, David Lo, John Grundy, and Haoyu Wang.
\newblock Large language models for software engineering: A systematic literature review.
\newblock {\em arXiv preprint arXiv:2308.10620}, 2023.

\bibitem{LLM_solveproblem5}
Zeming Lin, Halil Akin, Roshan Rao, Brian Hie, Zhongkai Zhu, Wenting Lu, Nikita Smetanin, Robert Verkuil, Ori Kabeli, Yaniv Shmueli, et~al.
\newblock Evolutionary-scale prediction of atomic-level protein structure with a language model.
\newblock {\em Science}, 379(6637):1123--1130, 2023.

\bibitem{LLM_solveproblem6}
Joshua Meier, Roshan Rao, Robert Verkuil, Jason Liu, Tom Sercu, and Alex Rives.
\newblock Language models enable zero-shot prediction of the effects of mutations on protein function.
\newblock {\em Advances in Neural Information Processing Systems}, 34:29287--29303, 2021.

\bibitem{LLM_solveproblem7}
Humza Naveed, Asad~Ullah Khan, Shi Qiu, Muhammad Saqib, Saeed Anwar, Muhammad Usman, Nick Barnes, and Ajmal Mian.
\newblock A comprehensive overview of large language models.
\newblock {\em arXiv preprint arXiv:2307.06435}, 2023.

\bibitem{LLM_speech1}
Jaeho Jeon, Seongyong Lee, and Seongyune Choi.
\newblock A systematic review of research on speech-recognition chatbots for language learning: Implications for future directions in the era of large language models.
\newblock {\em Interactive Learning Environments}, pages 1--19, 2023.

\bibitem{LLM_speech2}
Dong Zhang, Shimin Li, Xin Zhang, Jun Zhan, Pengyu Wang, Yaqian Zhou, and Xipeng Qiu.
\newblock Speechgpt: Empowering large language models with intrinsic cross-modal conversational abilities.
\newblock {\em arXiv preprint arXiv:2305.11000}, 2023.

\bibitem{LLM_media1}
KunChang Li, Yinan He, Yi~Wang, Yizhuo Li, Wenhai Wang, Ping Luo, Yali Wang, Limin Wang, and Yu~Qiao.
\newblock Videochat: Chat-centric video understanding.
\newblock {\em arXiv preprint arXiv:2305.06355}, 2023.

\bibitem{LLM_media2}
Chenfei Wu, Shengming Yin, Weizhen Qi, Xiaodong Wang, Zecheng Tang, and Nan Duan.
\newblock Visual chatgpt: Talking, drawing and editing with visual foundation models.
\newblock {\em arXiv preprint arXiv:2303.04671}, 2023.

\bibitem{LLM_coding}
Russell~A Poldrack, Thomas Lu, and Ga{\v{s}}per Begu{\v{s}}.
\newblock Ai-assisted coding: Experiments with gpt-4.
\newblock {\em arXiv preprint arXiv:2304.13187}, 2023.

\bibitem{LLM_exam1}
Daniel~Martin Katz, Michael~James Bommarito, Shang Gao, and Pablo Arredondo.
\newblock Gpt-4 passes the bar exam.
\newblock {\em Available at SSRN 4389233}, 2023.

\bibitem{LLM_exam2}
Vinay Pursnani, Yusuf Sermet, Musa Kurt, and Ibrahim Demir.
\newblock Performance of chatgpt on the us fundamentals of engineering exam: Comprehensive assessment of proficiency and potential implications for professional environmental engineering practice.
\newblock {\em Computers and Education: Artificial Intelligence}, page 100183, 2023.

\bibitem{LLM_exam3}
Yiran Wu, Feiran Jia, Shaokun Zhang, Qingyun Wu, Hangyu Li, Erkang Zhu, Yue Wang, Yin~Tat Lee, Richard Peng, and Chi Wang.
\newblock An empirical study on challenging math problem solving with gpt-4.
\newblock {\em arXiv preprint arXiv:2306.01337}, 2023.

\bibitem{gpt-4inscience}
Microsoft~Research AI4Science and Microsoft~Azure Quantum.
\newblock The impact of large language models on scientific discovery: a preliminary study using gpt-4.
\newblock {\em arXiv preprint arXiv:2311.07361}, 2023.

\bibitem{gpt4_technical_report}
OpenAI.
\newblock Gpt-4 technical report, 2023.

\bibitem{palm2}
Rohan Anil, Andrew~M Dai, Orhan Firat, Melvin Johnson, Dmitry Lepikhin, Alexandre Passos, Siamak Shakeri, Emanuel Taropa, Paige Bailey, Zhifeng Chen, et~al.
\newblock Palm 2 technical report.
\newblock {\em arXiv preprint arXiv:2305.10403}, 2023.

\bibitem{Anthropic}
Claude 2.
\newblock \url{https://www.anthropic.com/index/claude-2}, 2023.

\bibitem{llama}
Hugo Touvron, Louis Martin, Kevin Stone, Peter Albert, Amjad Almahairi, Yasmine Babaei, Nikolay Bashlykov, Soumya Batra, Prajjwal Bhargava, Shruti Bhosale, et~al.
\newblock Llama 2: Open foundation and fine-tuned chat models.
\newblock {\em arXiv preprint arXiv:2307.09288}, 2023.

\bibitem{potential_publichealth1}
Som~S Biswas.
\newblock Role of chat gpt in public health.
\newblock {\em Annals of biomedical engineering}, 51(5):868--869, 2023.

\bibitem{potential_publichealth2}
Malik Sallam.
\newblock The utility of chatgpt as an example of large language models in healthcare education, research and practice: Systematic review on the future perspectives and potential limitations.
\newblock {\em medRxiv}, pages 2023--02, 2023.

\bibitem{potential_publichealth3}
Ridwan~Islam Sifat.
\newblock Chatgpt and the future of health policy analysis: potential and pitfalls of using chatgpt in policymaking.
\newblock {\em Annals of Biomedical Engineering}, pages 1--3, 2023.

\bibitem{potential_publichealth4}
Yuqing Wang, Yun Zhao, and Linda Petzold.
\newblock Are large language models ready for healthcare? a comparative study on clinical language understanding.
\newblock {\em arXiv preprint arXiv:2304.05368}, 2023.

\bibitem{potential_medicine1}
Asser~Abou Elkassem and Andrew~D Smith.
\newblock Potential use cases for chatgpt in radiology reporting.
\newblock {\em American Journal of Roentgenology}, 2023.

\bibitem{potential_medicine2}
Ahmed Ismail, Nima~S Ghorashi, and Ramin Javan.
\newblock New horizons: the potential role of openai’s chatgpt in clinical radiology.
\newblock {\em Journal of the American College of Radiology}, 2023.

\bibitem{potential_medicine3}
Peter Lee, Sebastien Bubeck, and Joseph Petro.
\newblock Benefits, limits, and risks of gpt-4 as an ai chatbot for medicine.
\newblock {\em New England Journal of Medicine}, 388(13):1233--1239, 2023.

\bibitem{potential_medicine4}
Harsha Nori, Nicholas King, Scott~Mayer McKinney, Dean Carignan, and Eric Horvitz.
\newblock Capabilities of gpt-4 on medical challenge problems.
\newblock {\em arXiv preprint arXiv:2303.13375}, 2023.

\bibitem{potential_medicine5}
Ethan Waisberg, Joshua Ong, Mouayad Masalkhi, Sharif~Amit Kamran, Nasif Zaman, Prithul Sarker, Andrew~G Lee, and Alireza Tavakkoli.
\newblock Gpt-4: a new era of artificial intelligence in medicine.
\newblock {\em Irish Journal of Medical Science (1971-)}, pages 1--4, 2023.

\bibitem{potential_medicine6}
Shahab~Saquib Sohail.
\newblock A promising start and not a panacea: Chatgpt's early impact and potential in medical science and biomedical engineering research.
\newblock {\em Annals of Biomedical Engineering}, pages 1--5, 2023.

\bibitem{potential_medicine7}
Zhengliang Liu, Aoxiao Zhong, Yiwei Li, Longtao Yang, Chao Ju, Zihao Wu, Chong Ma, Peng Shu, Cheng Chen, Sekeun Kim, et~al.
\newblock Radiology-gpt: A large language model for radiology.
\newblock {\em arXiv preprint arXiv:2306.08666}, 2023.

\bibitem{potential_medicine8}
Youwei Liang, Ruiyi Zhang, Li~Zhang, and Pengtao Xie.
\newblock Drugchat: towards enabling chatgpt-like capabilities on drug molecule graphs.
\newblock {\em arXiv preprint arXiv:2309.03907}, 2023.

\bibitem{potential_education1}
Hyunsu Lee.
\newblock The rise of chatgpt: Exploring its potential in medical education.
\newblock {\em Anatomical Sciences Education}, 2023.

\bibitem{potential_education2}
Tufan Adiguzel, Mehmet~Haldun Kaya, and Fatih~K{\"u}r{\c{s}}at Cansu.
\newblock Revolutionizing education with ai: Exploring the transformative potential of chatgpt.
\newblock {\em Contemporary Educational Technology}, 15(3):ep429, 2023.

\bibitem{potential_education3}
David Baidoo-Anu and Leticia~Owusu Ansah.
\newblock Education in the era of generative artificial intelligence (ai): Understanding the potential benefits of chatgpt in promoting teaching and learning.
\newblock {\em Journal of AI}, 7(1):52--62, 2023.

\bibitem{potential_education4}
Simone Grassini.
\newblock Shaping the future of education: exploring the potential and consequences of ai and chatgpt in educational settings.
\newblock {\em Education Sciences}, 13(7):692, 2023.

\bibitem{potential_education5}
Md~Mostafizer Rahman and Yutaka Watanobe.
\newblock Chatgpt for education and research: Opportunities, threats, and strategies.
\newblock {\em Applied Sciences}, 13(9):5783, 2023.

\bibitem{potential_education6}
Shunsuke Koga.
\newblock The potential of chatgpt in medical education: Focusing on usmle preparation.
\newblock {\em Annals of Biomedical Engineering}, pages 1--2, 2023.

\bibitem{potential_education7}
Henner Gimpel, Kristina Hall, Stefan Decker, Torsten Eymann, Luis L{\"a}mmermann, Alexander M{\"a}dche, Maximilian R{\"o}glinger, Caroline Ruiner, Manfred Schoch, Mareike Schoop, et~al.
\newblock Unlocking the power of generative ai models and systems such as gpt-4 and chatgpt for higher education: A guide for students and lecturers.
\newblock Technical report, Hohenheim Discussion Papers in Business, Economics and Social Sciences, 2023.

\bibitem{potential_environment1}
Som~S Biswas.
\newblock Potential use of chat gpt in global warming.
\newblock {\em Annals of biomedical engineering}, 51(6):1126--1127, 2023.

\bibitem{potential_math1}
Giuseppe Orlando.
\newblock Assessing chatgpt for coding finite element methods.
\newblock {\em Journal of Machine Learning for Modeling and Computing}, 4(2), 2023.

\bibitem{adautogpt}
Haixing Dai, Yiwei Li, Zhengliang Liu, Lin Zhao, Zihao Wu, Suhang Song, Ye~Shen, Dajiang Zhu, Xiang Li, Sheng Li, et~al.
\newblock Ad-autogpt: An autonomous gpt for alzheimer's disease infodemiology.
\newblock {\em arXiv preprint arXiv:2306.10095}, 2023.

\bibitem{automlgpt}
Shujian Zhang, Chengyue Gong, Lemeng Wu, Xingchao Liu, and Mingyuan Zhou.
\newblock Automl-gpt: Automatic machine learning with gpt.
\newblock {\em arXiv preprint arXiv:2305.02499}, 2023.

\bibitem{chatdrug}
Shengchao Liu, Jiongxiao Wang, Yijin Yang, Chengpeng Wang, Ling Liu, Hongyu Guo, and Chaowei Xiao.
\newblock Chatgpt-powered conversational drug editing using retrieval and domain feedback.
\newblock {\em arXiv preprint arXiv:2305.18090}, 2023.

\bibitem{chemcrow}
Andres~M Bran, Sam Cox, Oliver Schilter, Carlo Baldassari, Andrew White, and Philippe Schwaller.
\newblock Augmenting large language models with chemistry tools.
\newblock In {\em NeurIPS 2023 AI for Science Workshop}, 2023.

\bibitem{cancergpt}
Tianhao Li, Sandesh Shetty, Advaith Kamath, Ajay Jaiswal, Xiaoqian Jiang, Ying Ding, and Yejin Kim.
\newblock Cancergpt: Few-shot drug pair synergy prediction using large pre-trained language models.
\newblock {\em ArXiv}, 2023.

\bibitem{cohortgpt}
Zihan Guan, Zihao Wu, Zhengliang Liu, Dufan Wu, Hui Ren, Quanzheng Li, Xiang Li, and Ninghao Liu.
\newblock Cohortgpt: An enhanced gpt for participant recruitment in clinical study.
\newblock {\em arXiv preprint arXiv:2307.11346}, 2023.

\bibitem{genegpt}
Qiao Jin, Yifan Yang, Qingyu Chen, and Zhiyong Lu.
\newblock Genegpt: Augmenting large language models with domain tools for improved access to biomedical information.
\newblock {\em ArXiv}, 2023.

\bibitem{geogpt}
Yifan Zhang, Cheng Wei, Shangyou Wu, Zhengting He, and Wenhao Yu.
\newblock Geogpt: Understanding and processing geospatial tasks through an autonomous gpt.
\newblock {\em arXiv preprint arXiv:2307.07930}, 2023.

\bibitem{geotechgpt}
Krishna Kumar.
\newblock Geotechnical parrot tales (gpt): Harnessing large language models in geotechnical engineering.
\newblock {\em Journal of Geotechnical and Geoenvironmental Engineering}, 150(1):02523001, 2024.

\bibitem{mycrunchgpt}
Varun Kumar, Leonard Gleyzer, Adar Kahana, Khemraj Shukla, and George~Em Karniadakis.
\newblock Mycrunchgpt: A llm assisted framework for scientific machine learning.
\newblock {\em Journal of Machine Learning for Modeling and Computing}, 4(4), 2023.

\bibitem{pharmacygpt}
Zhengliang Liu, Zihao Wu, Mengxuan Hu, Bokai Zhao, Lin Zhao, Tianyi Zhang, Haixing Dai, Xianyan Chen, Ye~Shen, Sheng Li, et~al.
\newblock Pharmacygpt: The ai pharmacist.
\newblock {\em arXiv preprint arXiv:2307.10432}, 2023.

\bibitem{roboticsgpt}
Sai Vemprala, Rogerio Bonatti, Arthur Bucker, and Ashish Kapoor.
\newblock Chatgpt for robotics: Design principles and model abilities.
\newblock {\em Microsoft Auton. Syst. Robot. Res}, 2:20, 2023.

\bibitem{surfacegpt}
Spyros Kamnis.
\newblock Generative pre-trained transformers (gpt) for surface engineering.
\newblock {\em Surface and Coatings Technology}, page 129680, 2023.

\bibitem{synergpt}
Carl~N Edwards, Aakanksha Naik, Tushar Khot, Martin~D Burke, Heng Ji, and Tom Hope.
\newblock Synergpt: In-context learning for personalized drug synergy prediction and drug design.
\newblock {\em bioRxiv}, pages 2023--07, 2023.

\bibitem{chatmof}
Yeonghun Kang and Jihan Kim.
\newblock Chatmof: An autonomous ai system for predicting and generating metal-organic frameworks.
\newblock {\em arXiv preprint arXiv:2308.01423}, 2023.

\bibitem{hypotheses}
Yang~Jeong Park, Daniel Kaplan, Zhichu Ren, Chia-Wei Hsu, Changhao Li, Haowei Xu, Sipei Li, and Ju~Li.
\newblock Can chatgpt be used to generate scientific hypotheses?
\newblock {\em arXiv preprint arXiv:2304.12208}, 2023.

\bibitem{LLM_literature}
Yupeng Chang, Xu~Wang, Jindong Wang, Yuan Wu, Kaijie Zhu, Hao Chen, Linyi Yang, Xiaoyuan Yi, Cunxiang Wang, Yidong Wang, et~al.
\newblock A survey on evaluation of large language models.
\newblock {\em arXiv preprint arXiv:2307.03109}, 2023.

\bibitem{LLM_interdisciplinary}
James Boyko, Joseph Cohen, Nathan Fox, Maria~Han Veiga, Jennifer~I Li, Jing Liu, Bernardo Modenesi, Andreas~H Rauch, Kenneth~N Reid, Soumi Tribedi, et~al.
\newblock An interdisciplinary outlook on large language models for scientific research.
\newblock {\em arXiv preprint arXiv:2311.04929}, 2023.

\bibitem{scispace_webpage}
Ai chat for scientific pdfs | scispace, Nov 2023.
\newblock Available at \url{https://typeset.io/t/about/}.

\bibitem{chain-of-thought1}
Jason Wei, Xuezhi Wang, Dale Schuurmans, Maarten Bosma, Fei Xia, Ed~Chi, Quoc~V Le, Denny Zhou, et~al.
\newblock Chain-of-thought prompting elicits reasoning in large language models.
\newblock {\em Advances in Neural Information Processing Systems}, 35:24824--24837, 2022.

\bibitem{chain-of-thought2}
Shunyu Yao, Jeffrey Zhao, Dian Yu, Nan Du, Izhak Shafran, Karthik Narasimhan, and Yuan Cao.
\newblock React: Synergizing reasoning and acting in language models.
\newblock {\em arXiv preprint arXiv:2210.03629}, 2022.

\bibitem{LLM_media4}
Yongliang Shen, Kaitao Song, Xu~Tan, Dongsheng Li, Weiming Lu, and Yueting Zhuang.
\newblock Hugginggpt: Solving ai tasks with chatgpt and its friends in huggingface.
\newblock {\em arXiv preprint arXiv:2303.17580}, 2023.

\bibitem{autogen}
Qingyun Wu, Gagan Bansal, Jieyu Zhang, Yiran Wu, Shaokun Zhang, Erkang Zhu, Beibin Li, Li~Jiang, Xiaoyun Zhang, and Chi Wang.
\newblock Autogen: Enabling next-gen llm applications via multi-agent conversation framework.
\newblock {\em arXiv preprint arXiv:2308.08155}, 2023.

\end{thebibliography}

\clearpage 
\appendix
\section{Appendix} \label{Appendix}
\textbf{Medicines:}\\

SynerGPT: In-Context Learning for Personalized Drug Synergy Prediction and Drug Design \cite{synergpt}

\begin{figure}[htbp]
    \centering
    \includegraphics[scale = 0.6]{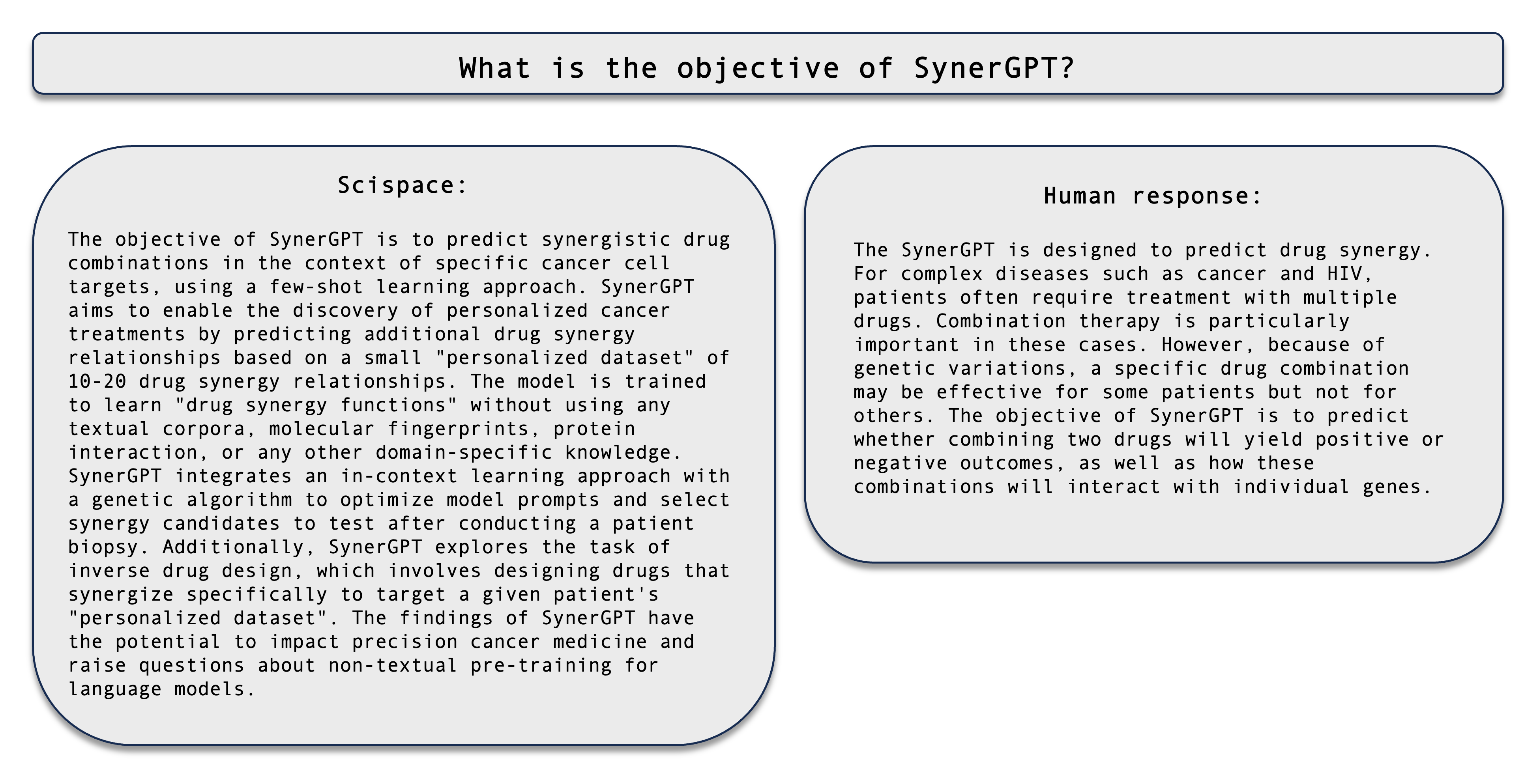}
\end{figure}

\begin{figure}[htbp]
    \includegraphics[scale = 0.6]{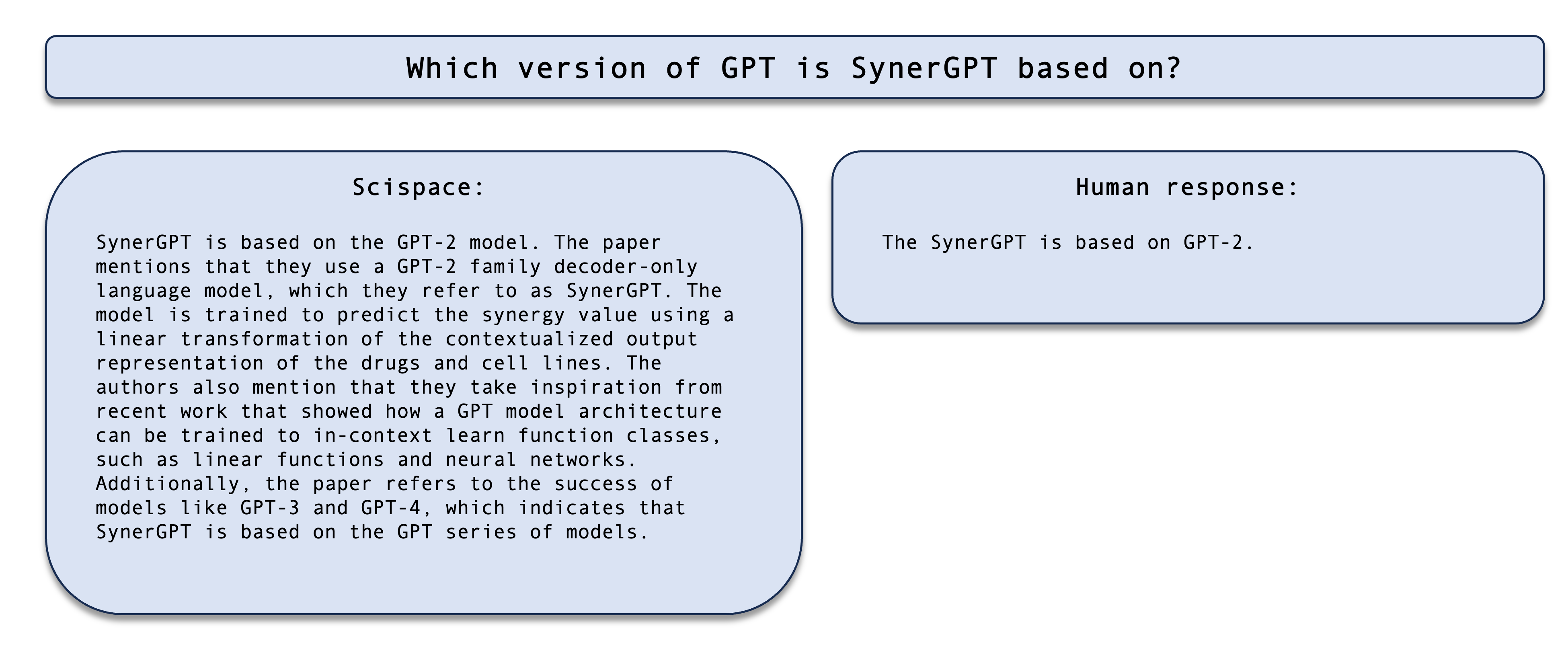}
\end{figure}

\begin{figure}[htbp]
    \includegraphics[scale = 0.6]{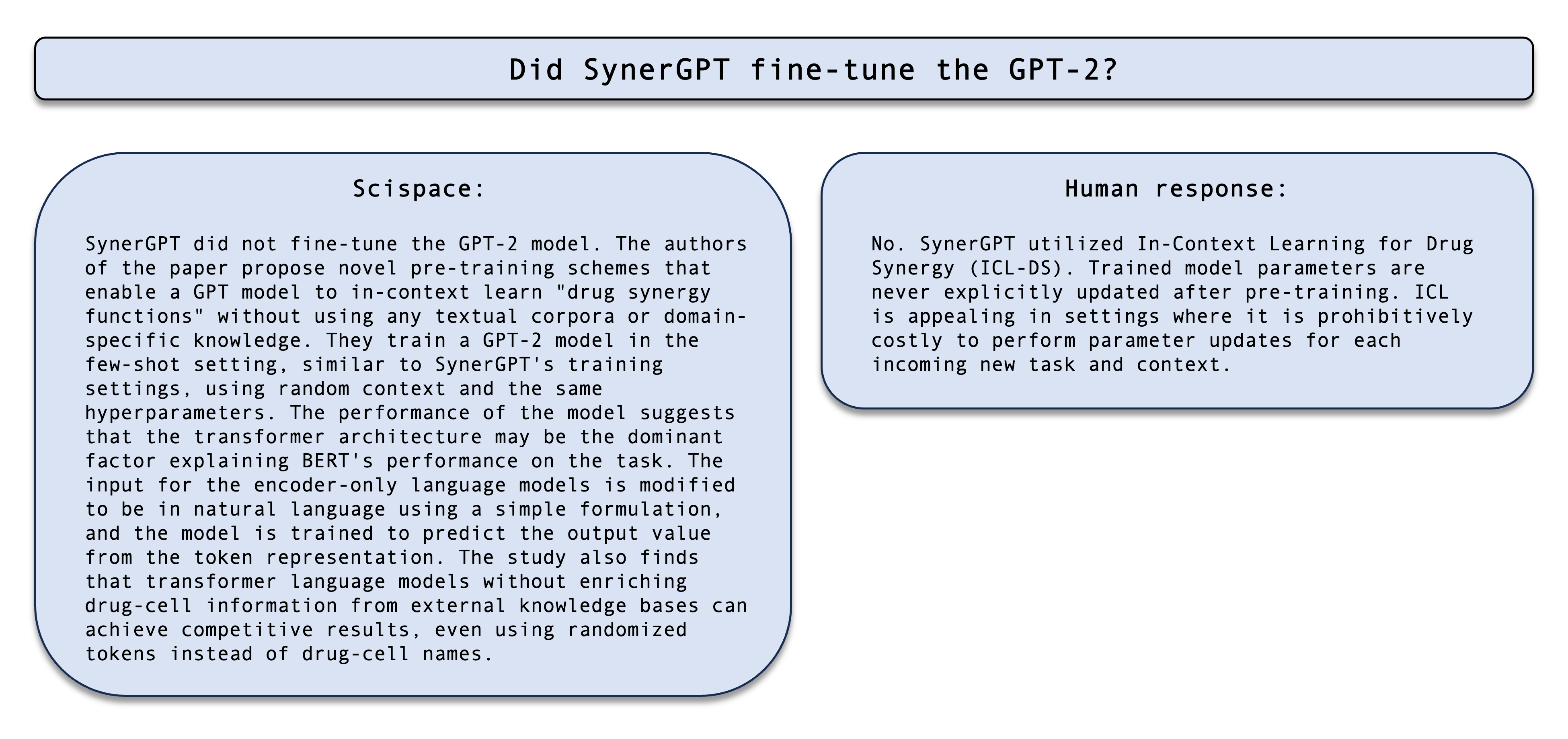}
\end{figure}

\begin{figure}[htbp]
    \includegraphics[scale = 0.6]{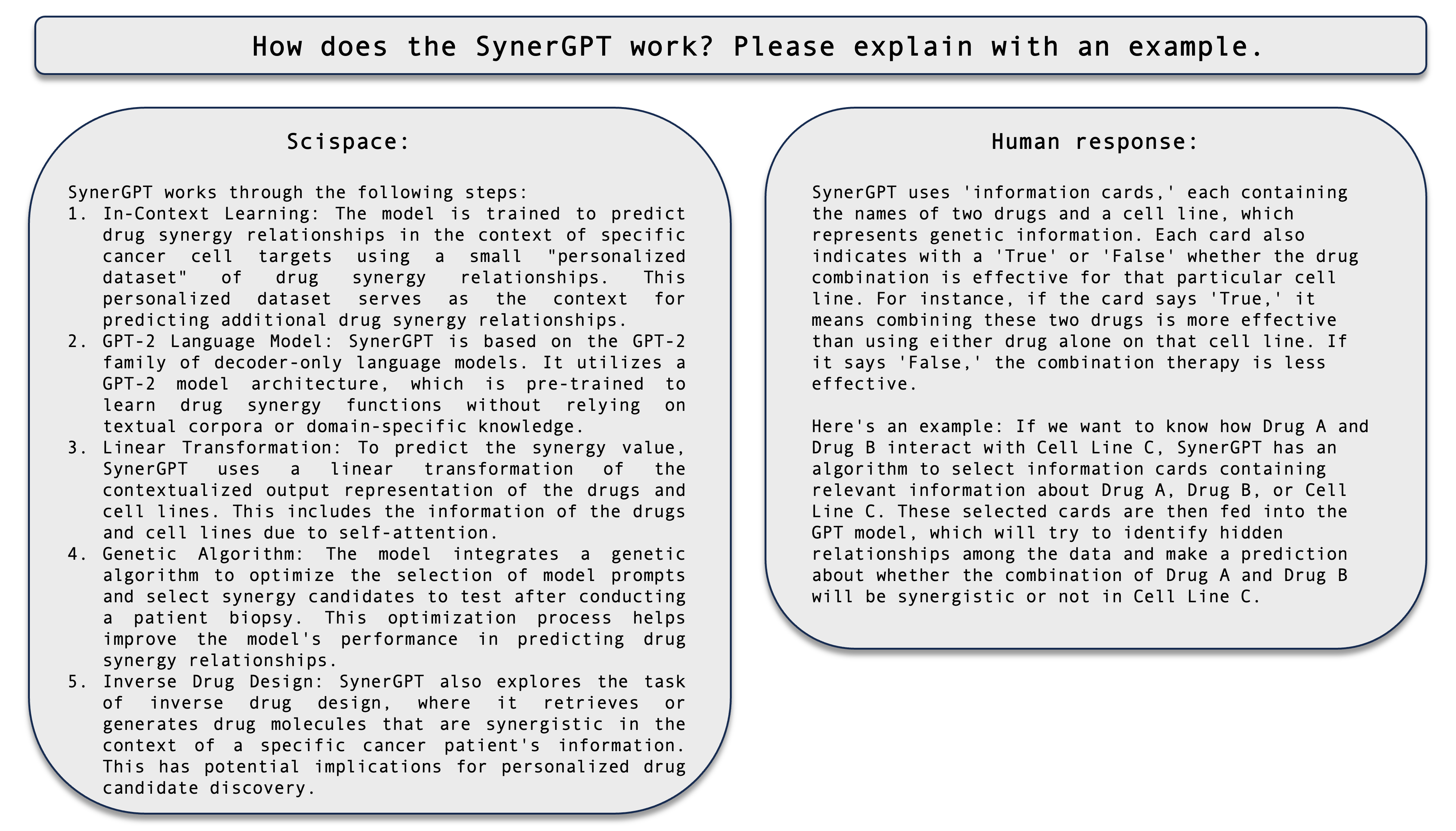}
\end{figure}

\begin{figure}[htbp]
    \includegraphics[scale = 0.6]{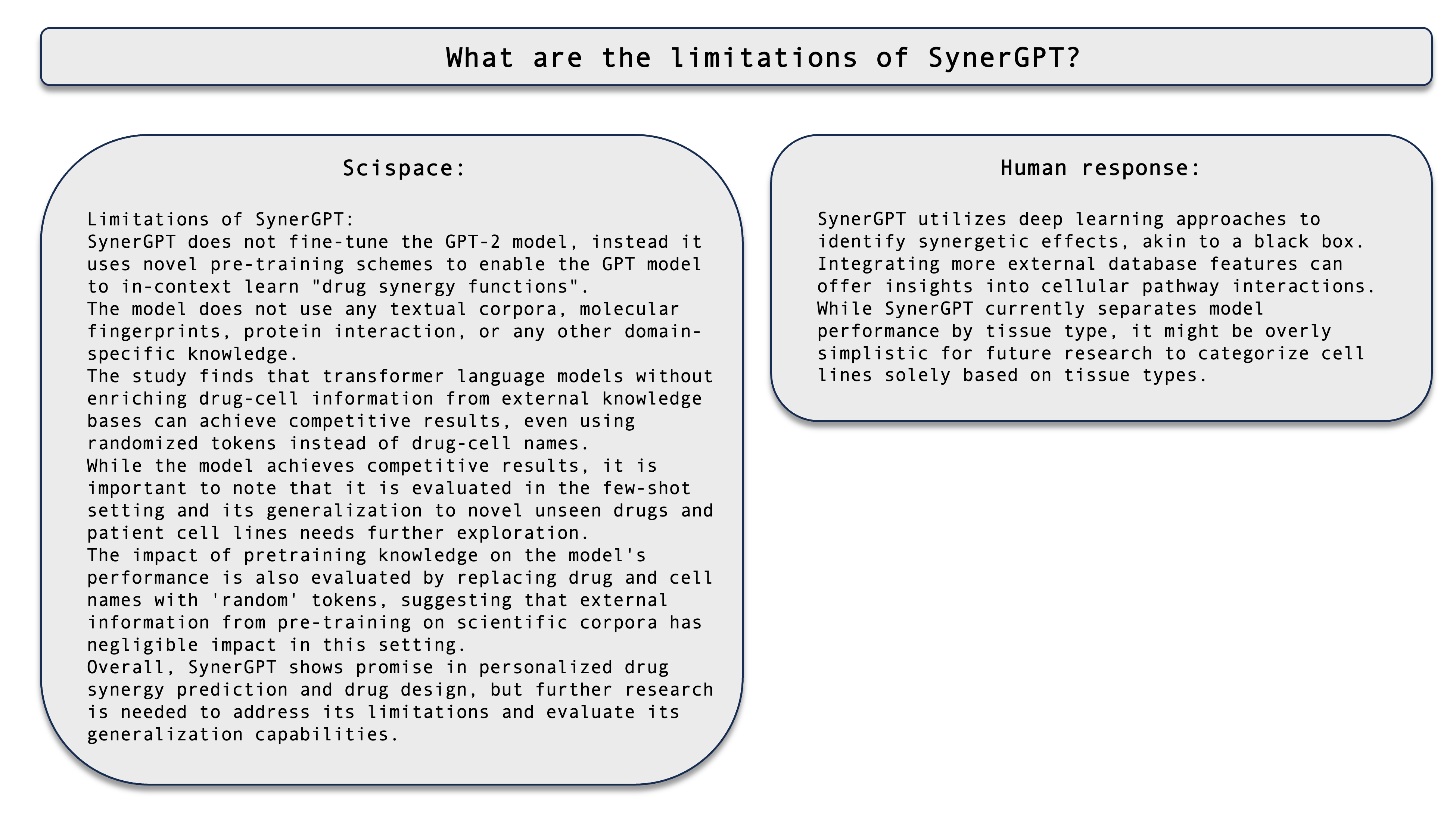}
\end{figure}

\clearpage
CancerGPT: Few-shot Drug Pair Synergy Prediction using Large Pre-trained Language Models \cite{cancergpt}

\begin{figure}[htbp]
    \centering
    \includegraphics[scale = 0.6]{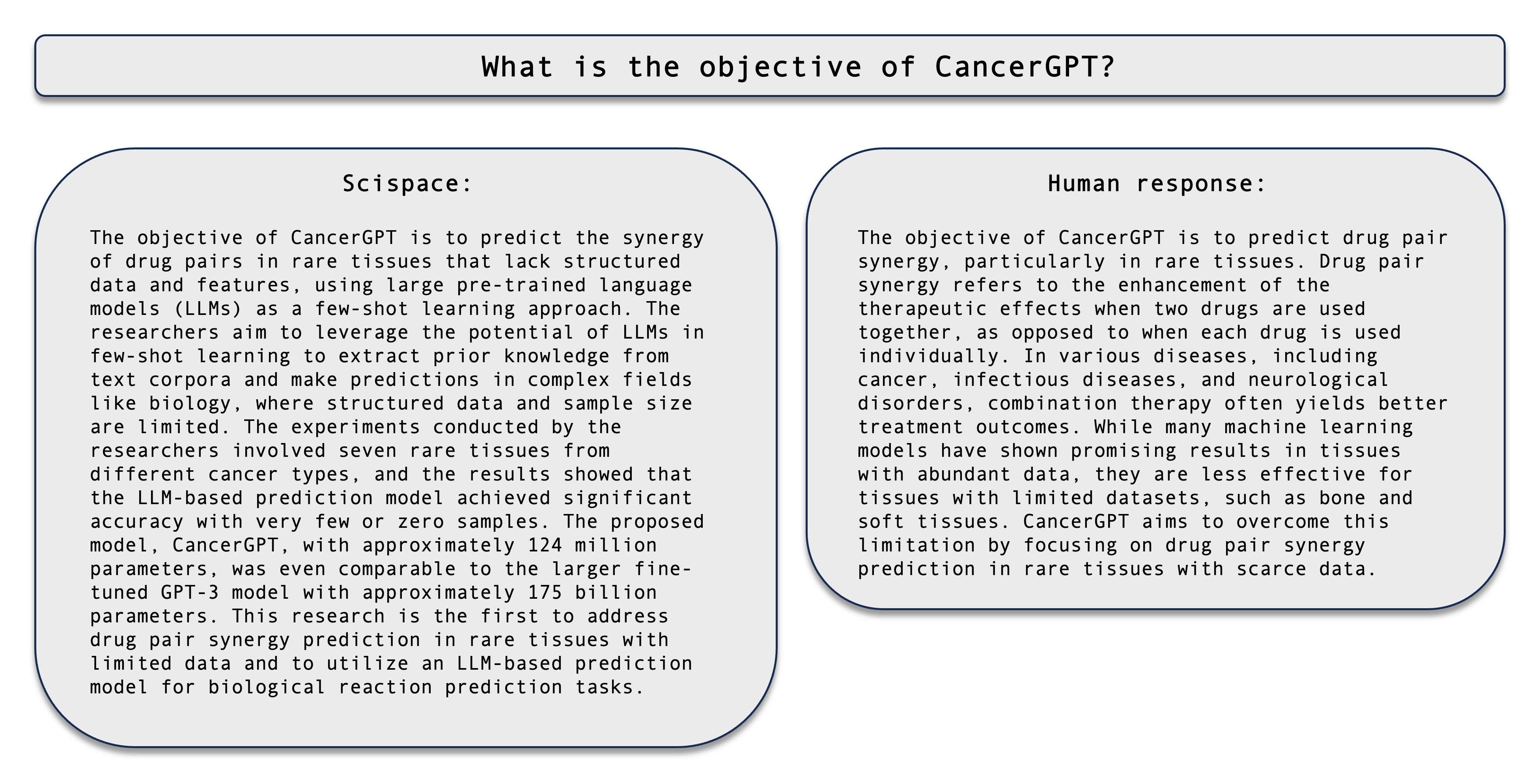}
\end{figure}

\begin{figure}[htbp]
    \centering
    \includegraphics[scale = 0.6]{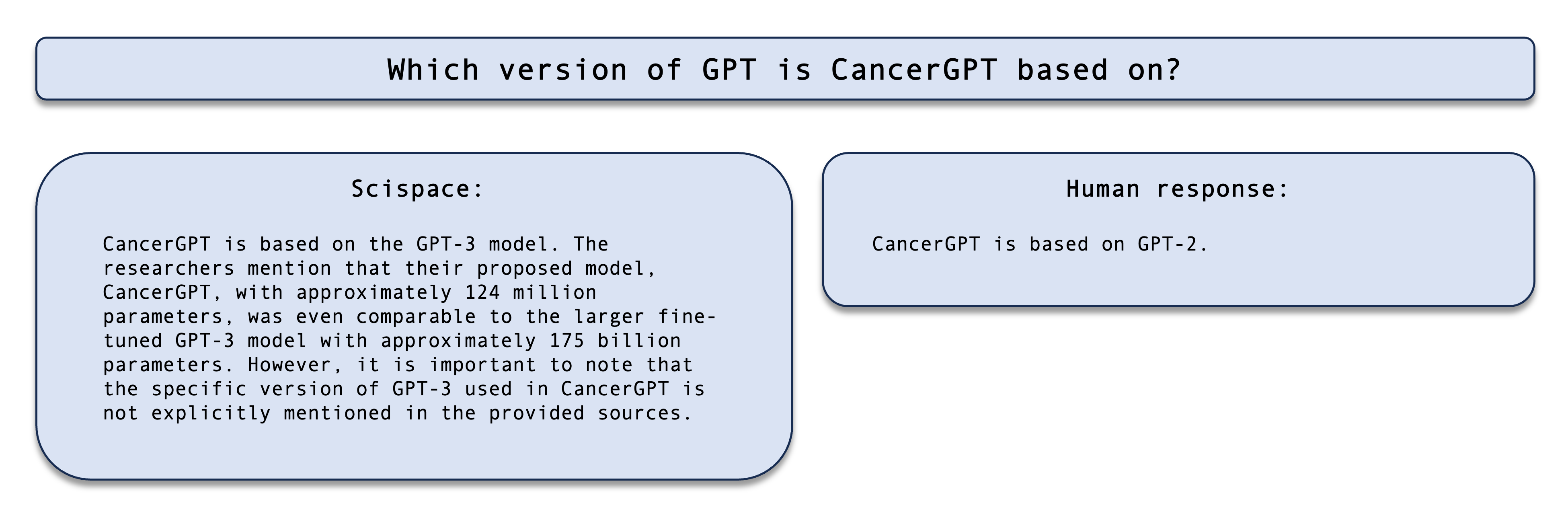}
\end{figure}

\begin{figure}[htbp]
    \centering
    \includegraphics[scale = 0.6]{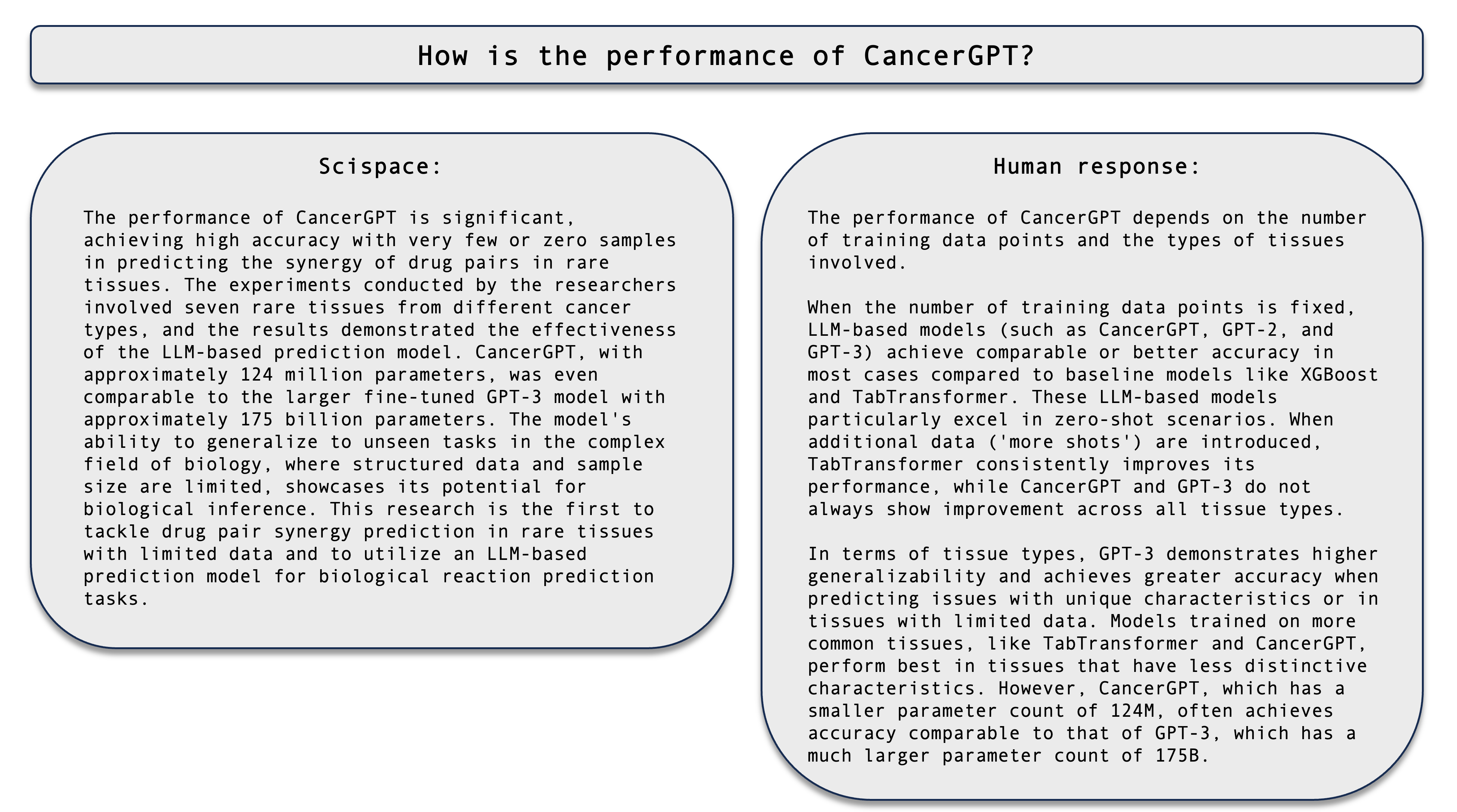}
\end{figure}

\begin{figure}[htbp]
    \centering
    \includegraphics[scale = 0.6]{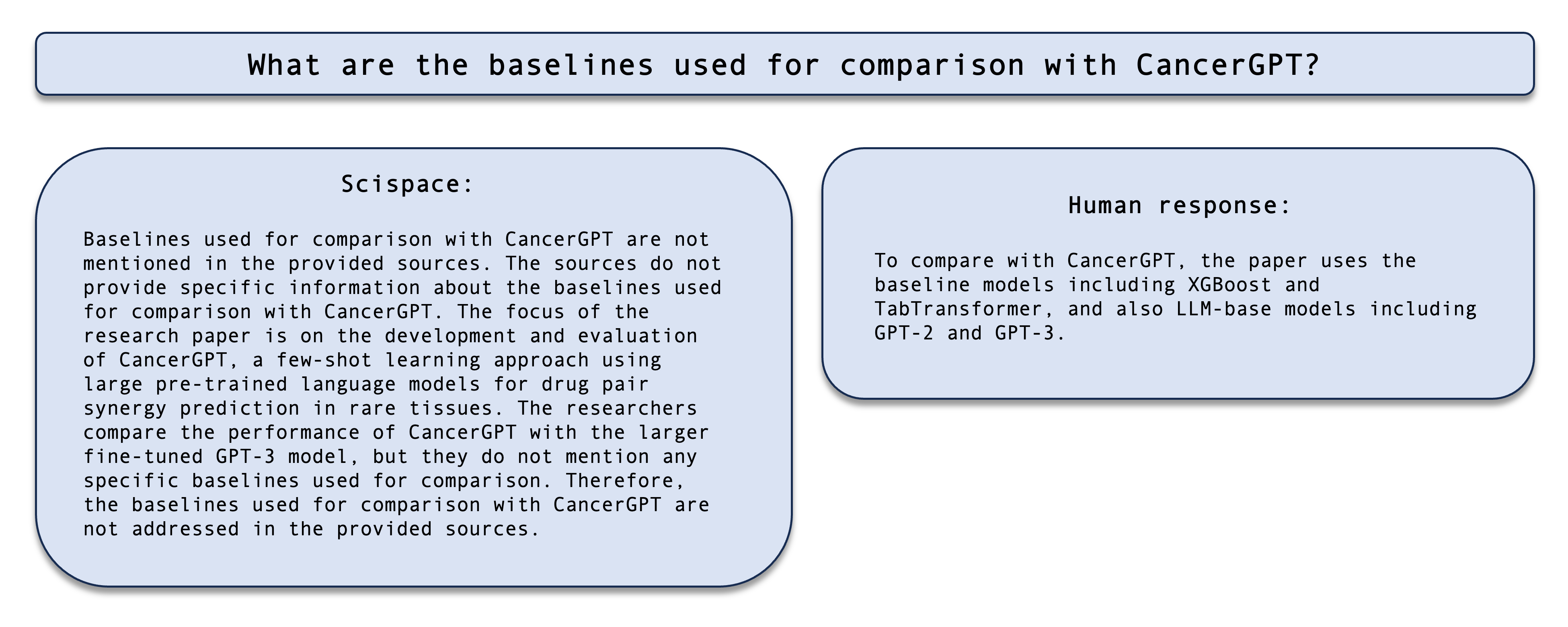}
\end{figure}

\begin{figure}[htbp]
    \centering
    \includegraphics[scale = 0.6]{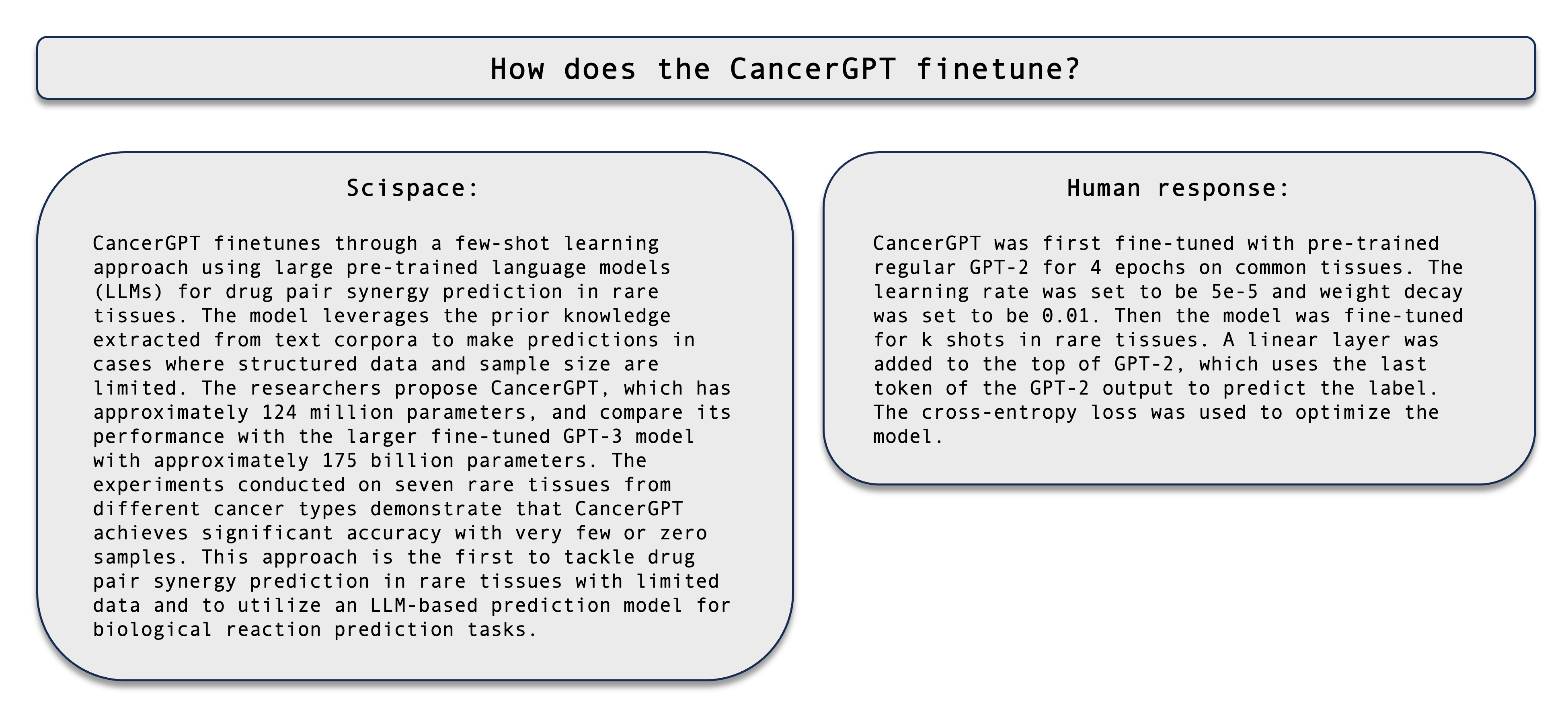}
\end{figure}

\clearpage
PharmacyGPT: The AI Pharmacist \cite{pharmacygpt}

\begin{figure}[htbp]
    \centering
    \includegraphics[scale = 0.6]{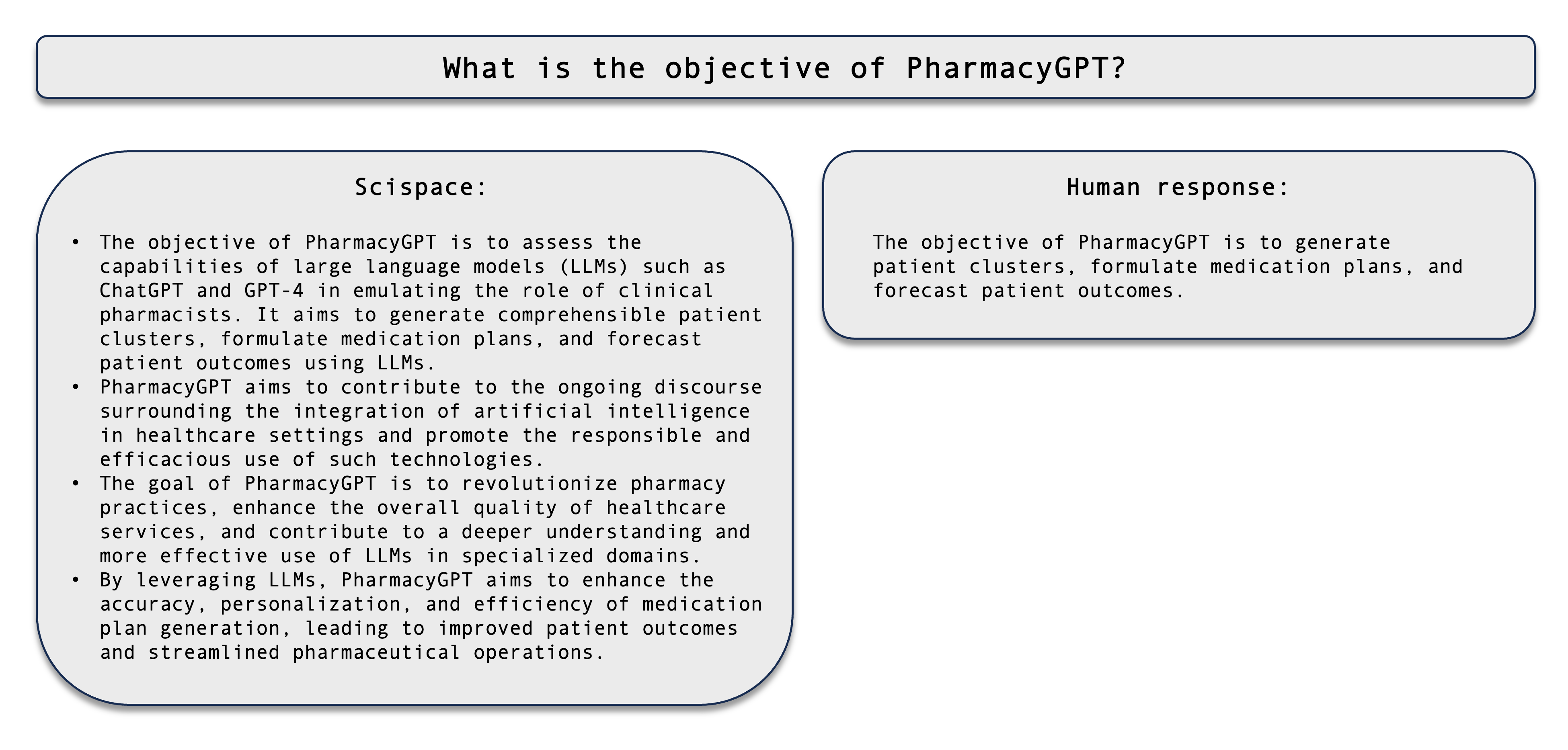}
\end{figure}

\begin{figure}[htbp]
    \centering
    \includegraphics[scale = 0.6]{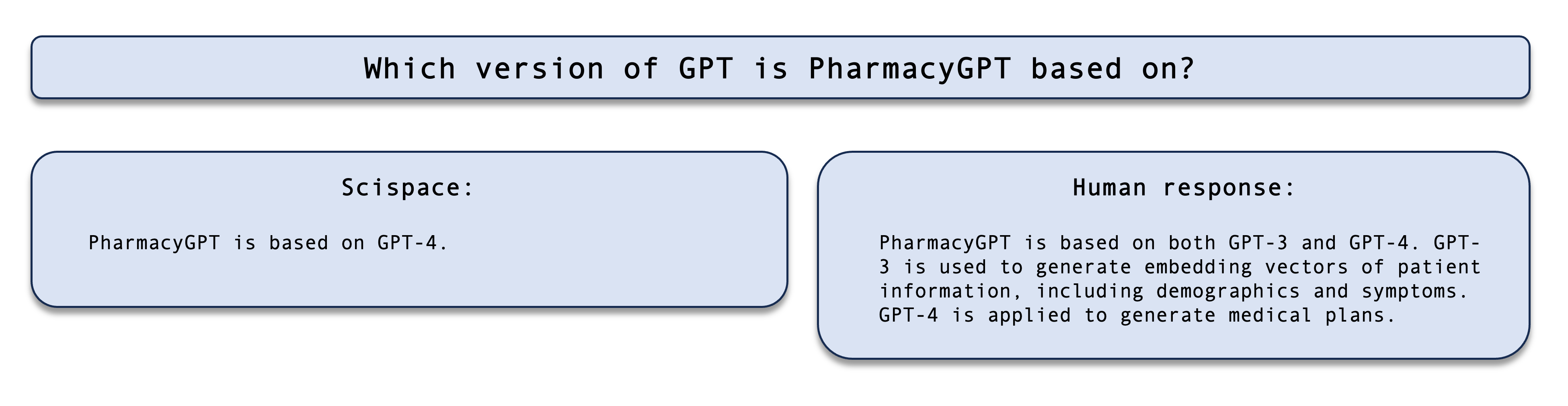}
\end{figure}

\begin{figure}[htbp]
    \centering
    \includegraphics[scale = 0.6]{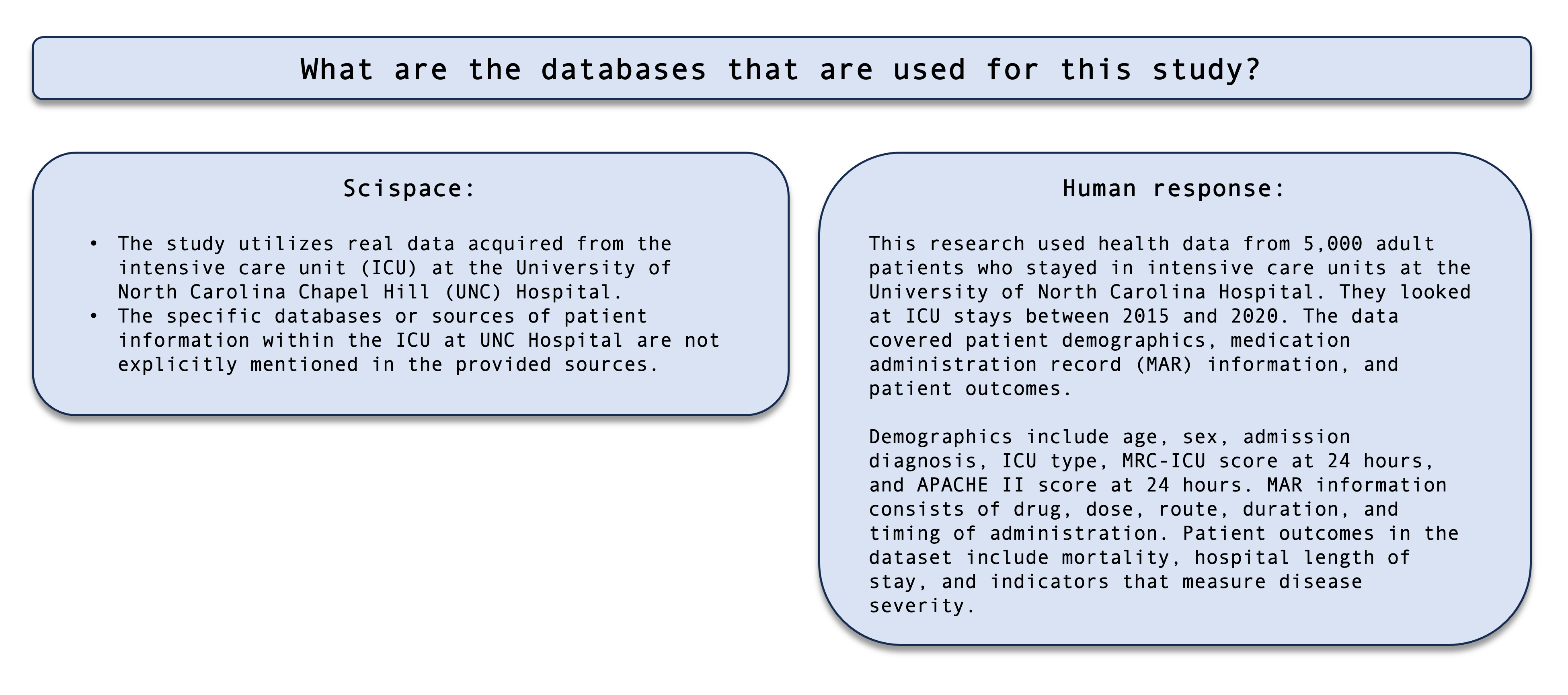}
\end{figure}

\begin{figure}[htbp]
    \centering
    \includegraphics[scale = 0.6]{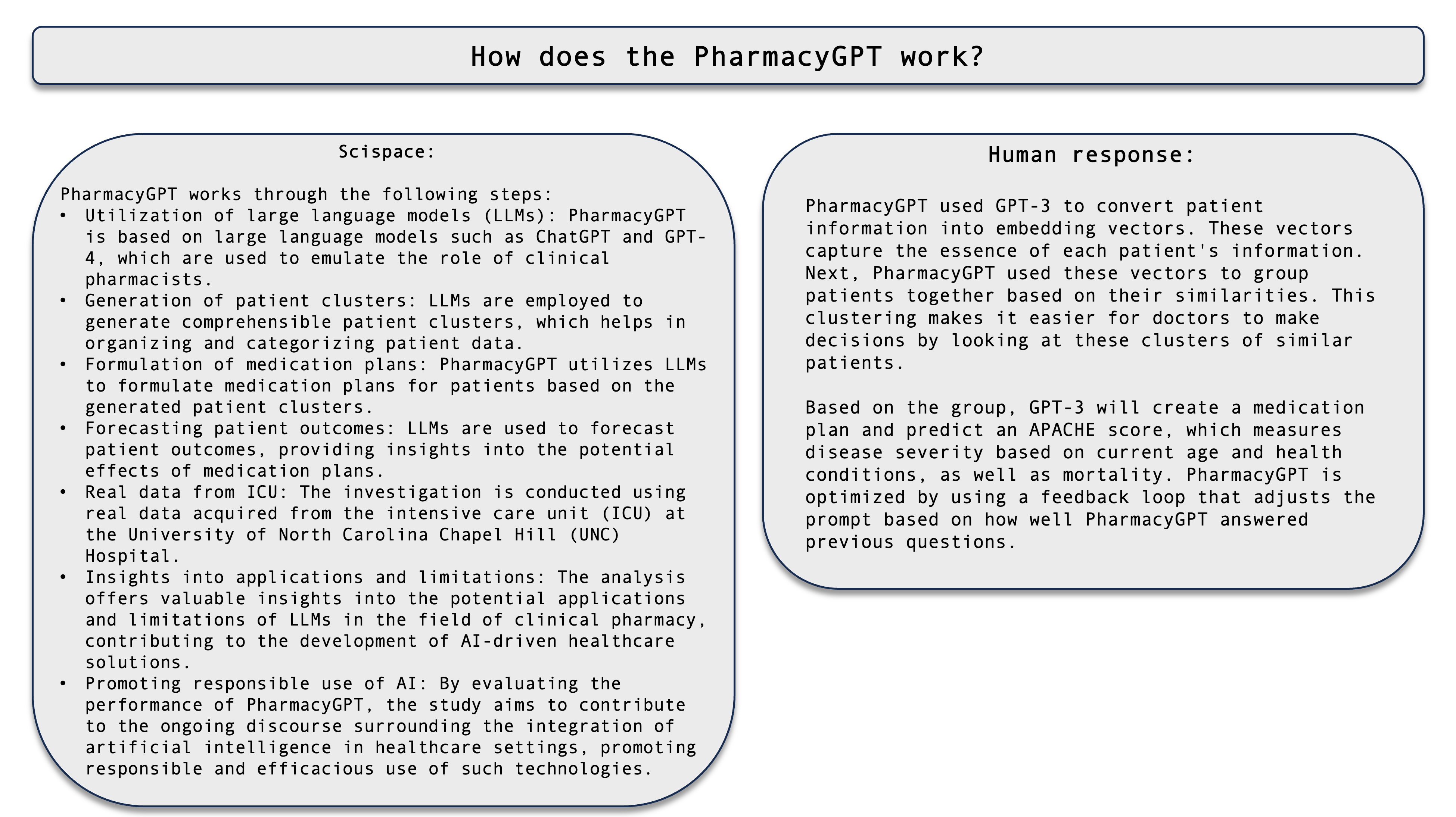}
\end{figure}

\begin{figure}[htbp]
    \centering
    \includegraphics[scale = 0.6]{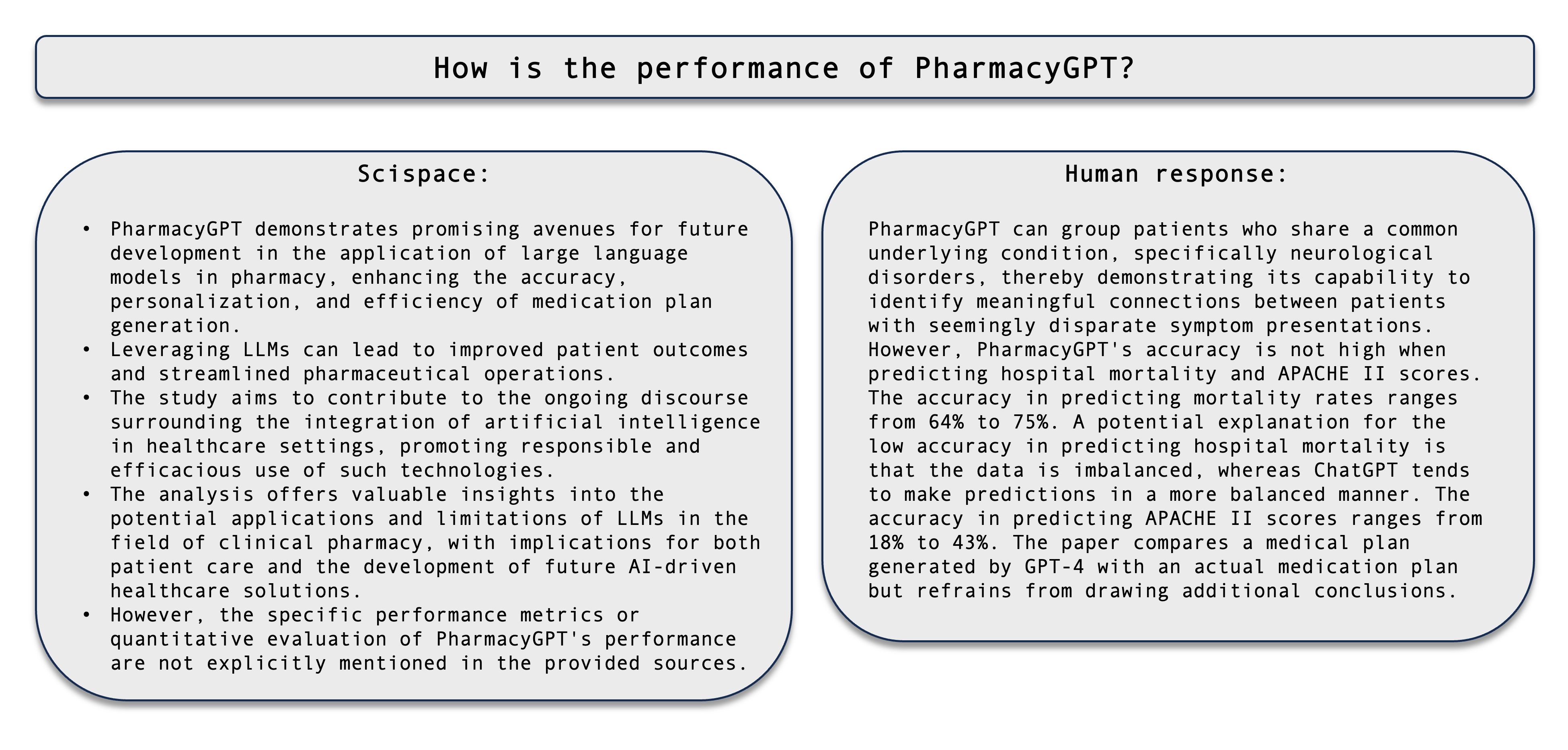}
\end{figure}

\clearpage
ChatGPT-powered Conversational Drug Editing Using Retrieval and Domain Feedback \cite{chatdrug}

\begin{figure}[htbp]
    \centering
    \includegraphics[scale = 0.6]{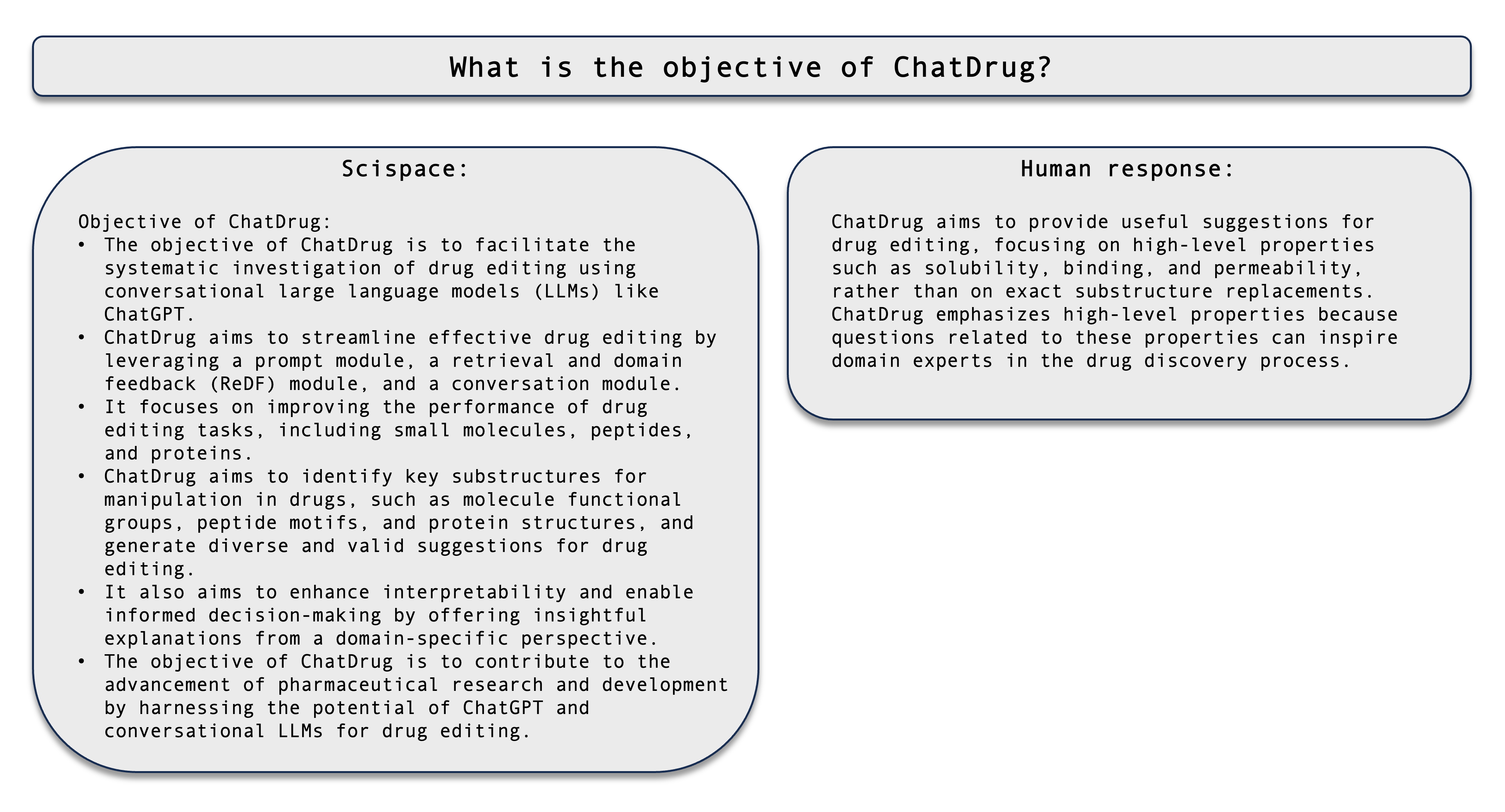}
\end{figure}

\begin{figure}[htbp]
    \centering
    \includegraphics[scale = 0.6]{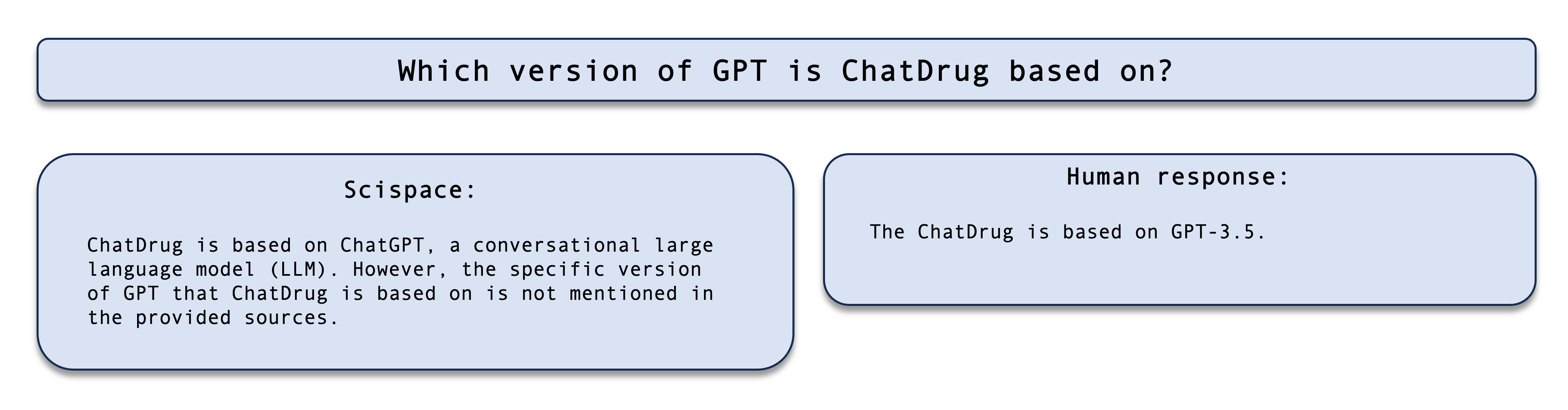}
\end{figure}

\begin{figure}[htbp]
    \centering
    \includegraphics[scale = 0.6]{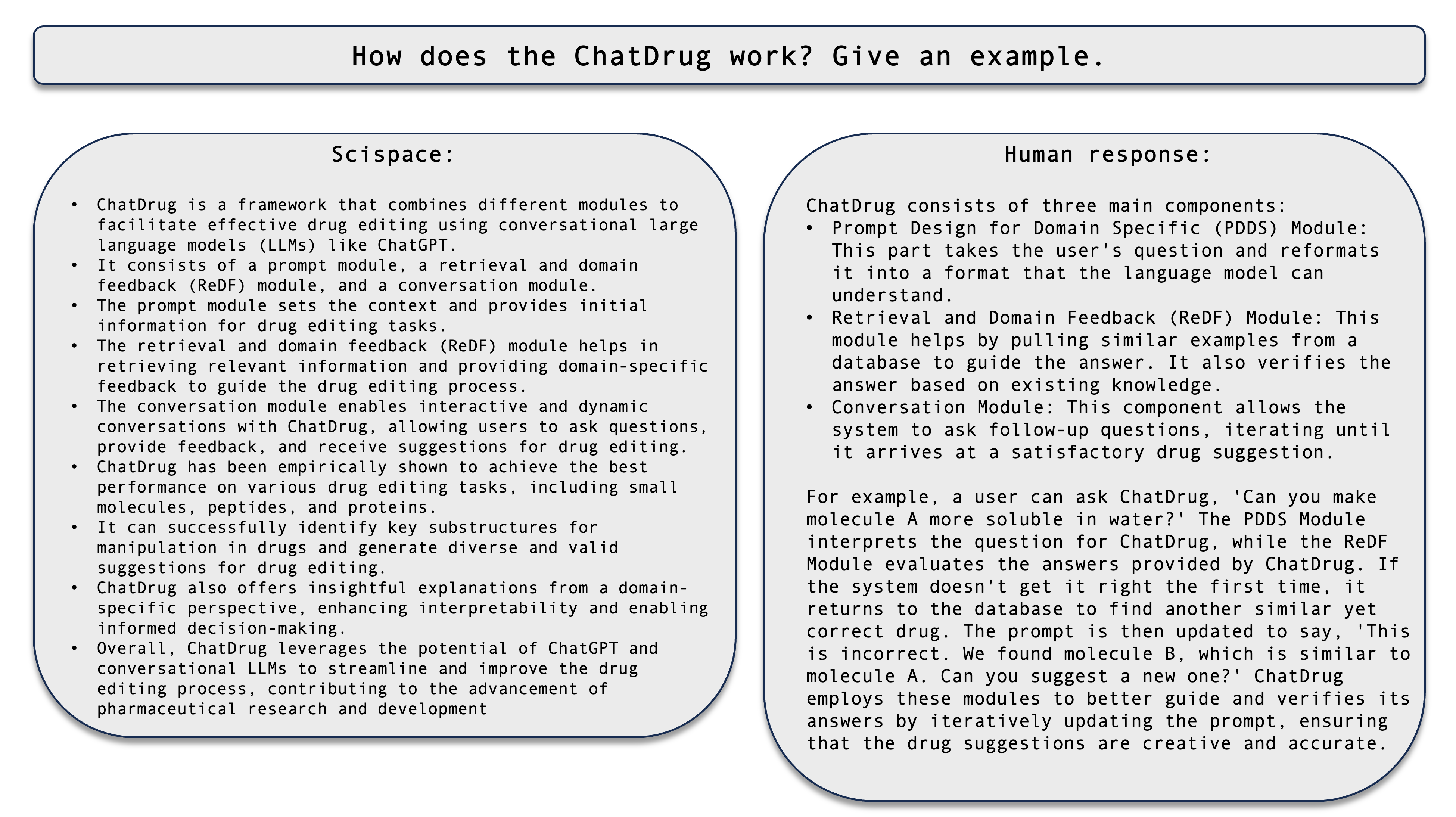}
\end{figure}

\begin{figure}[htbp]
    \centering
    \includegraphics[scale = 0.6]{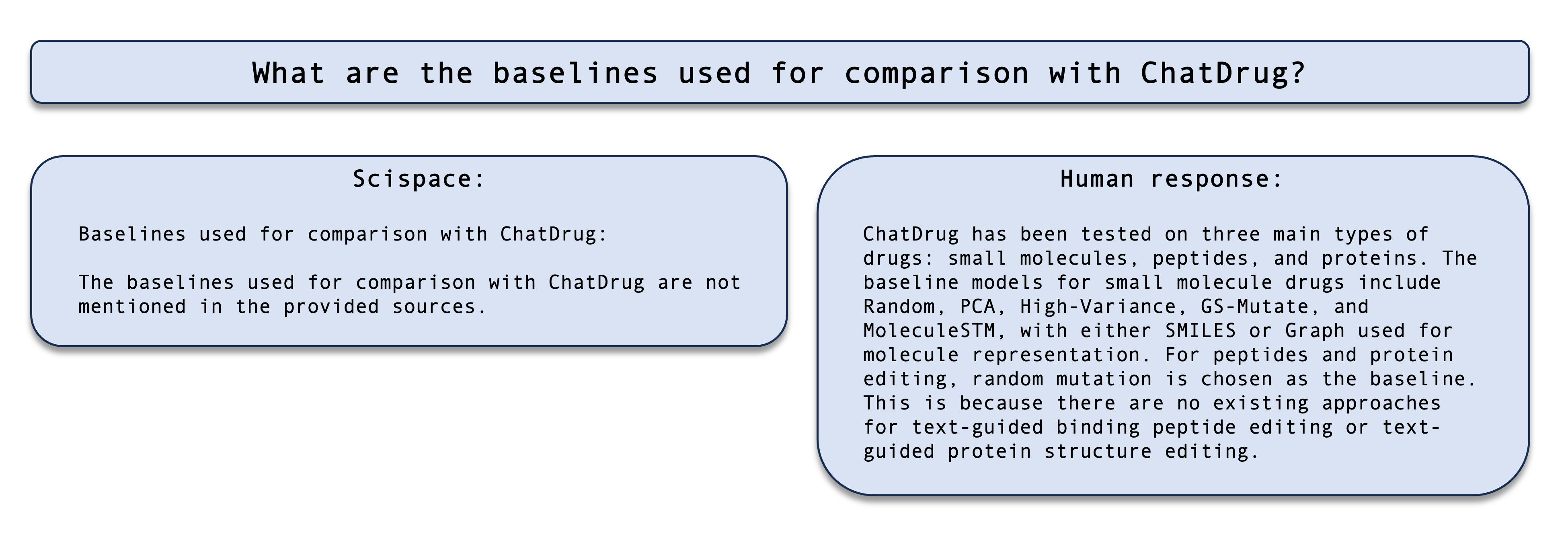}
\end{figure}

\begin{figure}[htbp]
    \centering
    \includegraphics[scale = 0.6]{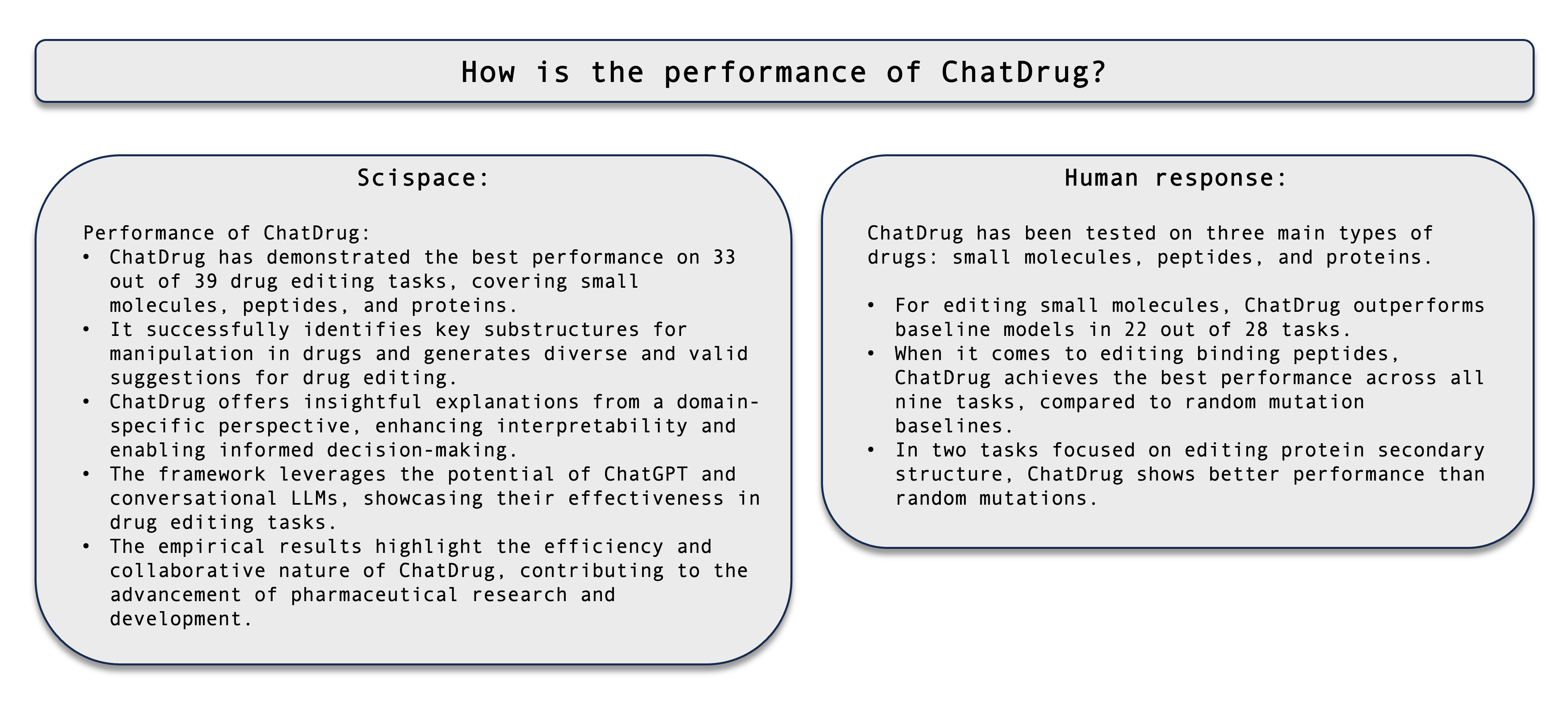}
\end{figure}

\clearpage
Augmenting large language models with chemistry tools \cite{chemcrow}

\begin{figure}[htbp]
    \centering
    \includegraphics[scale = 0.6]{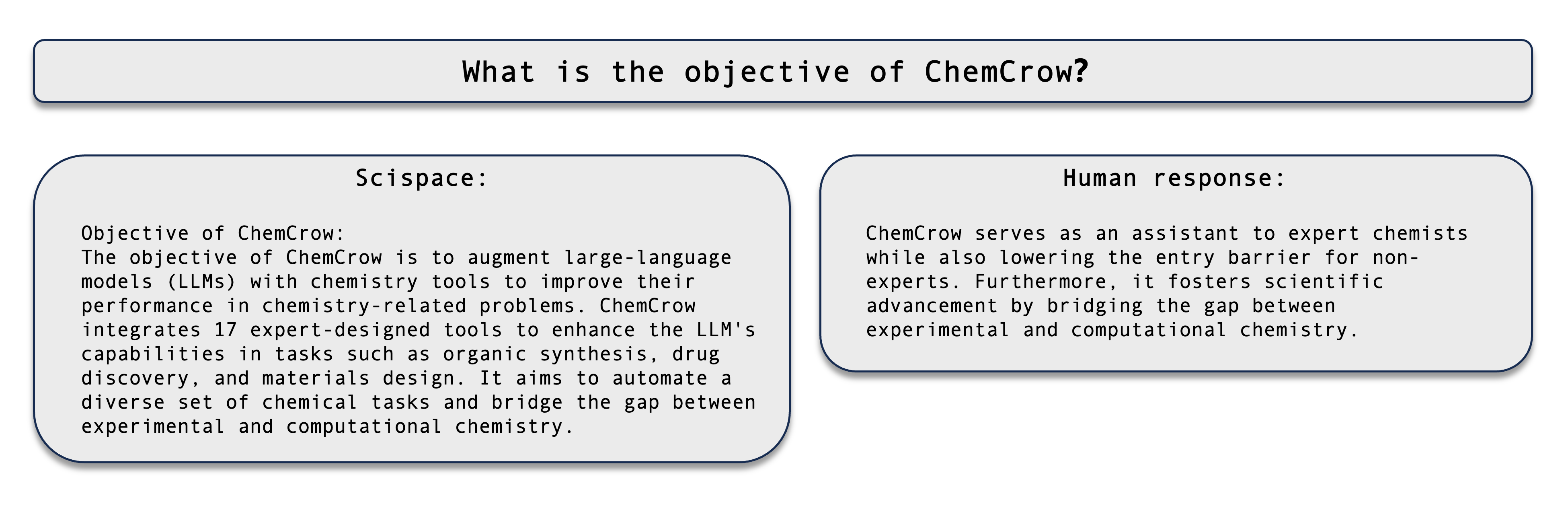}
\end{figure}

\begin{figure}[htbp]
    \centering
    \includegraphics[scale = 0.6]{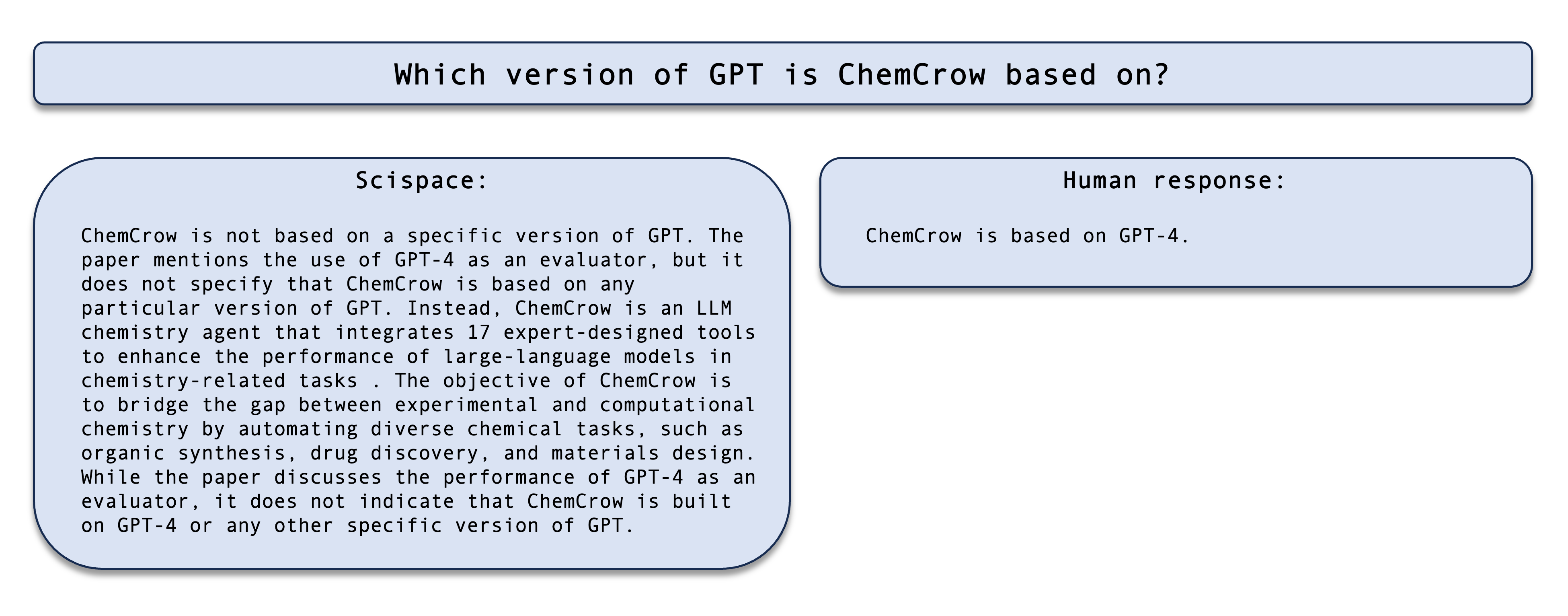}
\end{figure}

\begin{figure}[htbp]
    \centering
    \includegraphics[scale = 0.6]{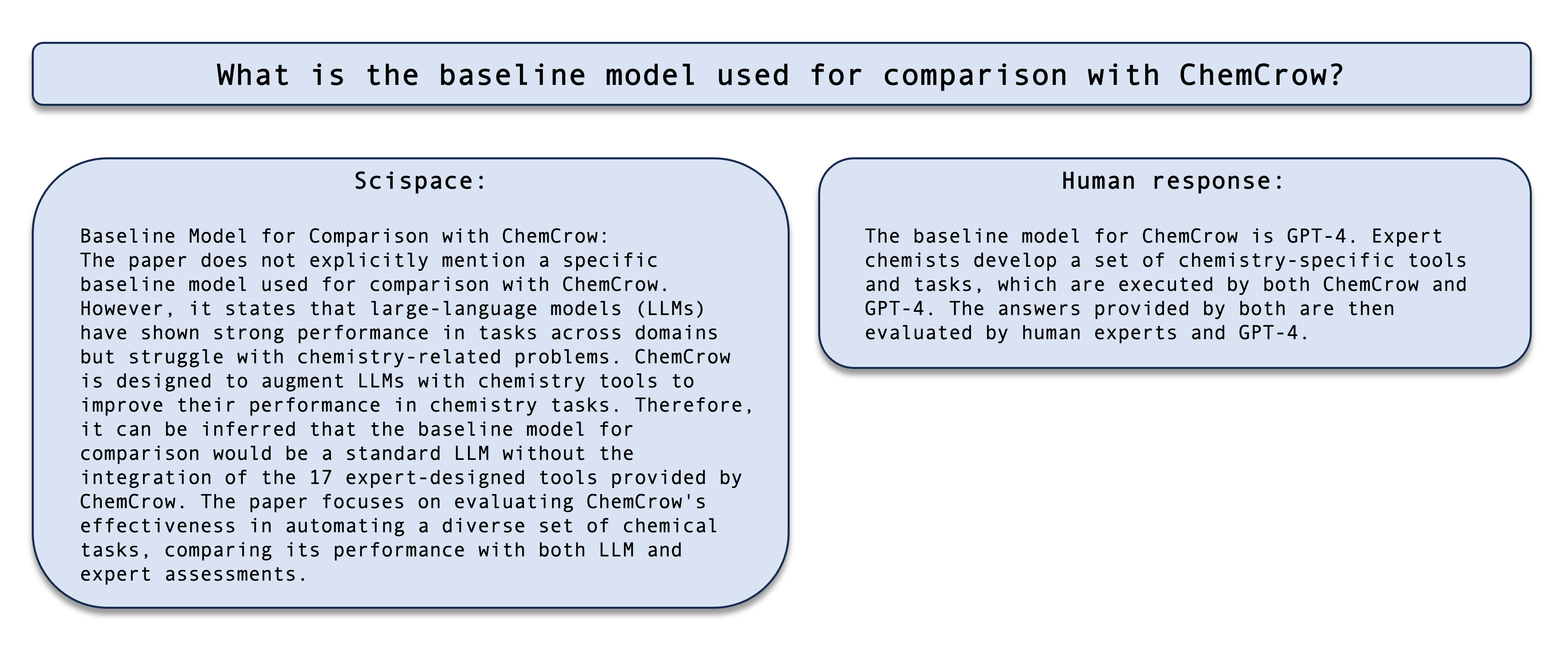}
\end{figure}

\begin{figure}[htbp]
    \centering
    \includegraphics[scale = 0.6]{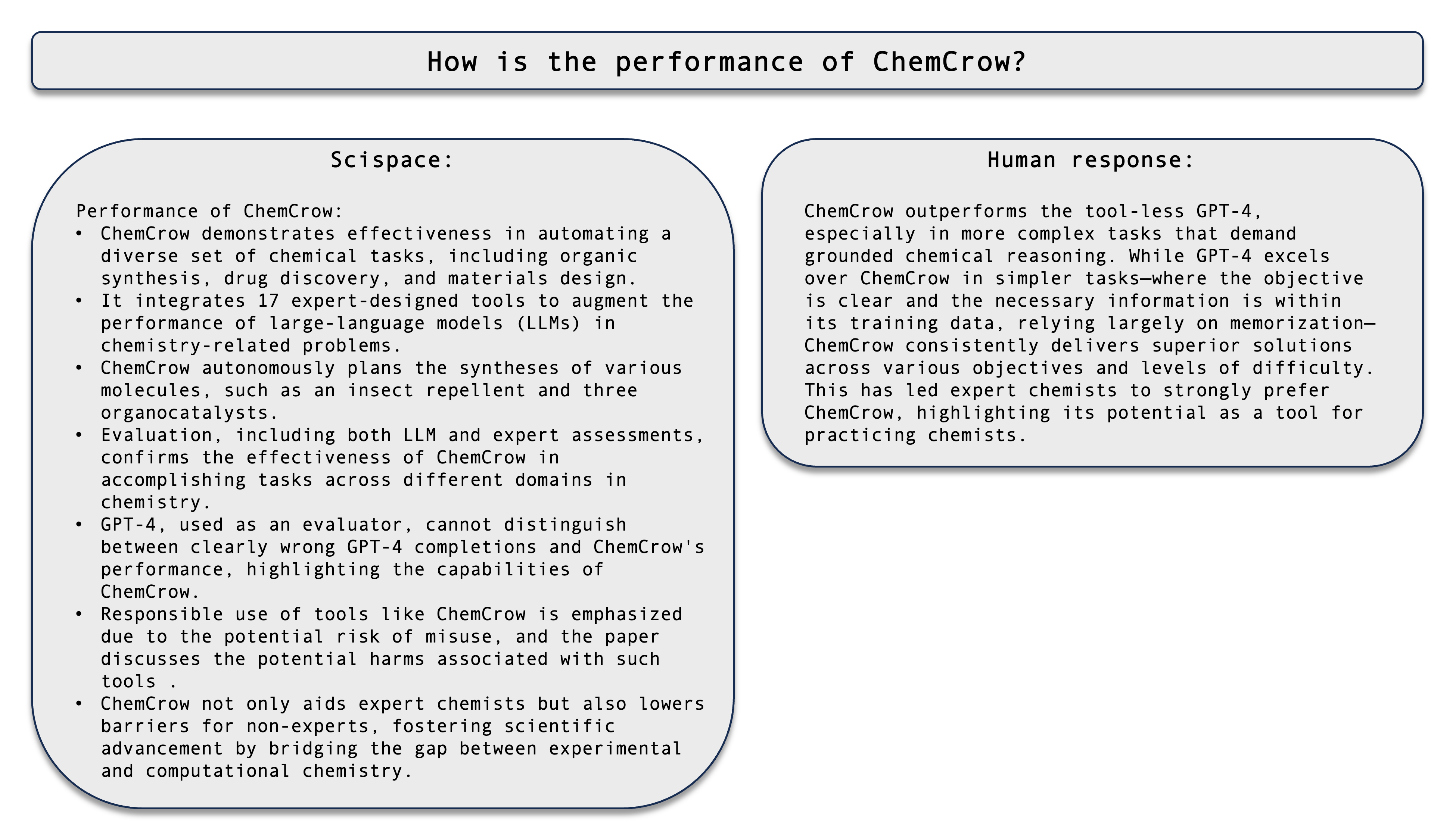}
\end{figure}

\begin{figure}[htbp]
    \centering
    \includegraphics[scale = 0.6]{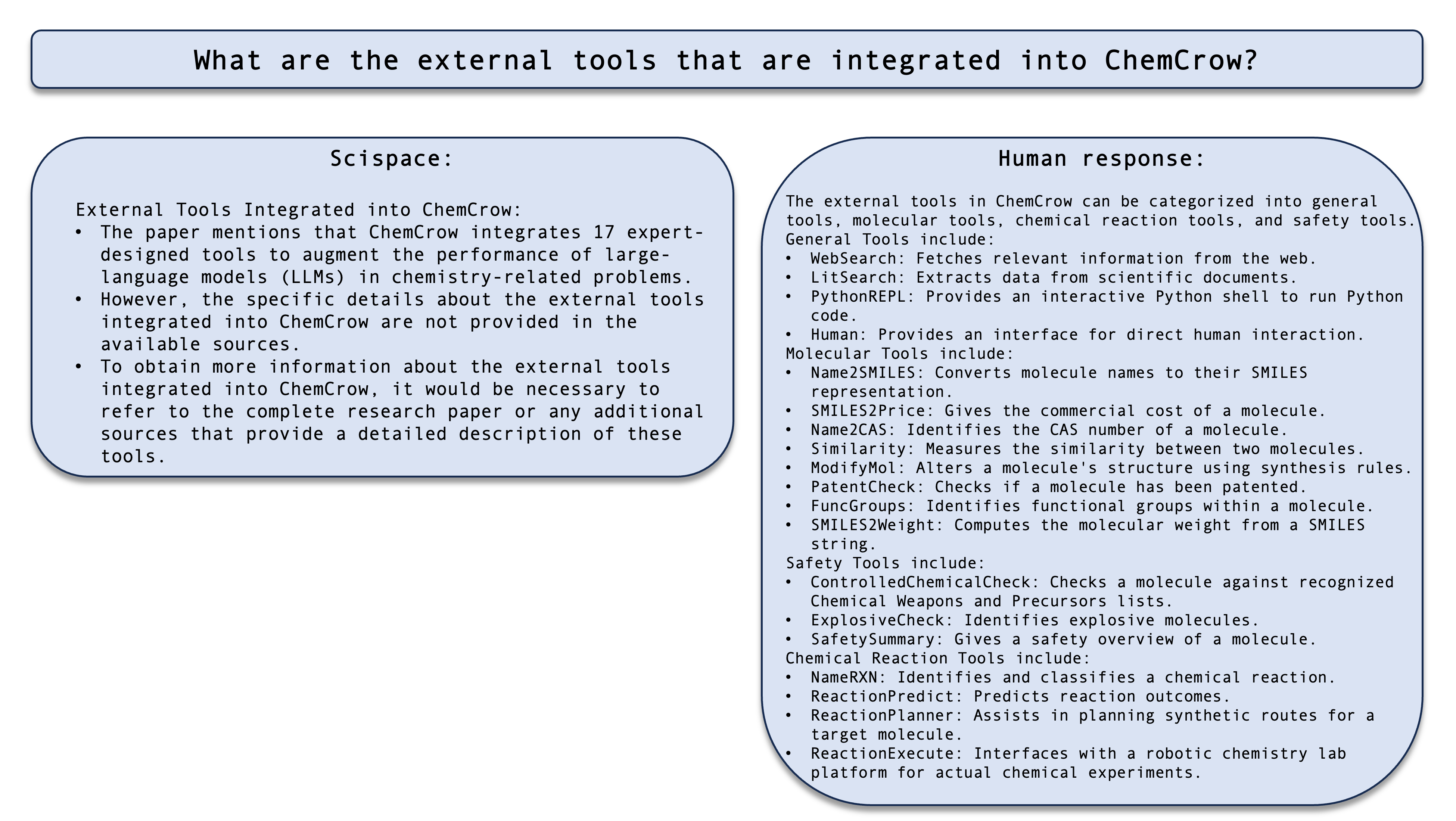}
\end{figure}

\begin{figure}[htbp]
    \centering
    \includegraphics[scale = 0.6]{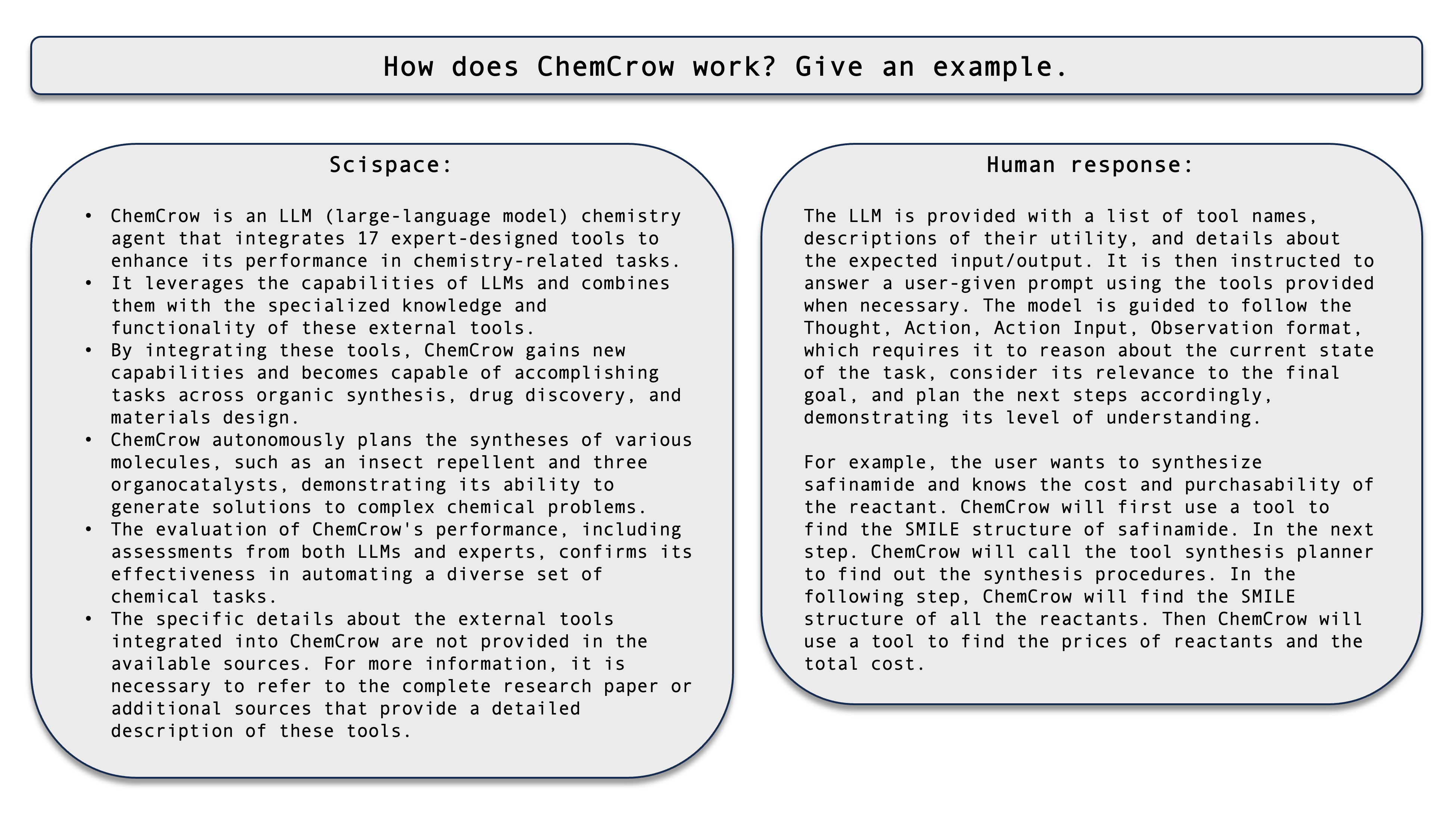}
\end{figure}

\clearpage
AD-AutoGPT: An Autonomous GPT for Alzheimer’s Disease Infodemiology \cite{adautogpt}

\begin{figure}[htbp]
    \centering
    \includegraphics[scale = 0.6]{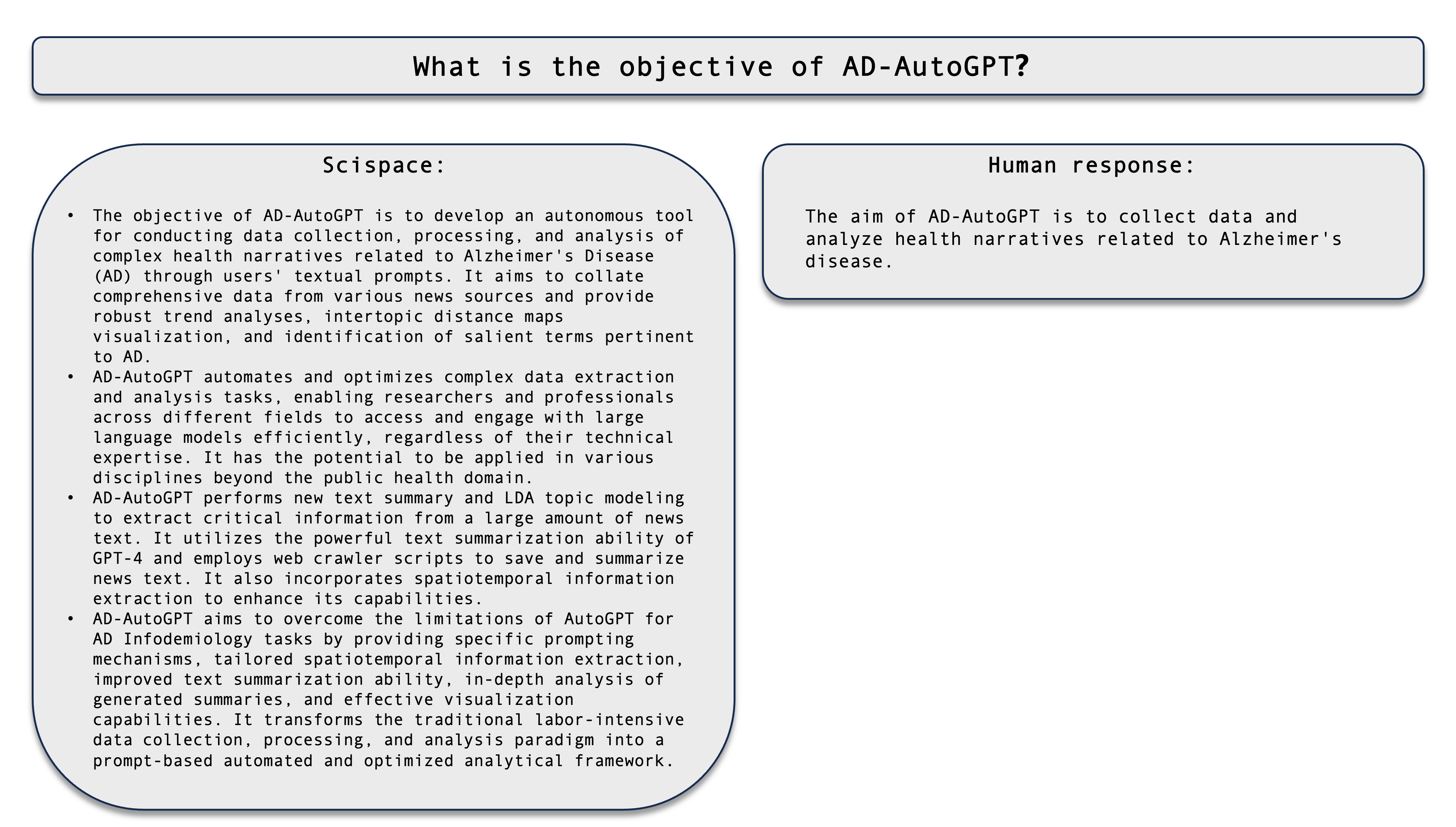}
\end{figure}

\begin{figure}[htbp]
    \centering
    \includegraphics[scale = 0.6]{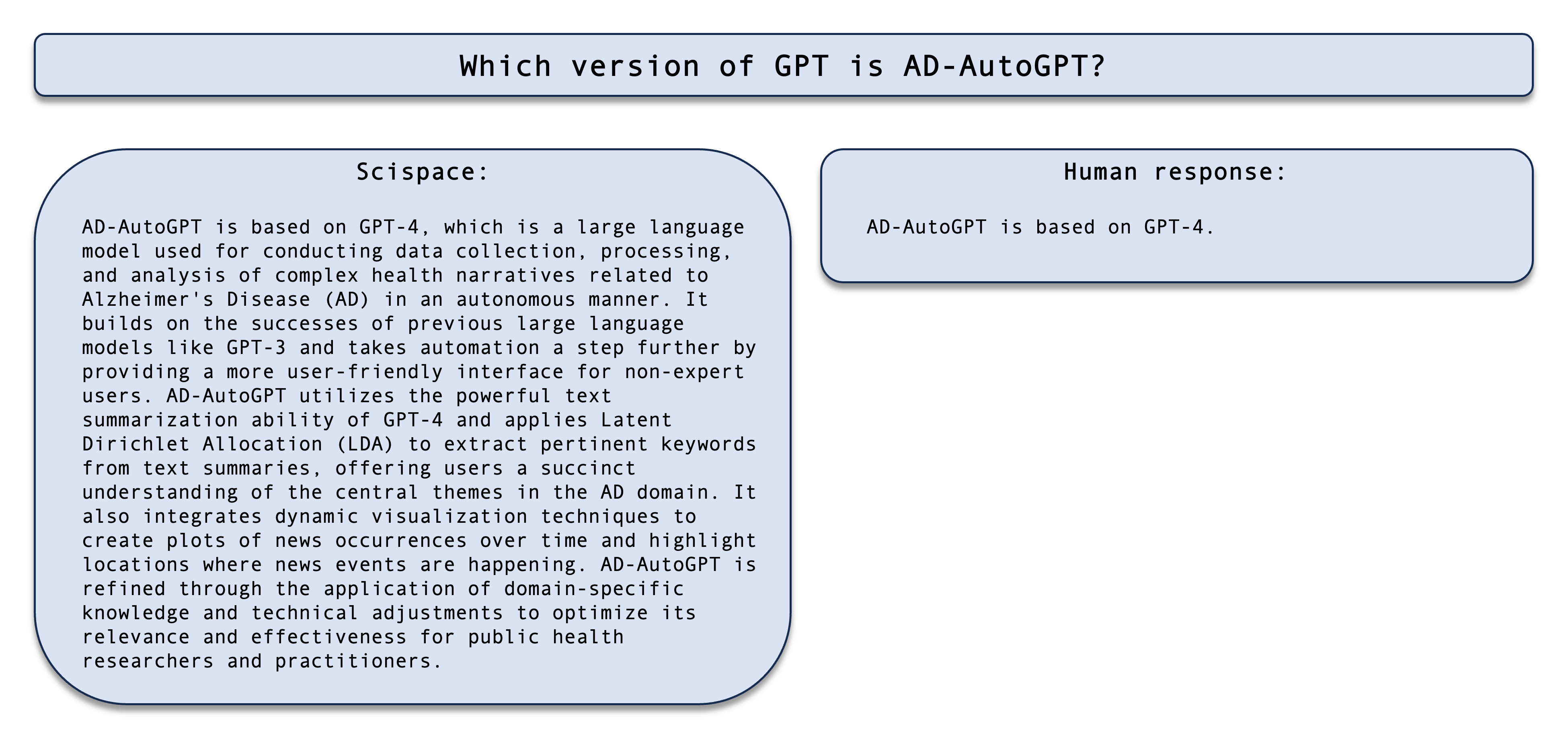}
\end{figure}

\begin{figure}[htbp]
    \centering
    \includegraphics[scale = 0.6]{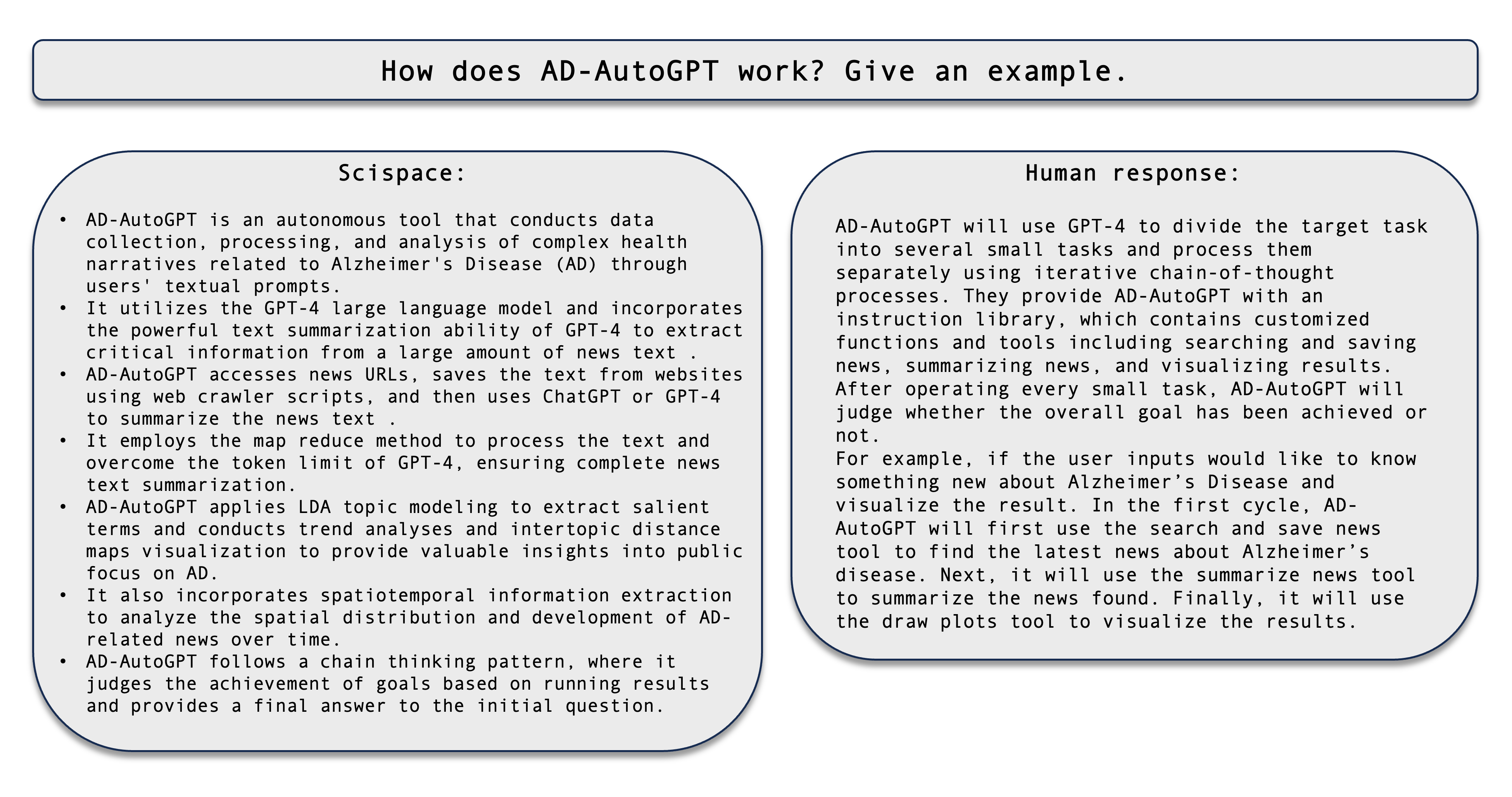}
\end{figure}

\begin{figure}[htbp]
    \centering
    \includegraphics[scale = 0.6]{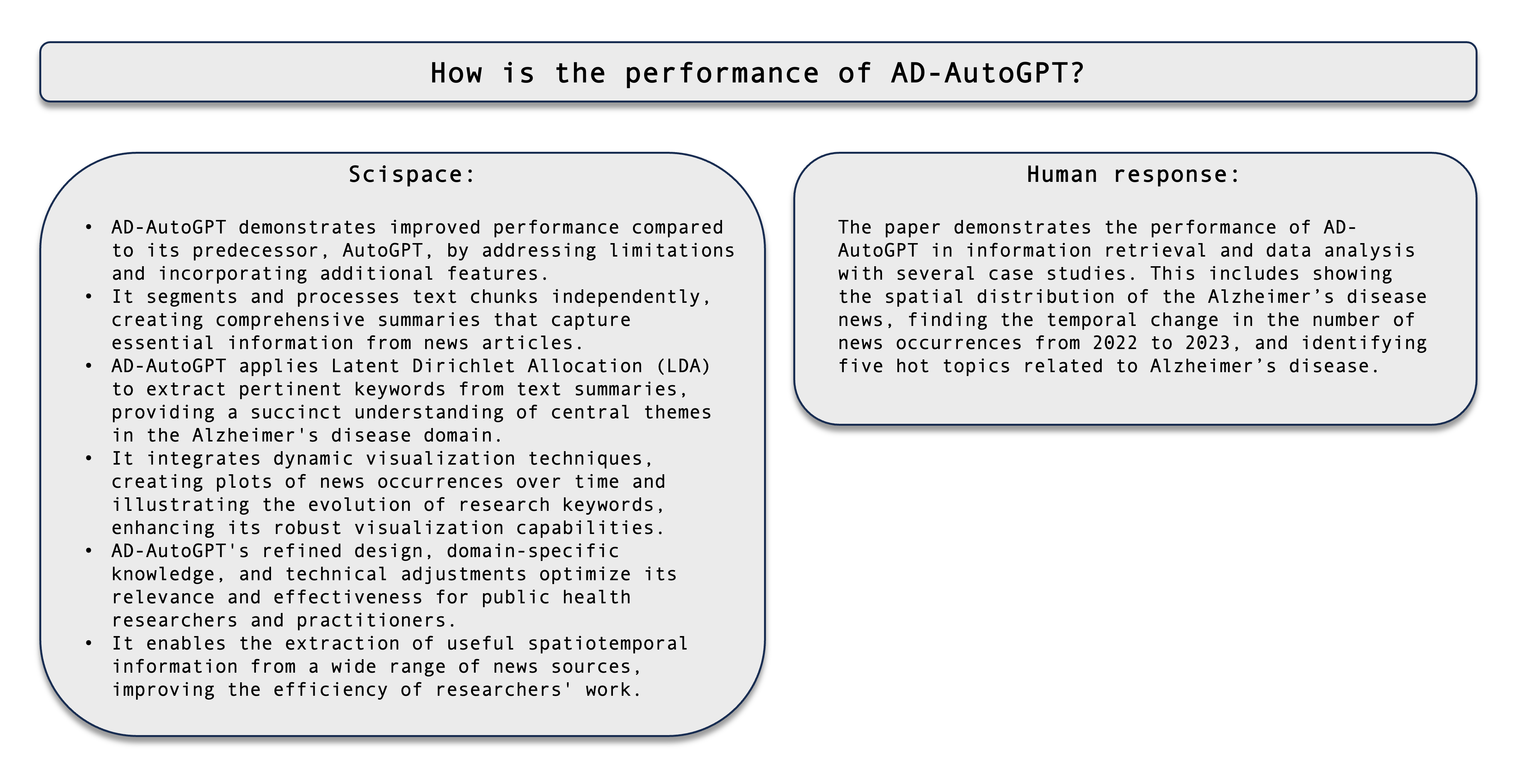}
\end{figure}

\begin{figure}[htbp]
    \centering
    \includegraphics[scale = 0.6]{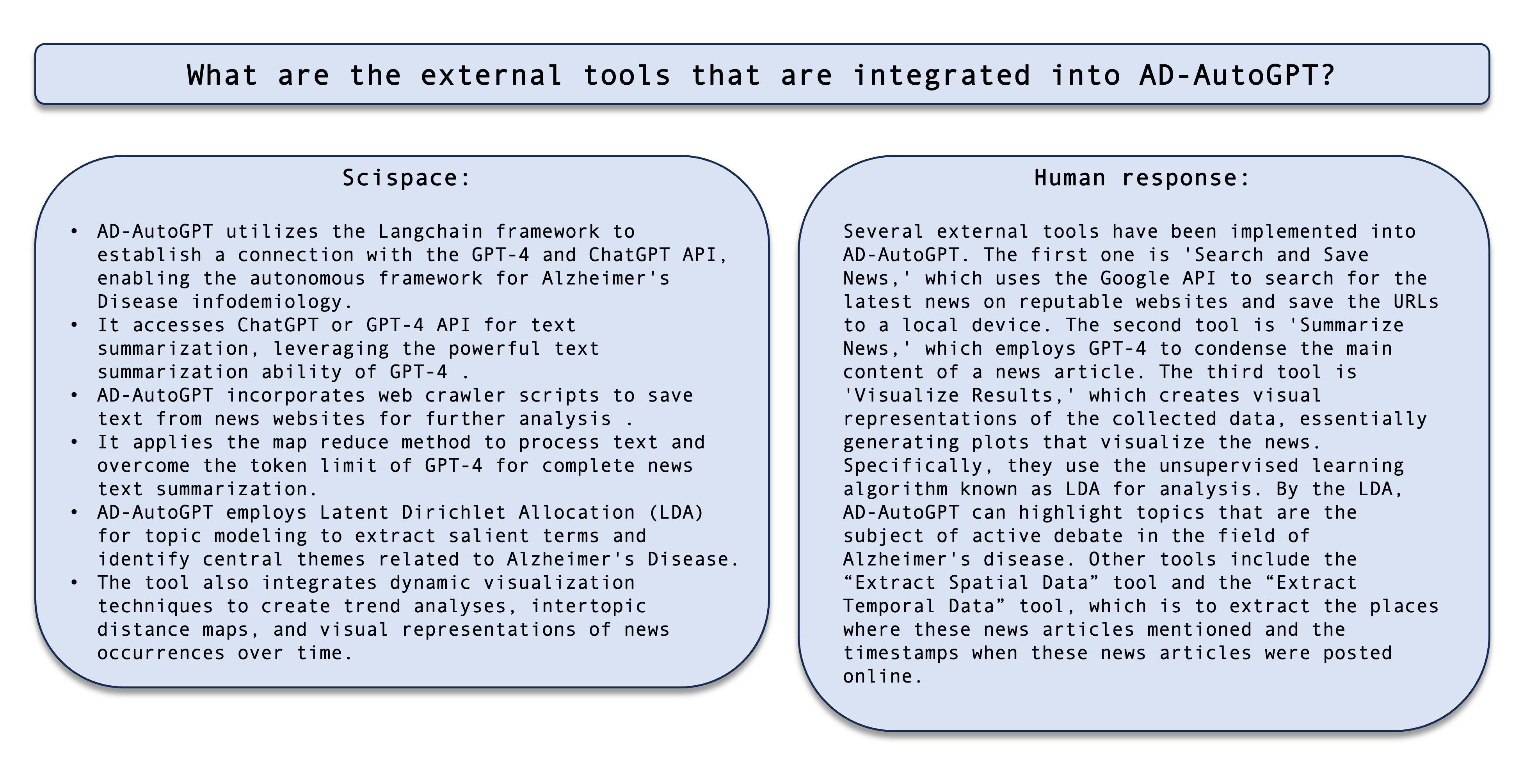}
\end{figure}

\clearpage
GeneGPT: Augmenting Large Language Models with Domain Tools for Improved Access to Biomedical Information \cite{genegpt}

\begin{figure}[htbp]
    \centering
    \includegraphics[scale = 0.6]{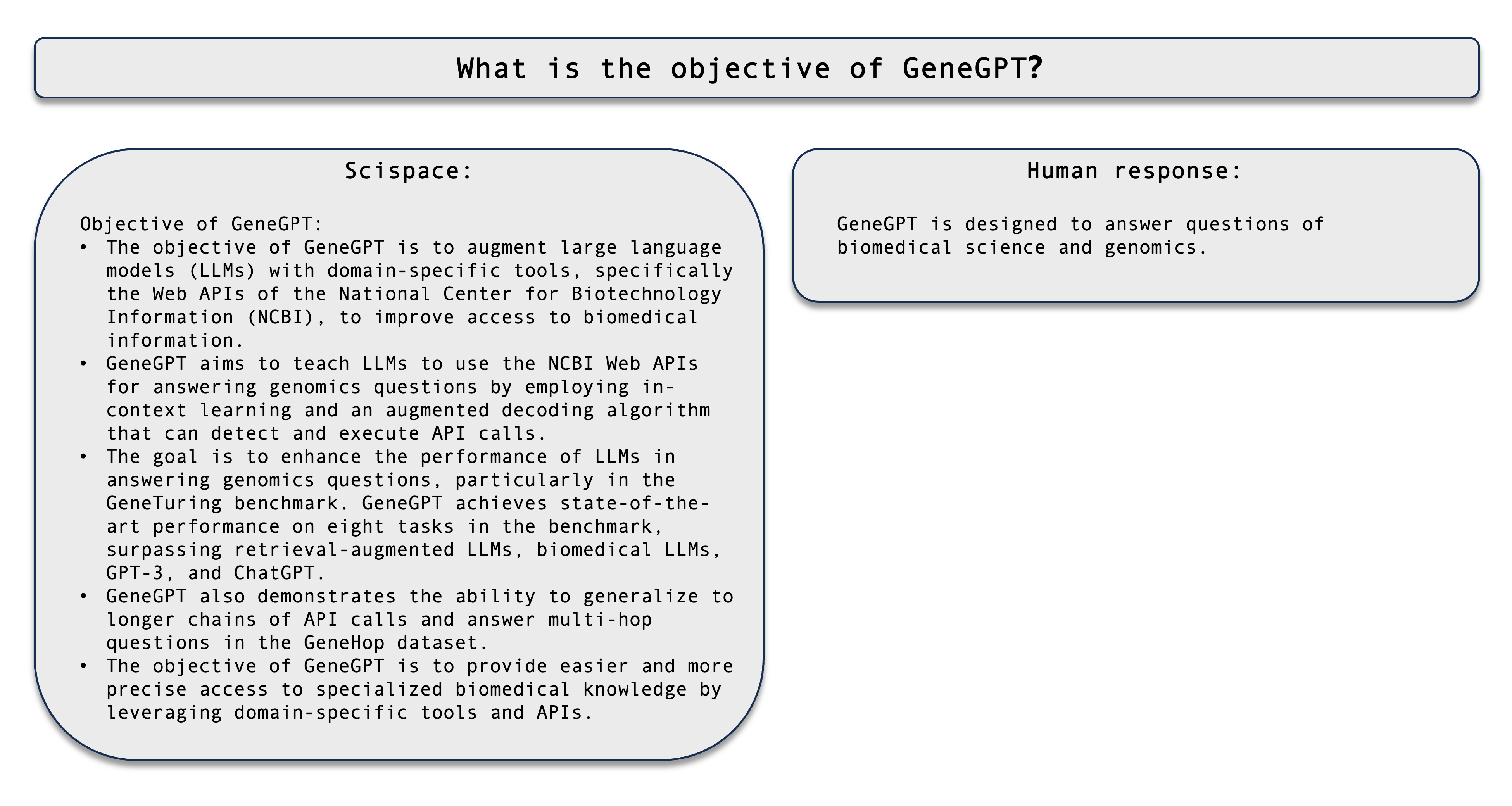}
\end{figure}

\begin{figure}[htbp]
    \centering
    \includegraphics[scale = 0.6]{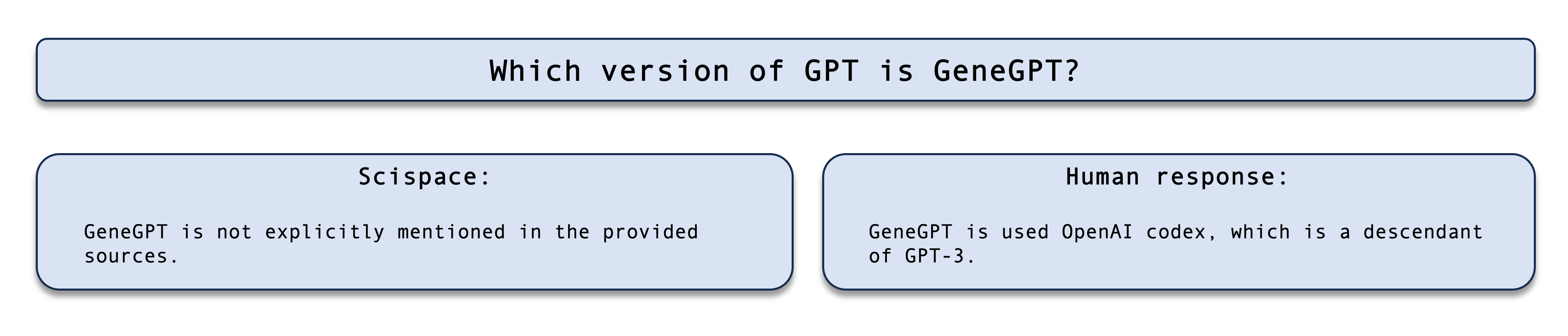}
\end{figure}

\begin{figure}[htbp]
    \centering
    \includegraphics[scale = 0.6]{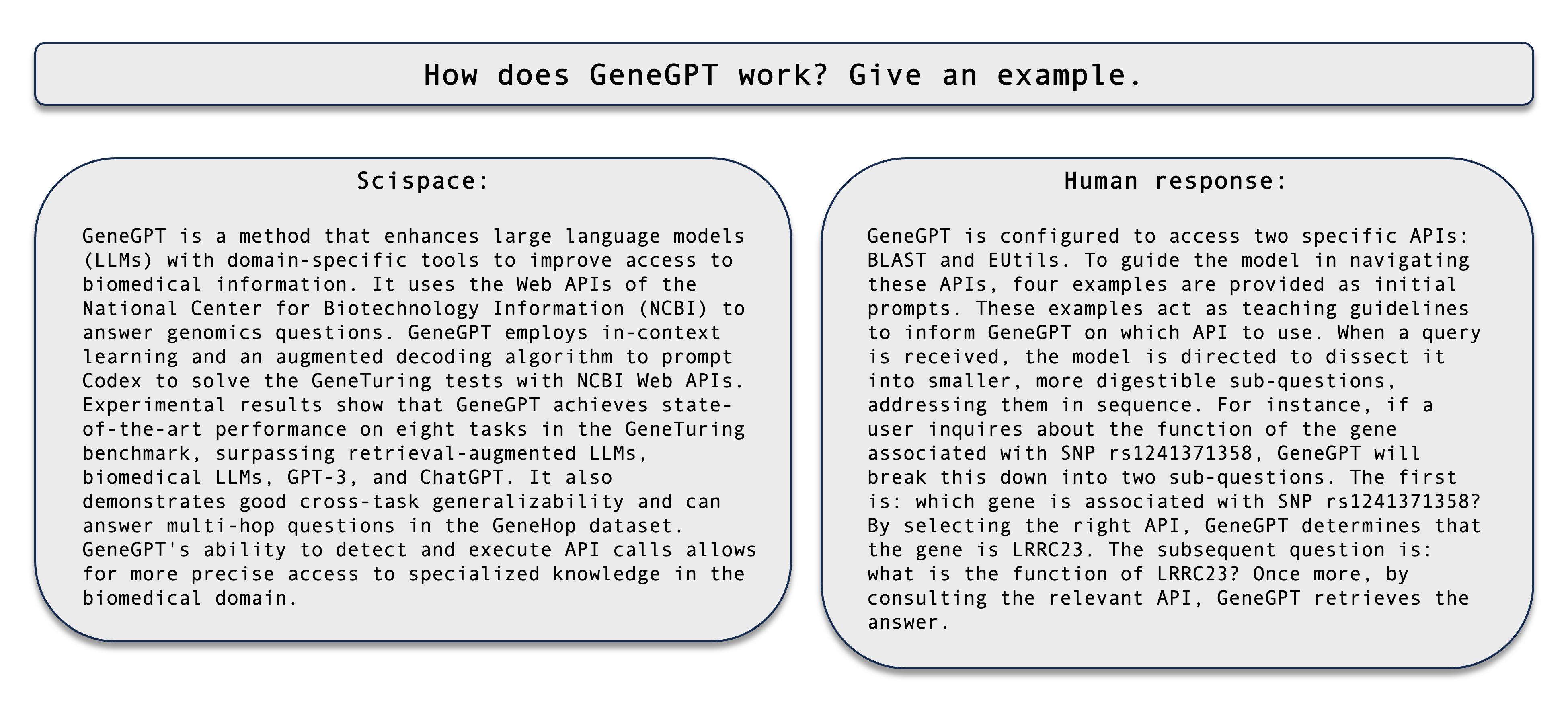}
\end{figure}

\begin{figure}[htbp]
    \centering
    \includegraphics[scale = 0.6]{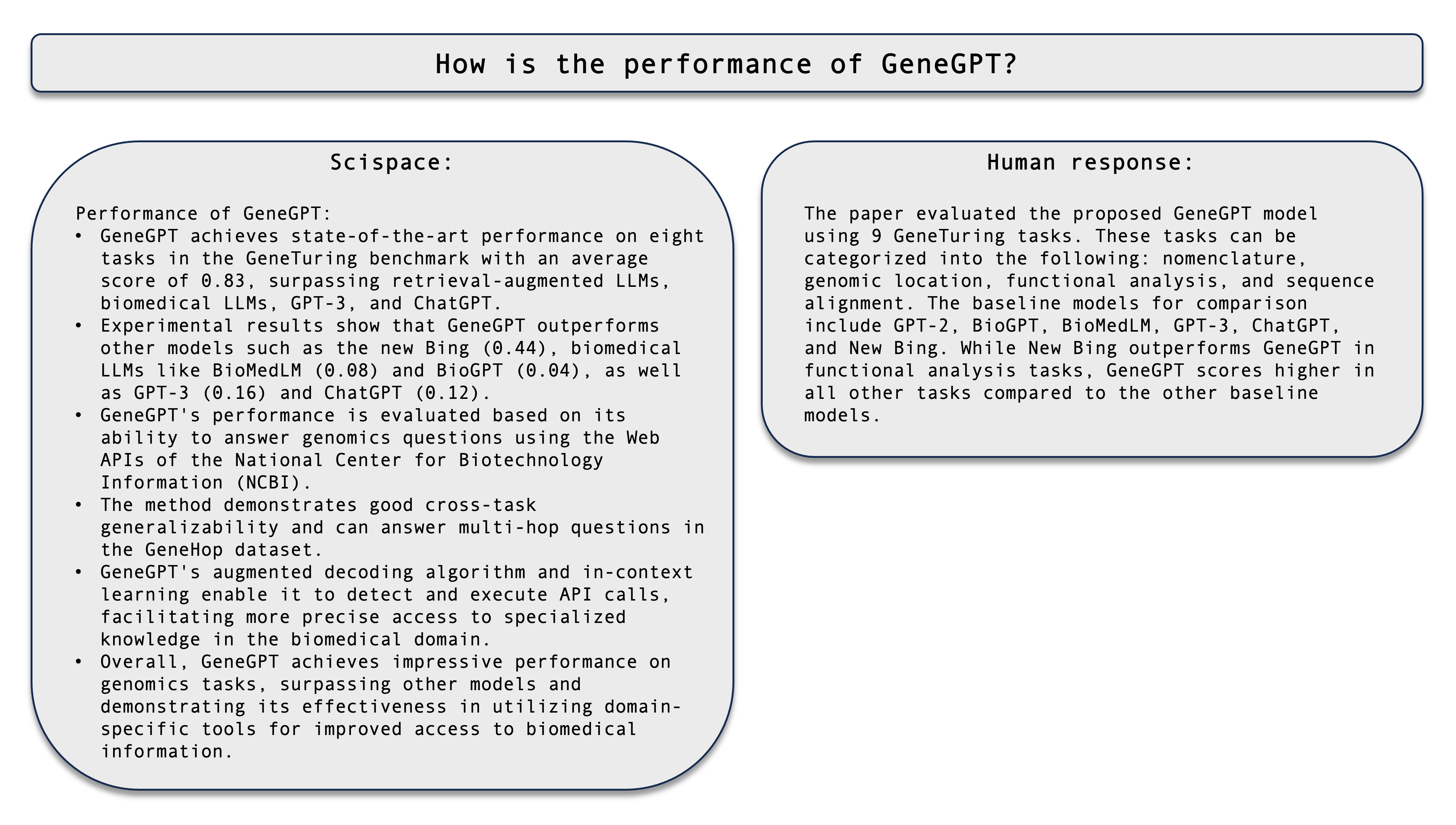}
\end{figure}

\begin{figure}[htbp]
    \centering
    \includegraphics[scale = 0.6]{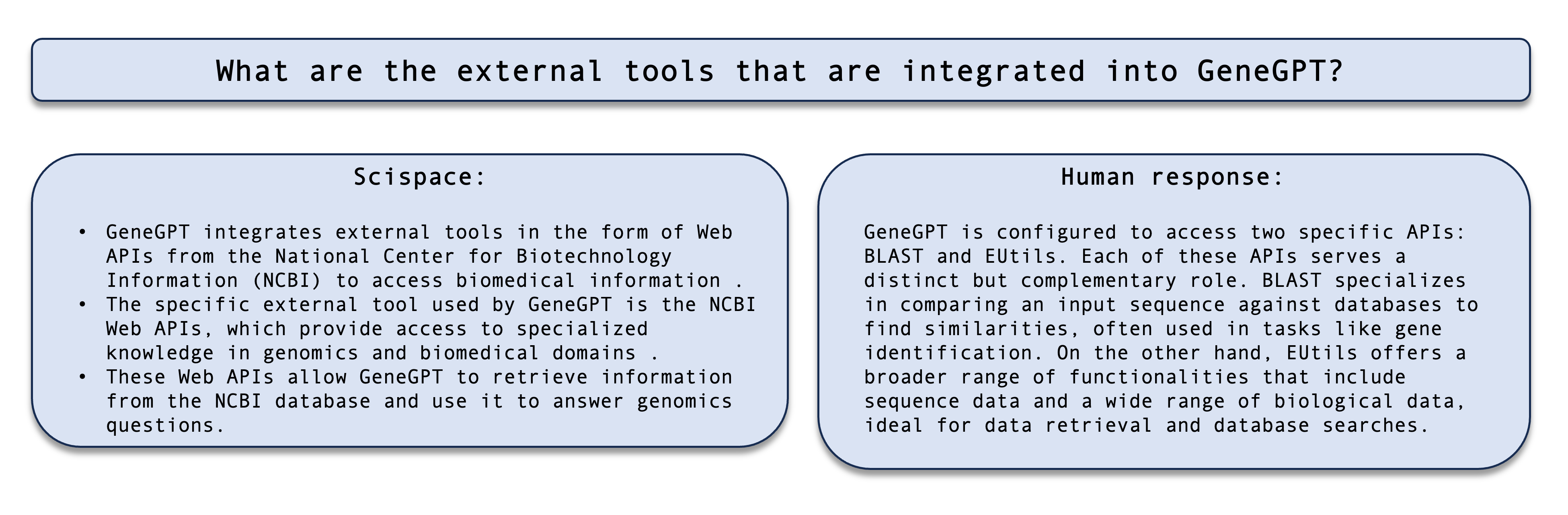}
\end{figure}

\clearpage
CohortGPT: An Enhanced GPT for Participant Recruitment in Clinical Study \cite{cohortgpt}

\begin{figure}[htbp]
    \centering
    \includegraphics[scale = 0.6]{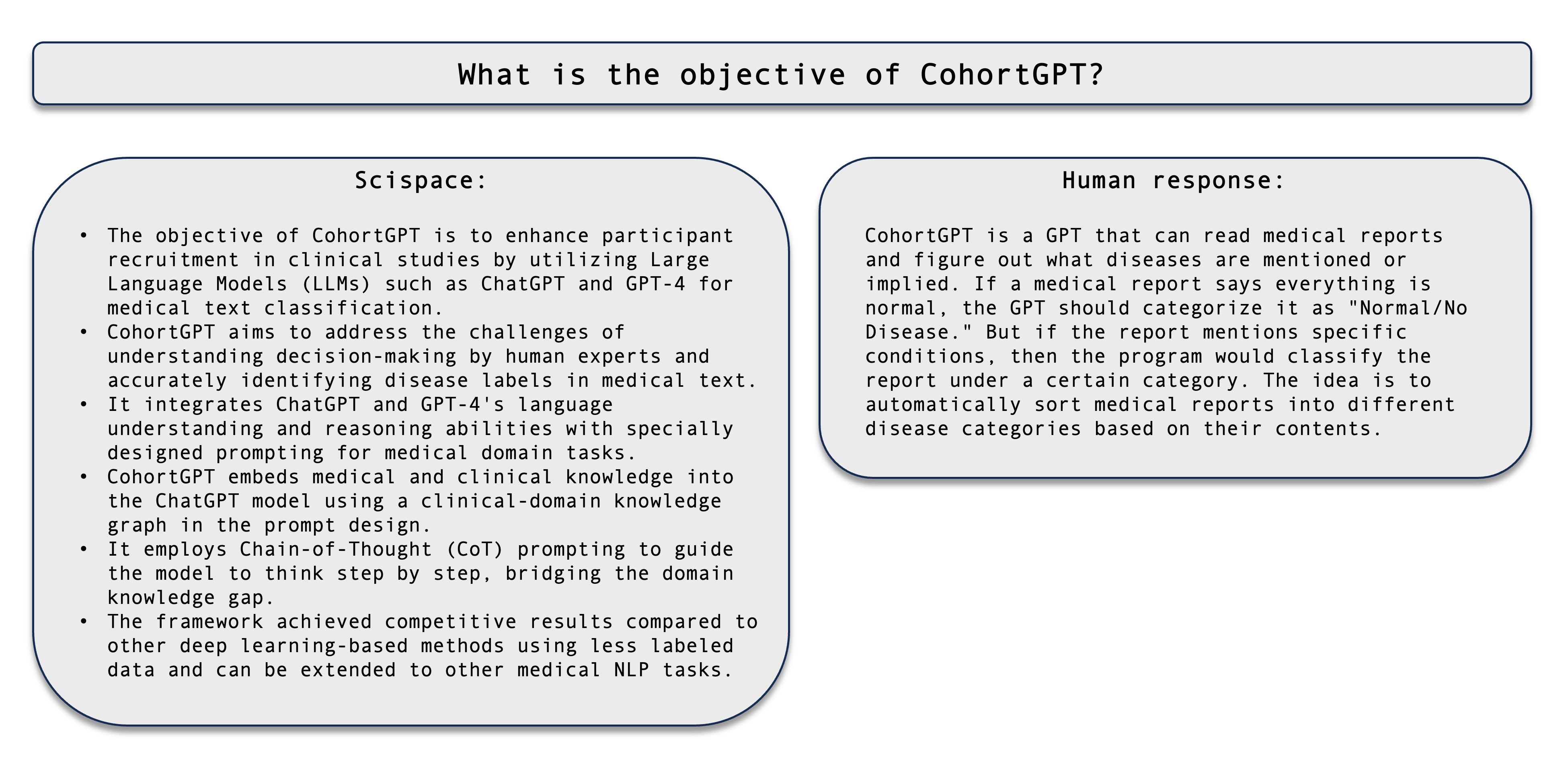}
\end{figure}

\begin{figure}[htbp]
    \centering
    \includegraphics[scale = 0.6]{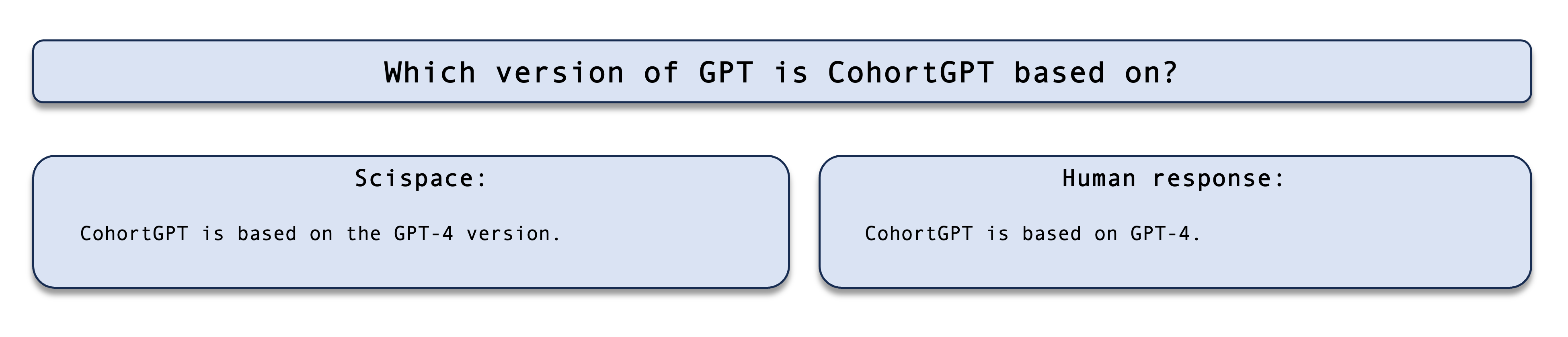}
\end{figure}

\begin{figure}[htbp]
    \centering
    \includegraphics[scale = 0.6]{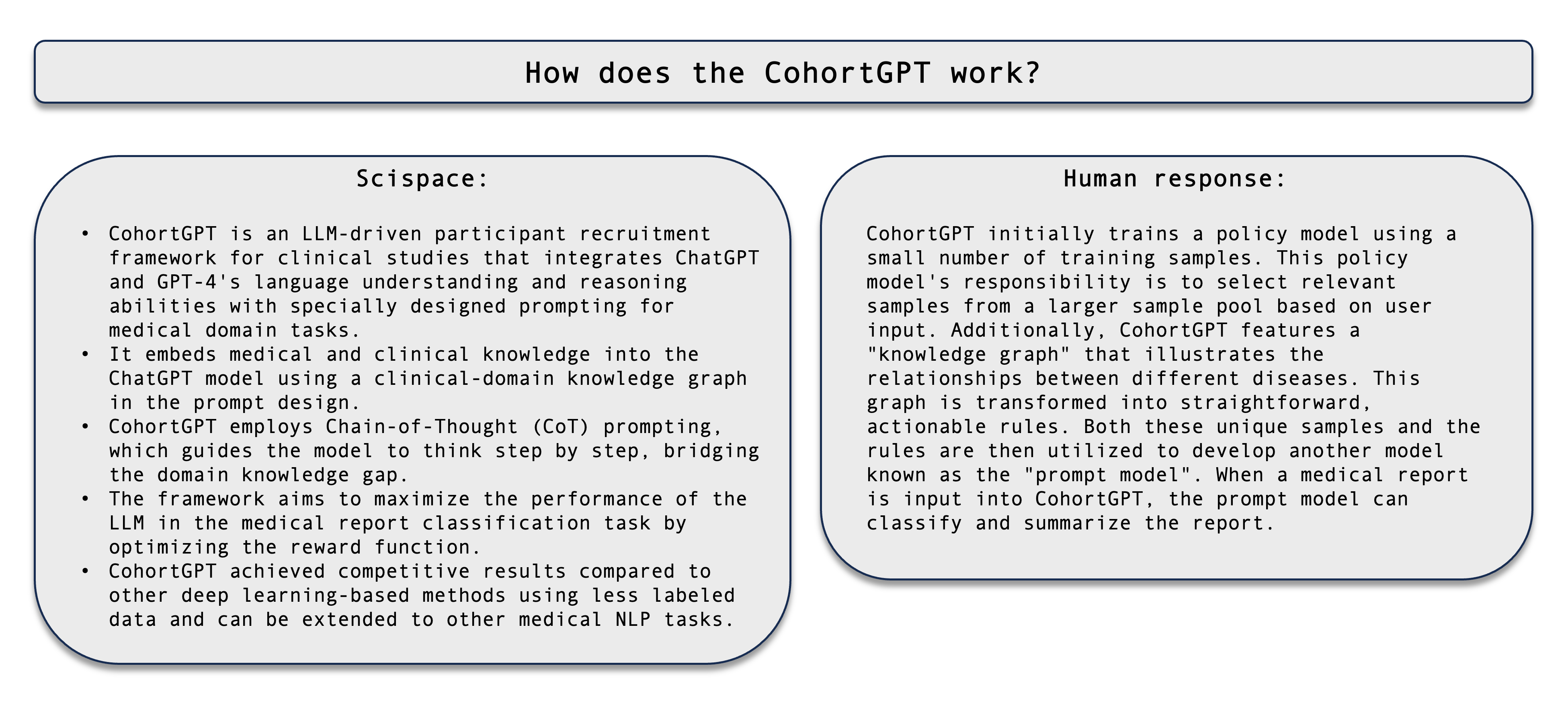}
\end{figure}

\begin{figure}[htbp]
    \centering
    \includegraphics[scale = 0.6]{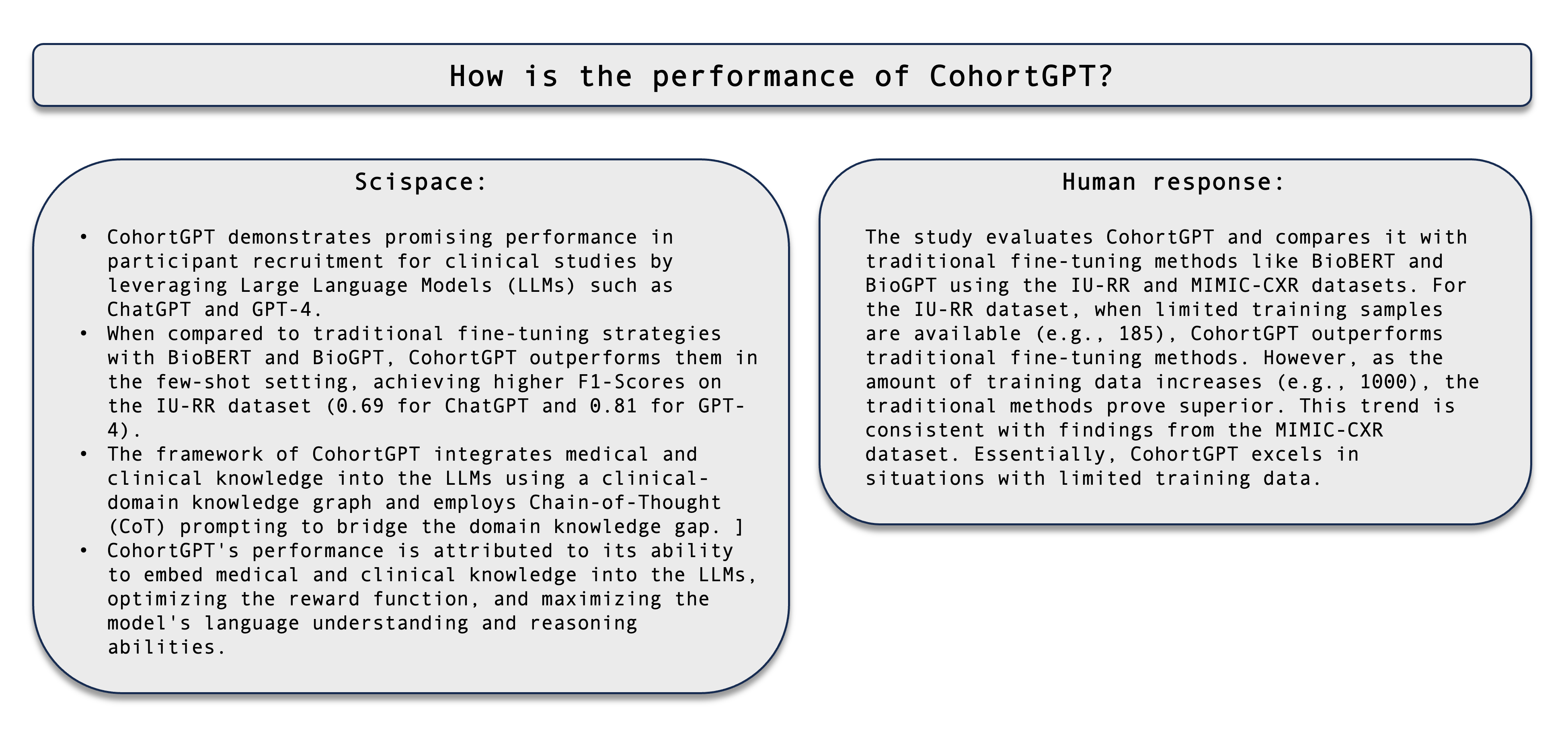}
\end{figure}

\begin{figure}[htbp]
    \centering
    \includegraphics[scale = 0.6]{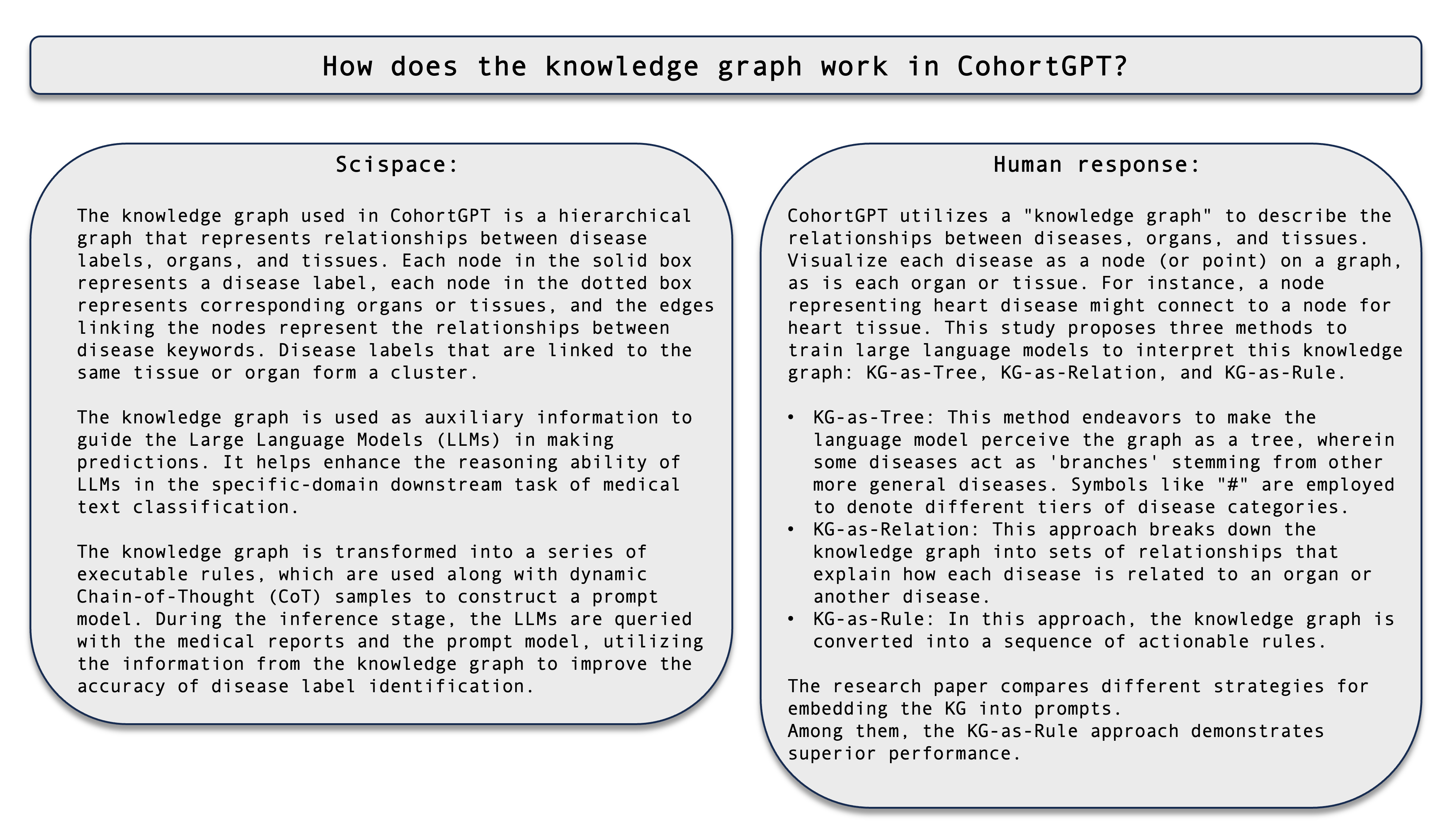}
\end{figure}

\clearpage
\textbf{Machine Learning:}\\

AutoML-GPT: Automatic Machine Learning with GPT \cite{automlgpt}

\begin{figure}[htbp]
    \centering
    \includegraphics[scale = 0.6]{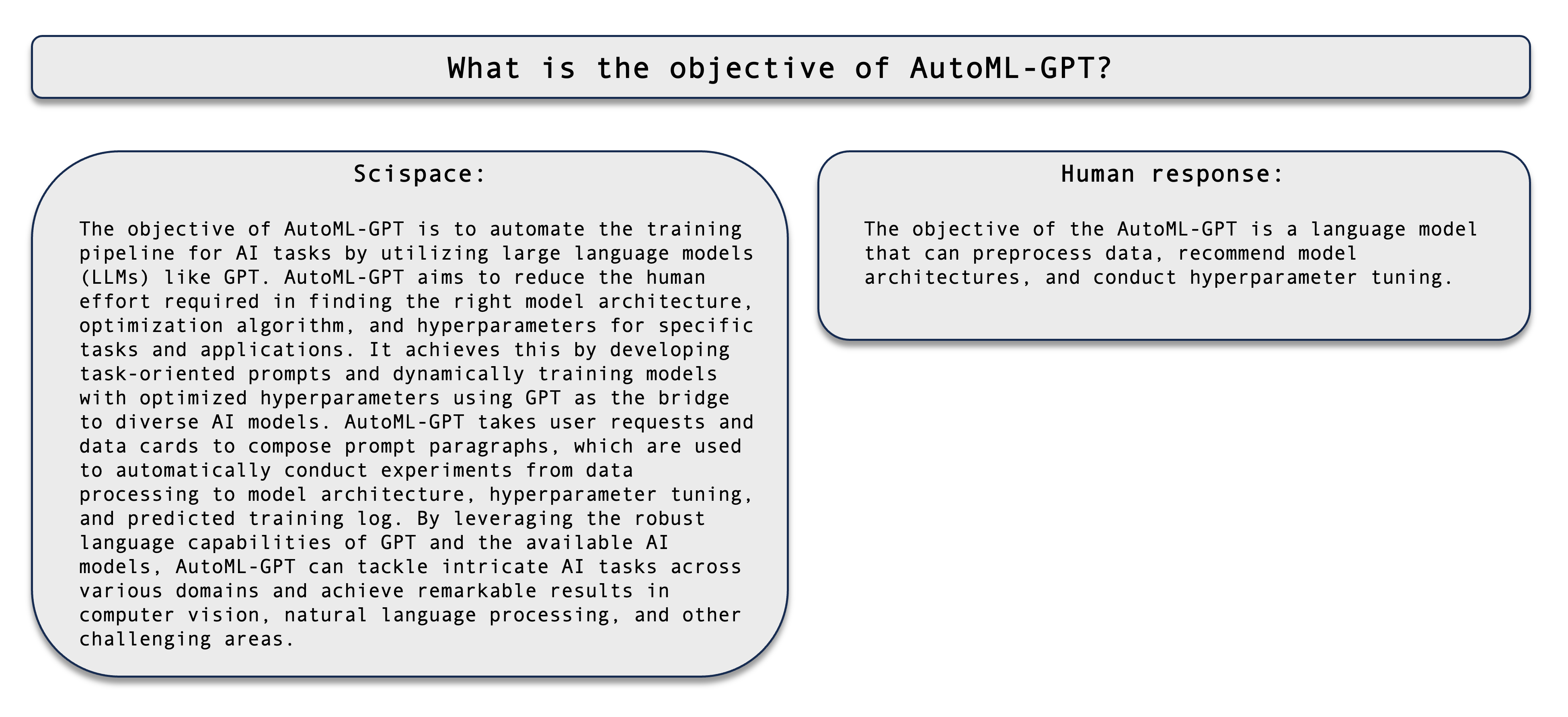}
\end{figure}

\begin{figure}[htbp]
    \centering
    \includegraphics[scale = 0.6]{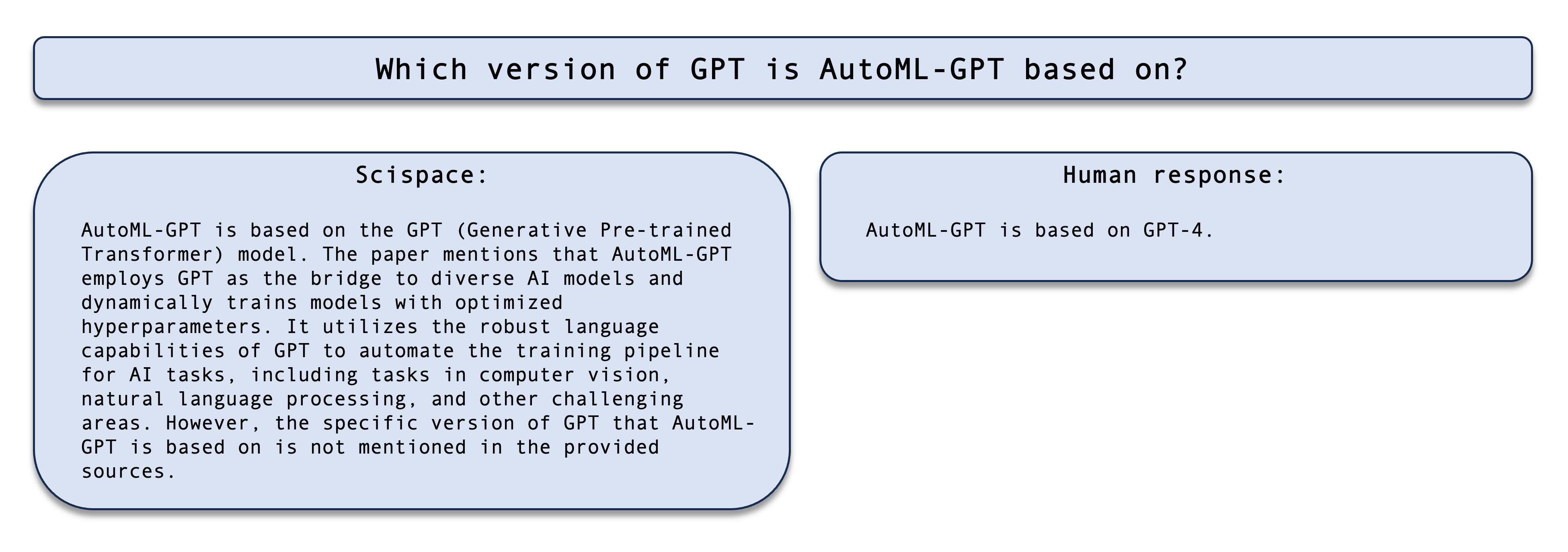}
\end{figure}

\begin{figure}[htbp]
    \centering
    \includegraphics[scale = 0.6]{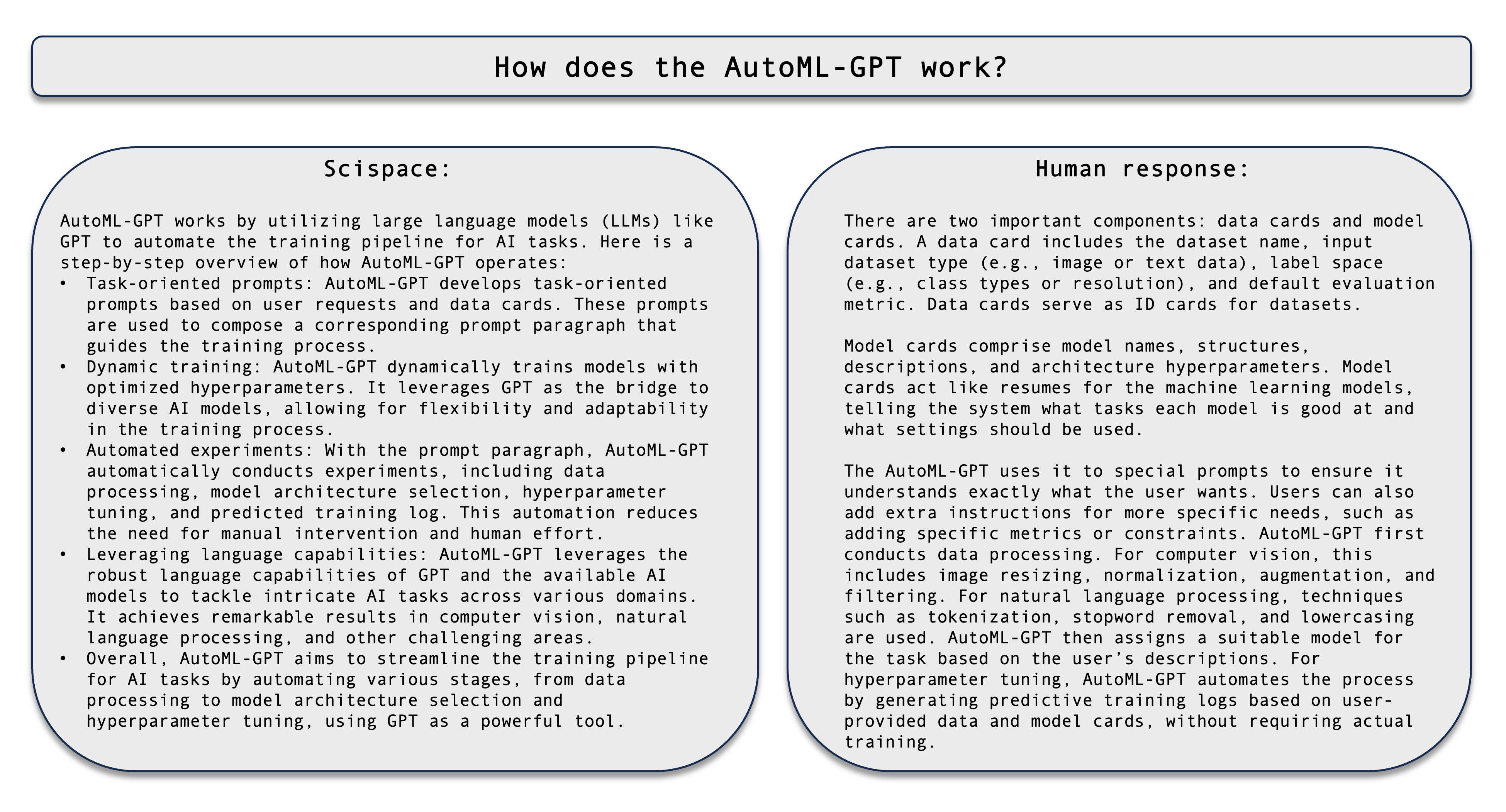}
\end{figure}

\begin{figure}[htbp]
    \centering
    \includegraphics[scale = 0.6]{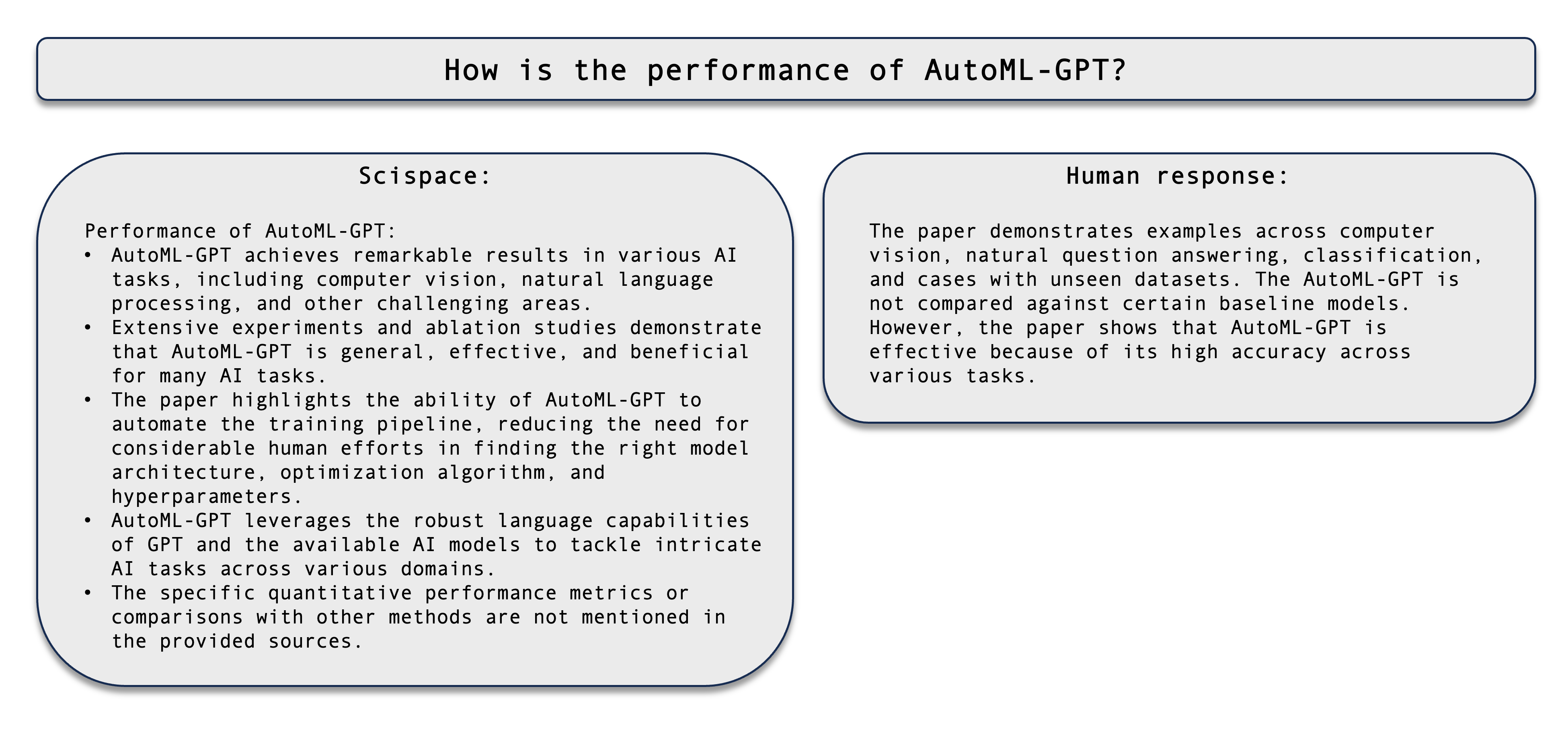}
\end{figure}

\begin{figure}[htbp]
    \centering
    \includegraphics[scale = 0.6]{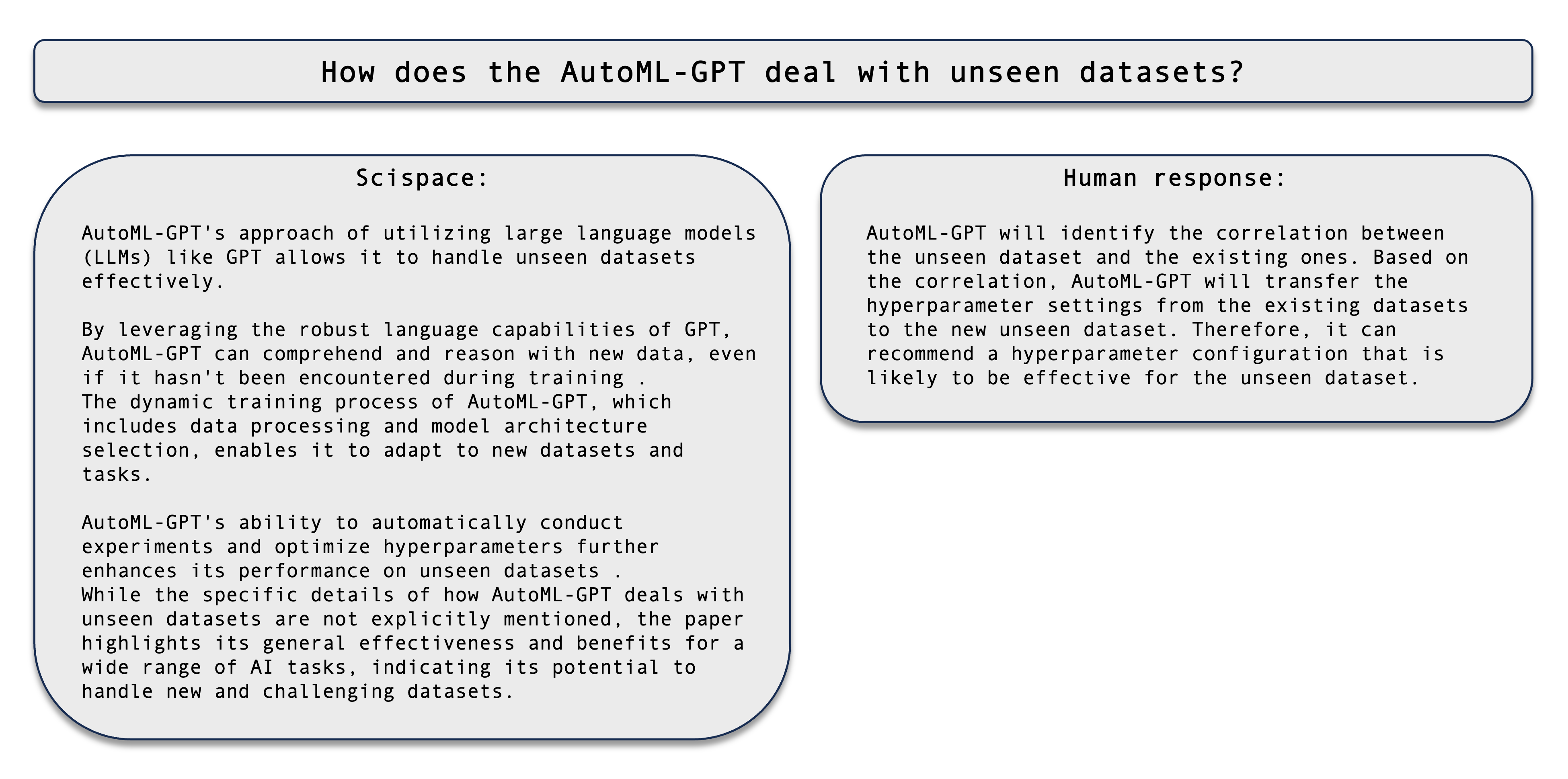}
\end{figure}

\clearpage

\textbf{Engineering:}\\

Generative pre-trained transformers (GPT) for surface engineering \cite{surfacegpt}

\begin{figure}[htbp]
    \centering
    \includegraphics[scale = 0.6]{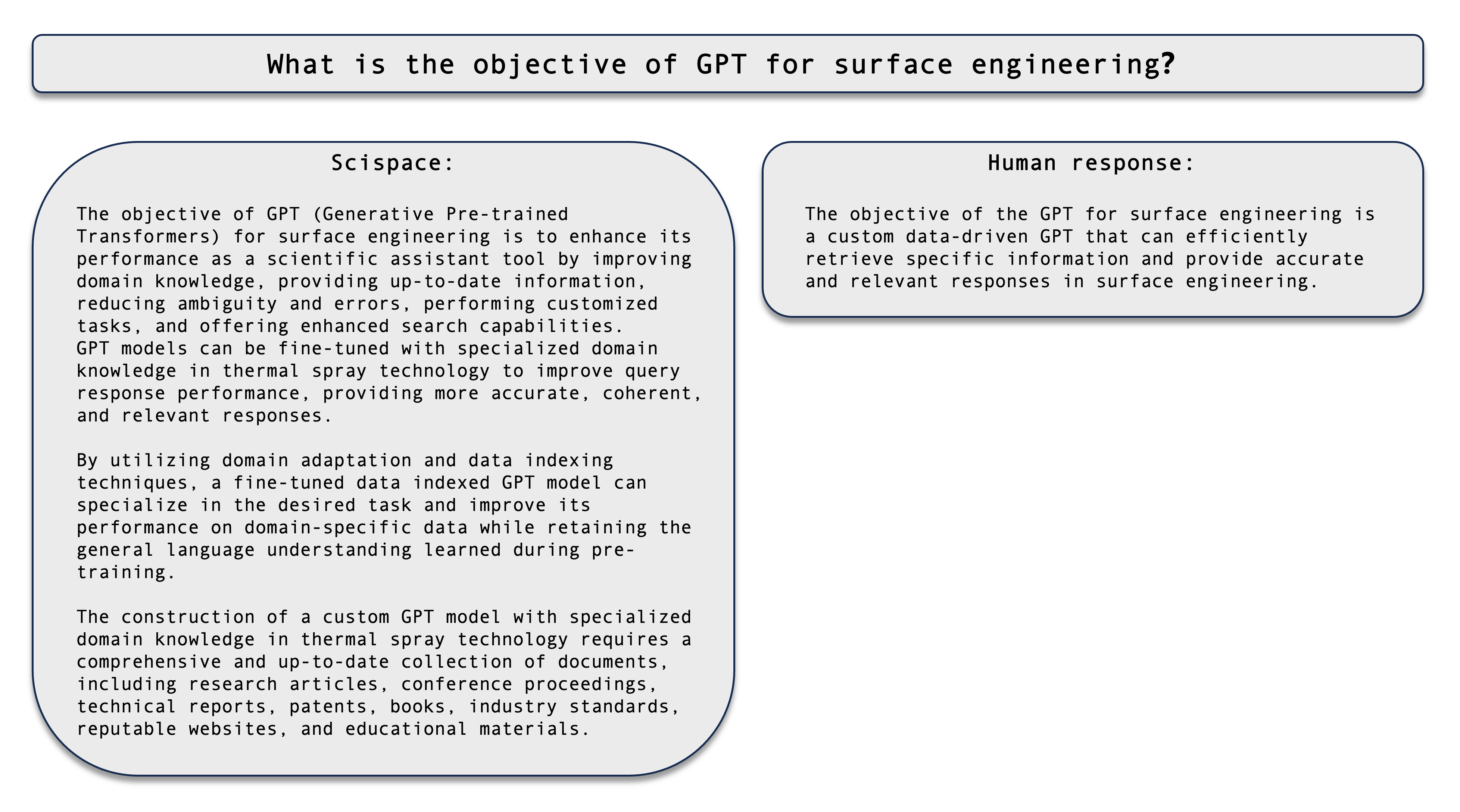}
\end{figure}

\begin{figure}[htbp]
    \centering
    \includegraphics[scale = 0.6]{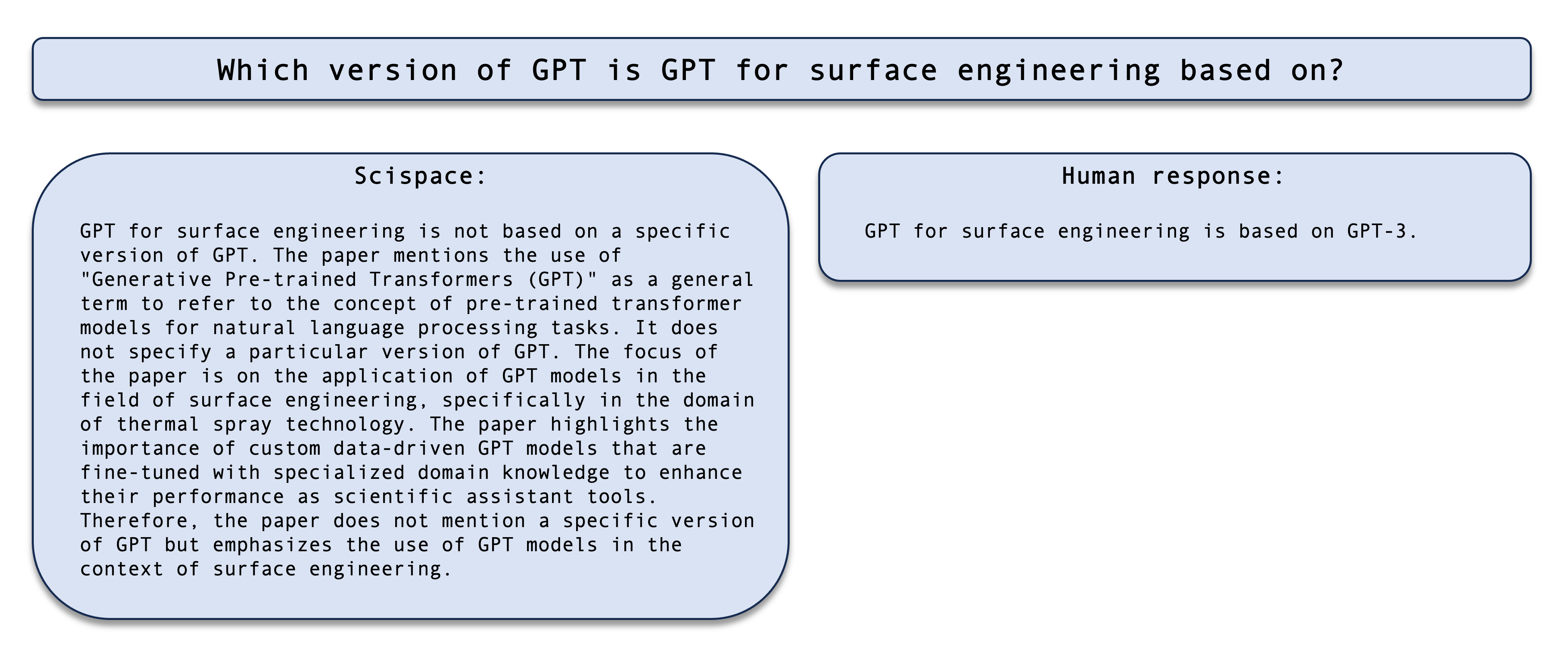}
\end{figure}

\begin{figure}[htbp]
    \centering
    \includegraphics[scale = 0.6]{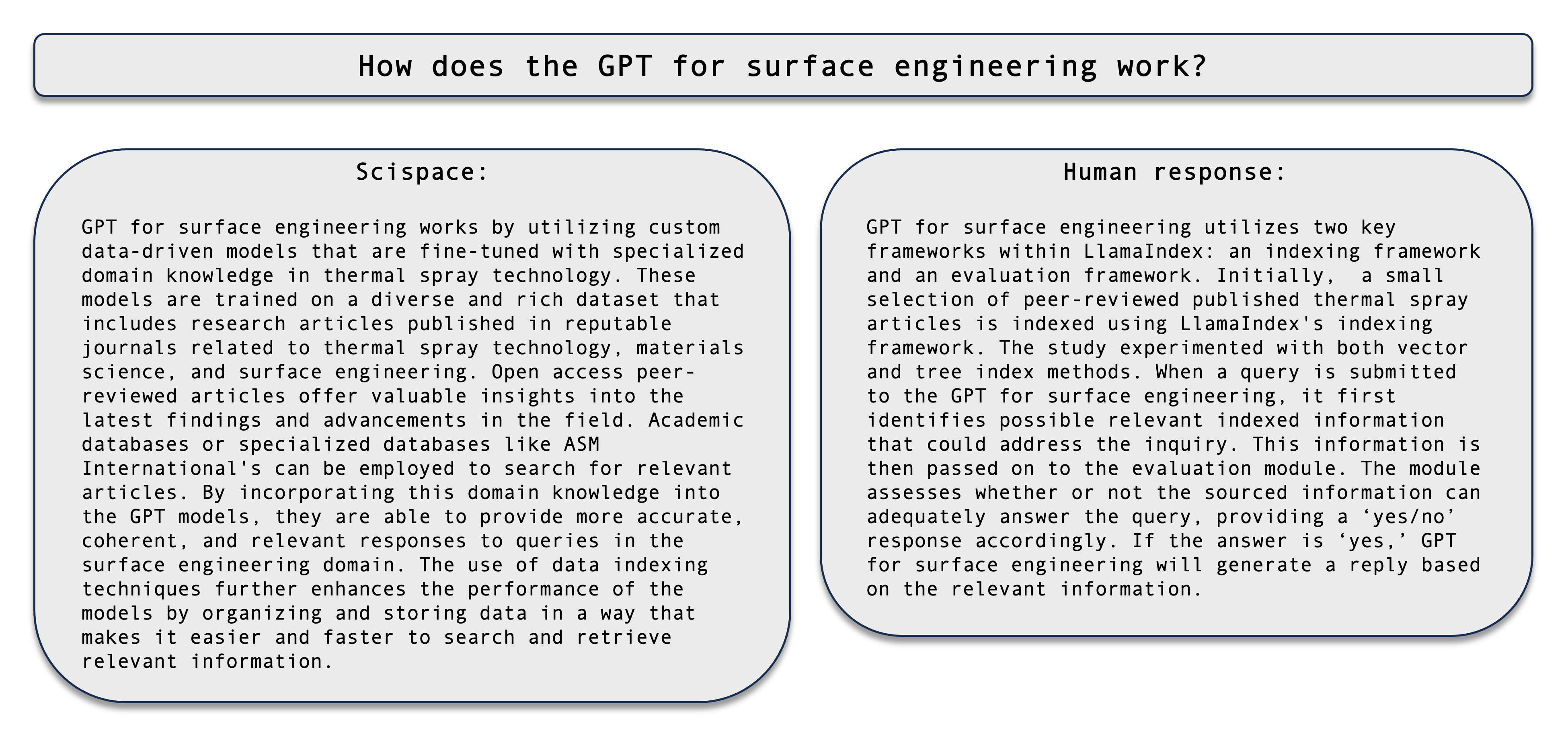}
\end{figure}

\begin{figure}[htbp]
    \centering
    \includegraphics[scale = 0.6]{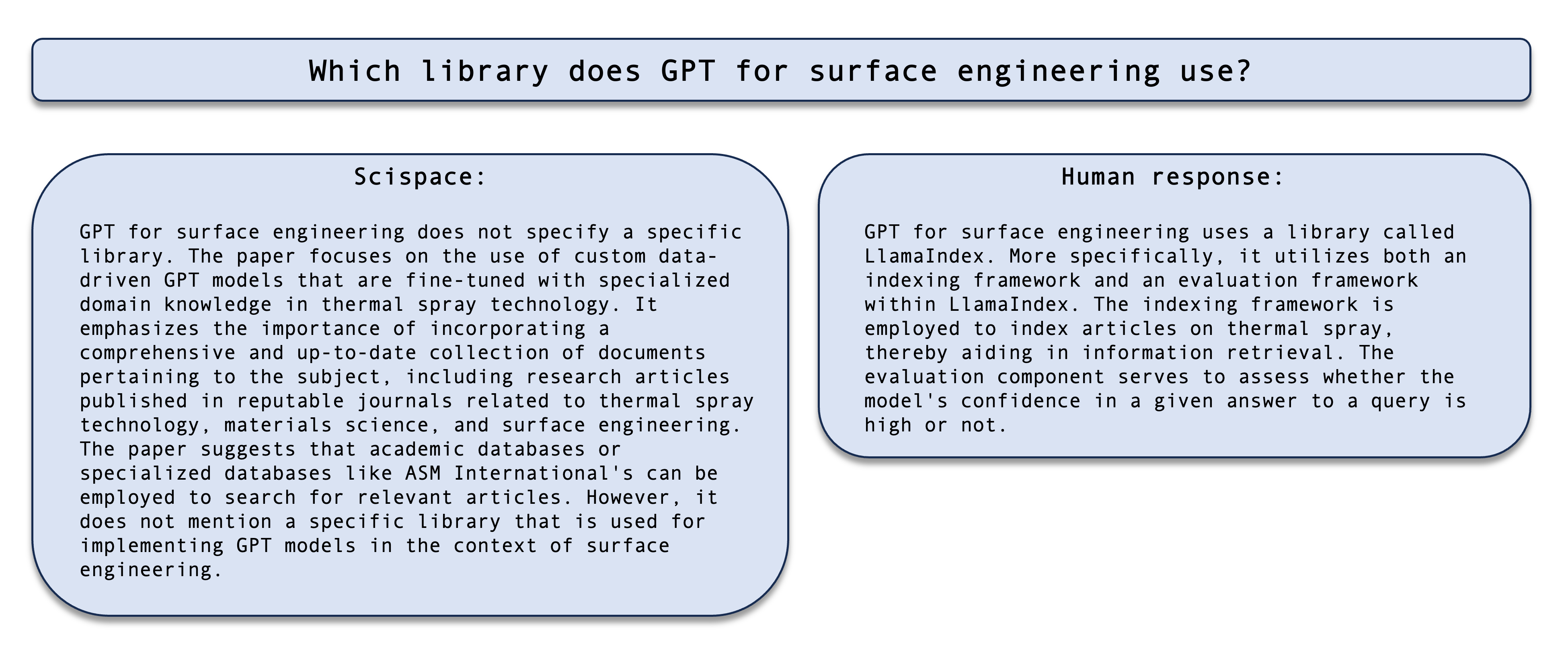}
\end{figure}

\begin{figure}[htbp]
    \centering
    \includegraphics[scale = 0.6]{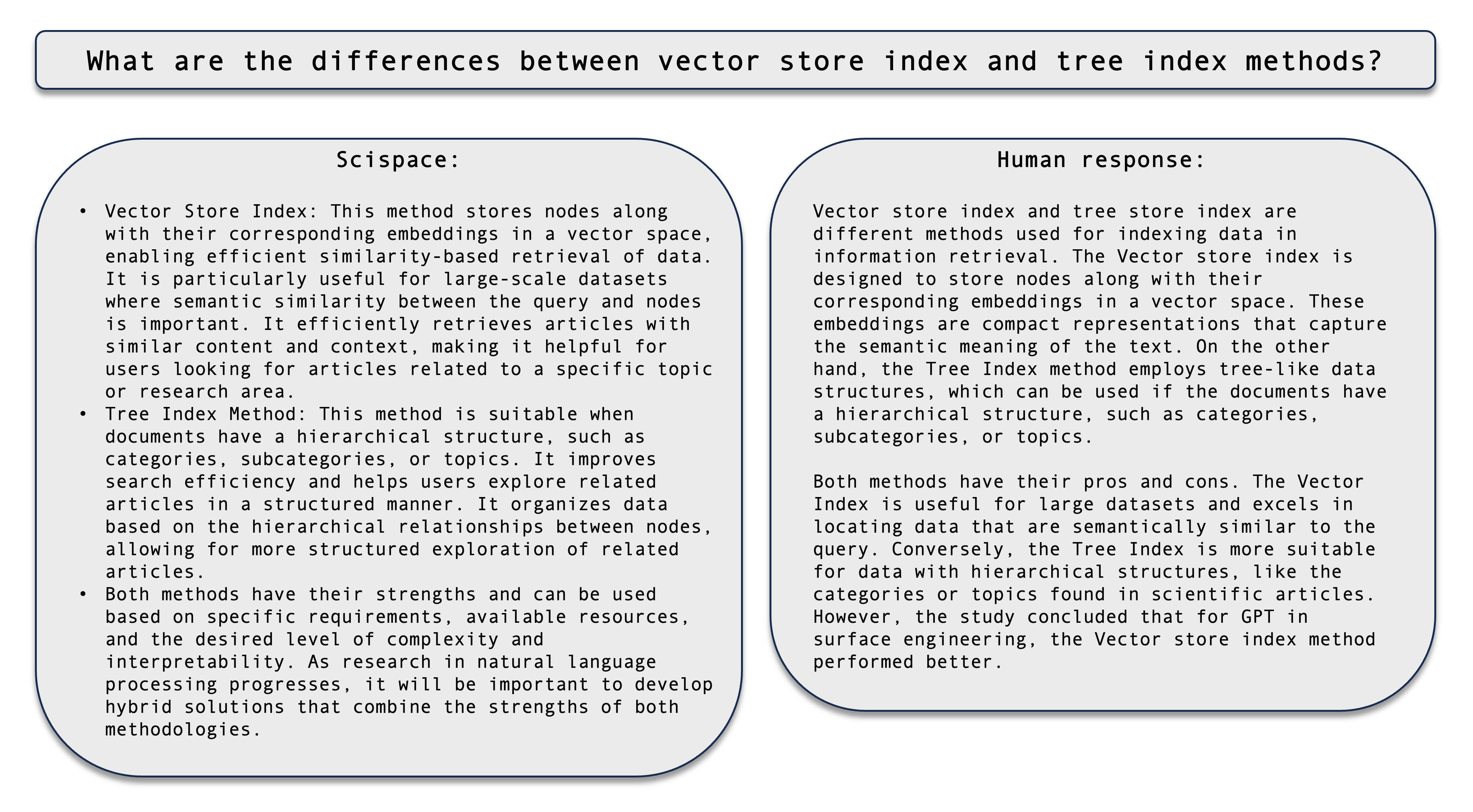}
\end{figure}

\clearpage
ChatGPT for Robotics: Design Principles and Model Abilities \cite{roboticsgpt}

\begin{figure}[htbp]
    \centering
    \includegraphics[scale = 0.6]{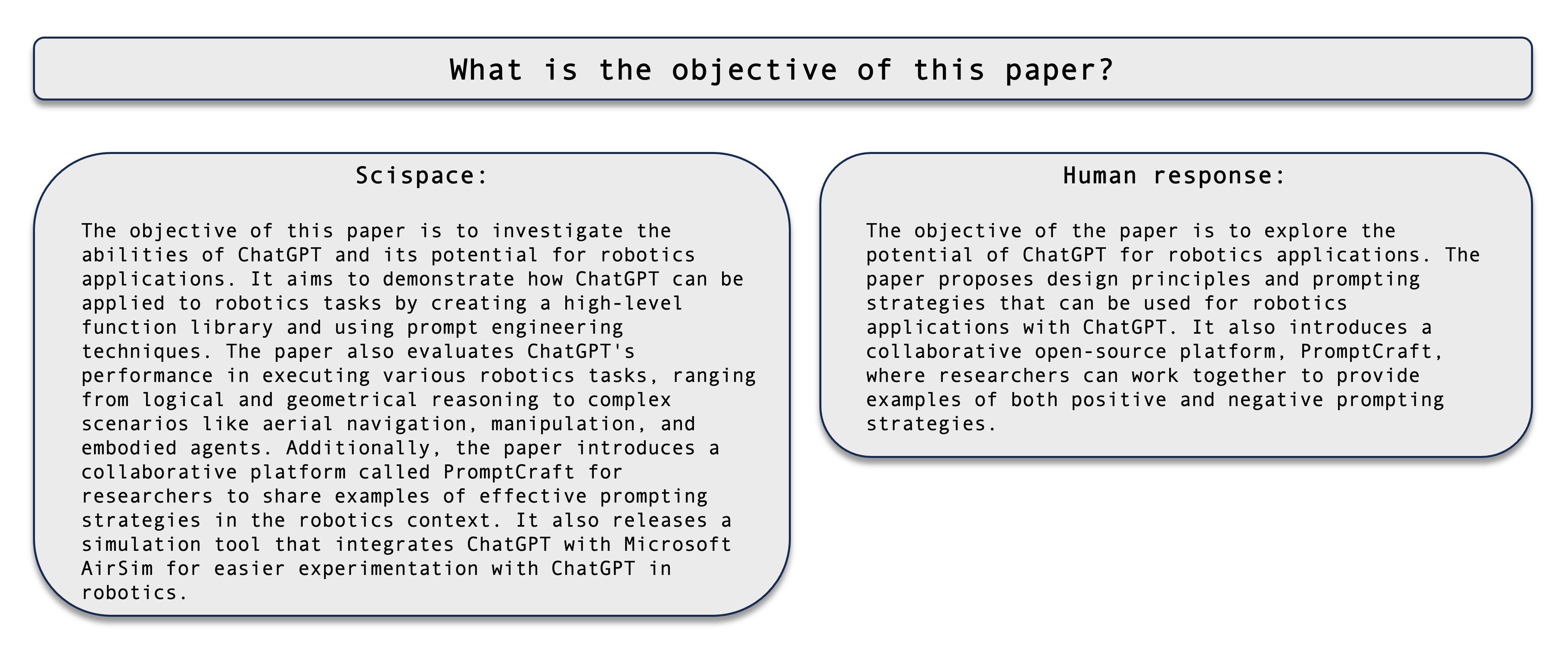}
\end{figure}

\begin{figure}[htbp]
    \centering
    \includegraphics[scale = 0.6]{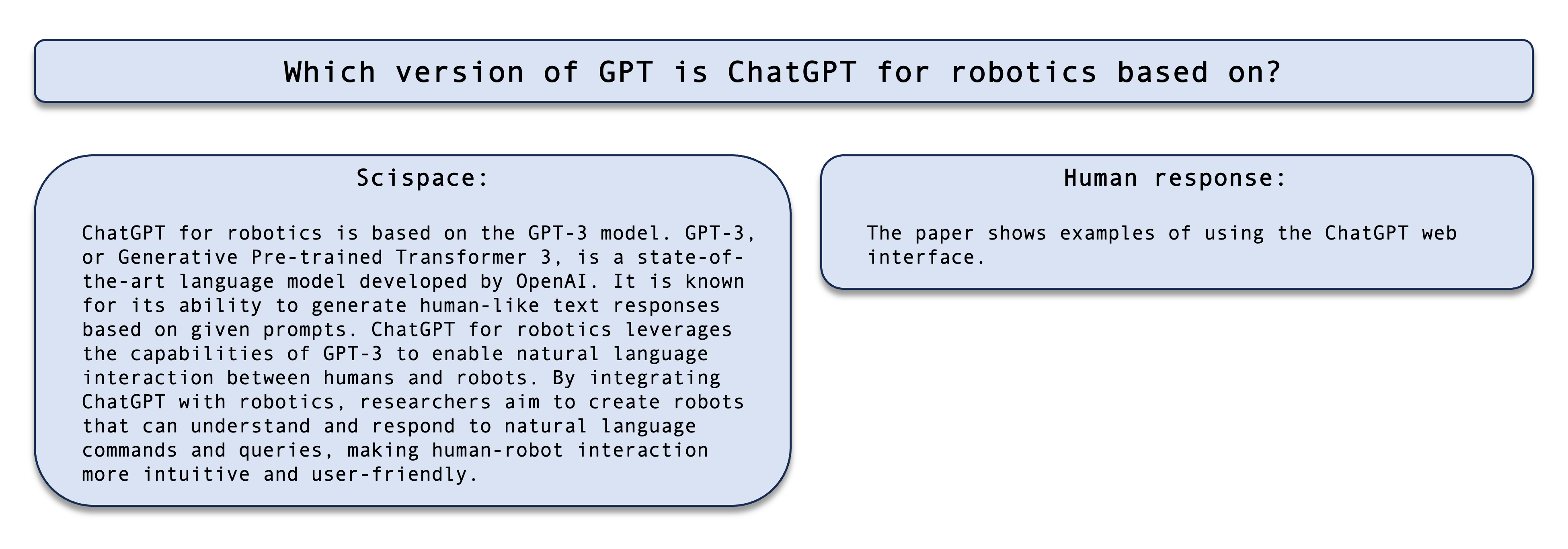}
\end{figure}

\begin{figure}[htbp]
    \centering
    \includegraphics[scale = 0.6]{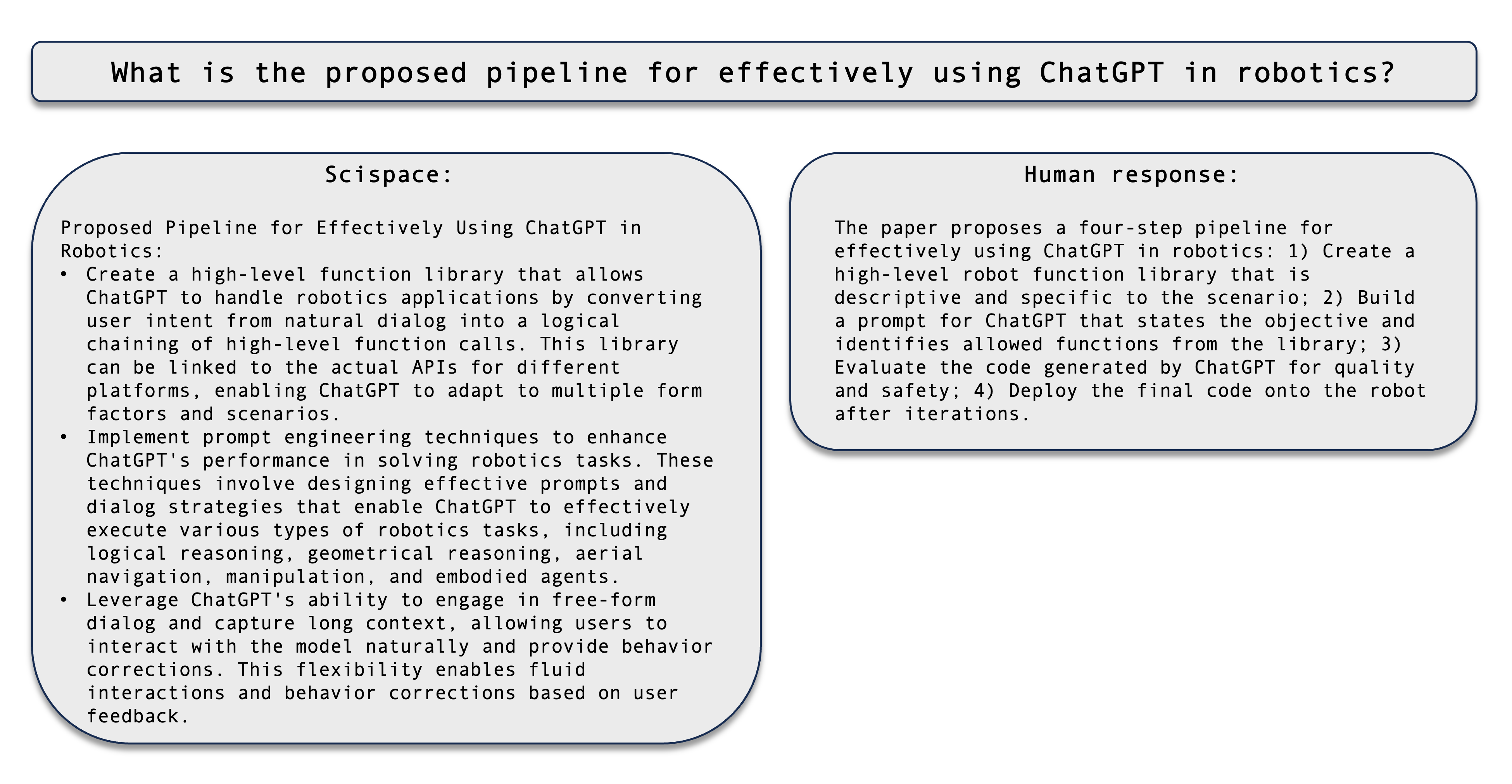}
\end{figure}

\begin{figure}[htbp]
    \centering
    \includegraphics[scale = 0.6]{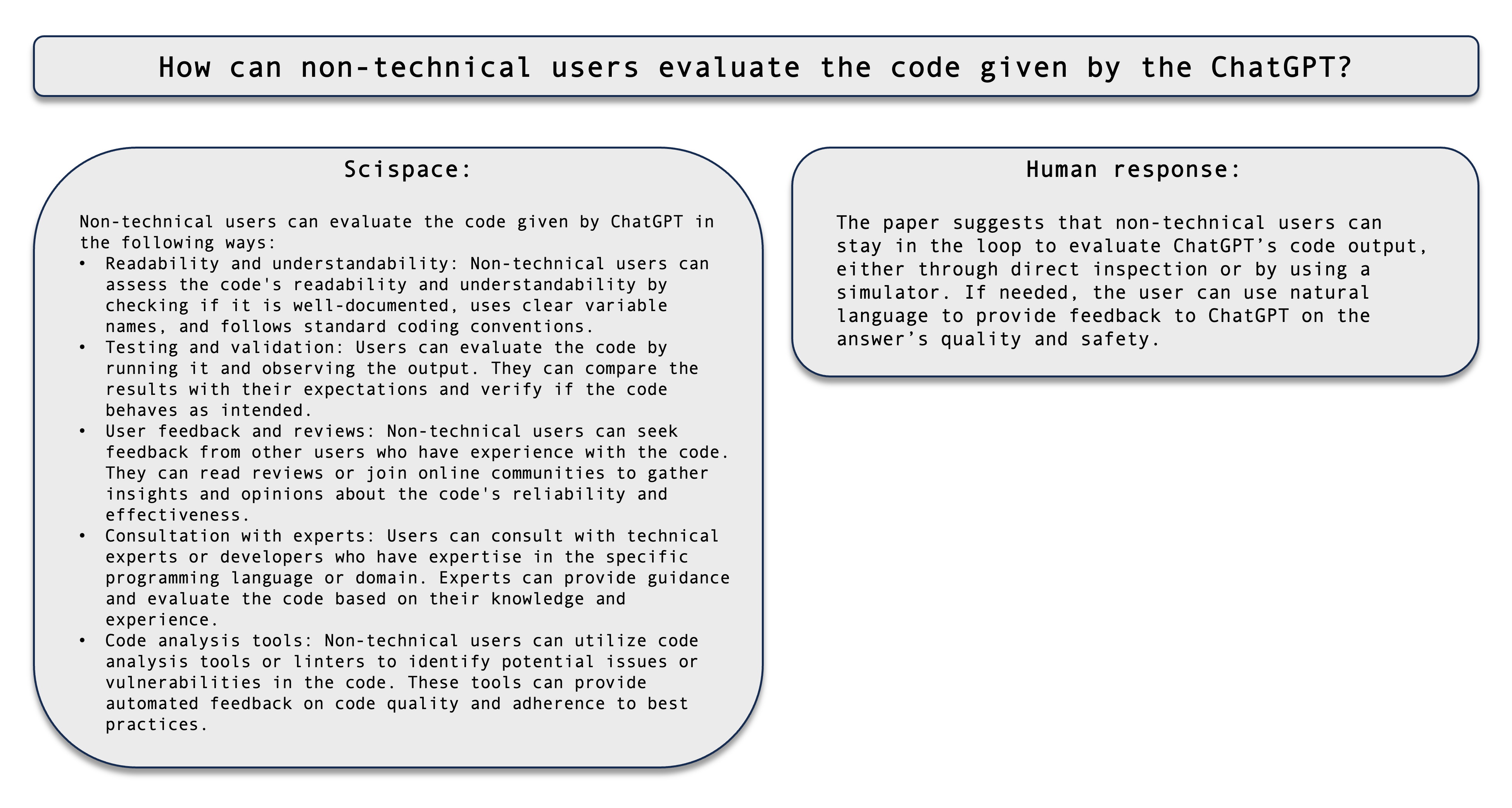}
\end{figure}

\begin{figure}[htbp]
    \centering
    \includegraphics[scale = 0.6]{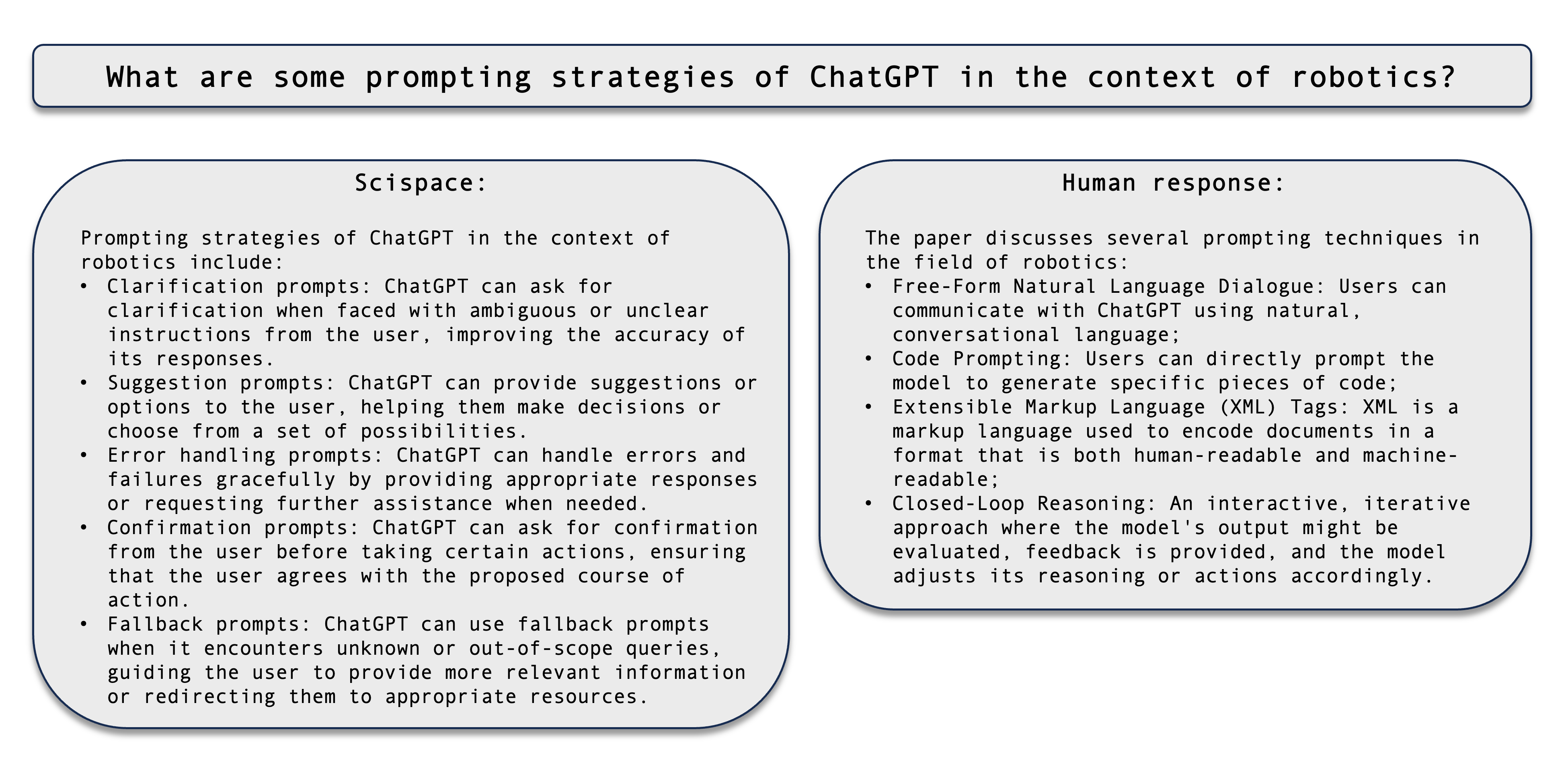}
\end{figure}

\clearpage

\textbf{Geography:}\\

GeoGPT: Understanding and Processing Geospatial Tasks through An Autonomous GPT \cite{geogpt}

\begin{figure}[htbp]
    \centering
    \includegraphics[scale = 0.6]{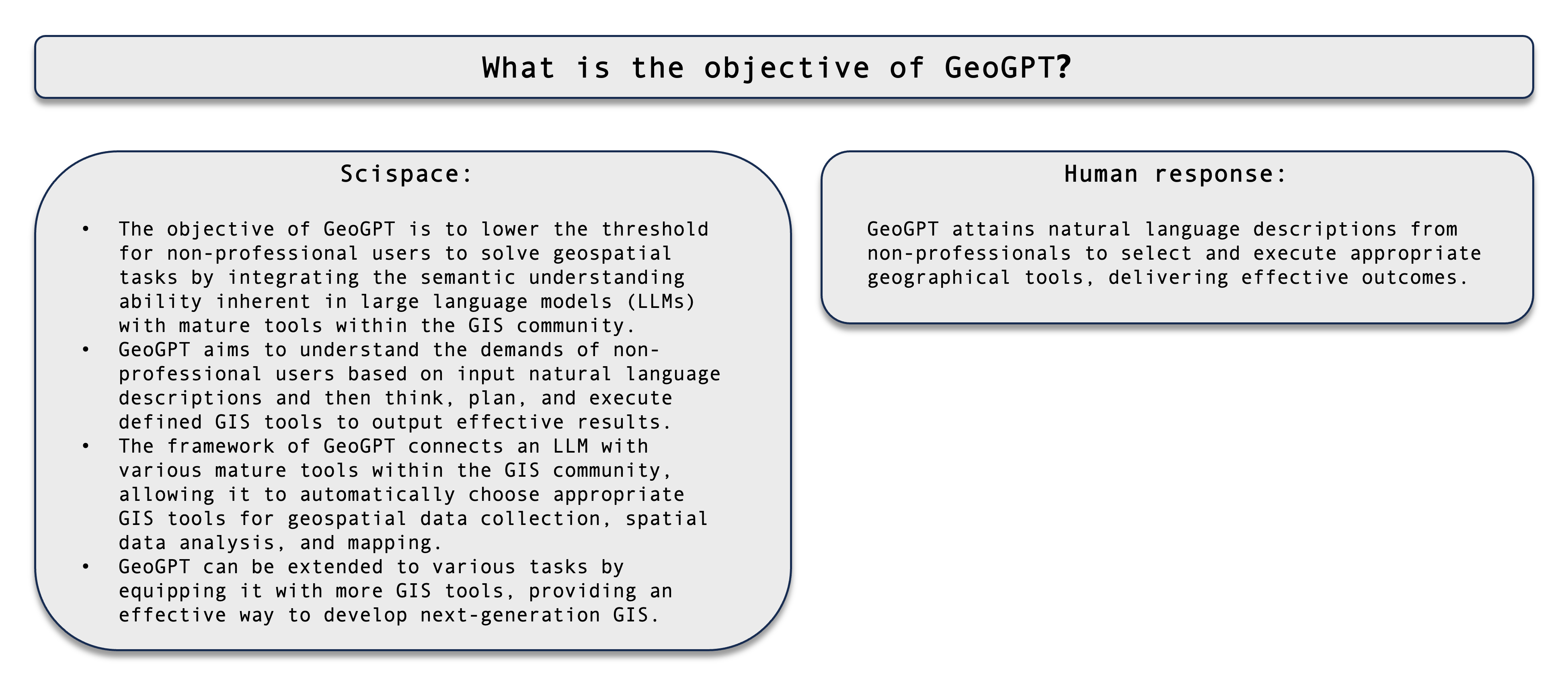}
\end{figure}

\begin{figure}[htbp]
    \centering
    \includegraphics[scale = 0.6]{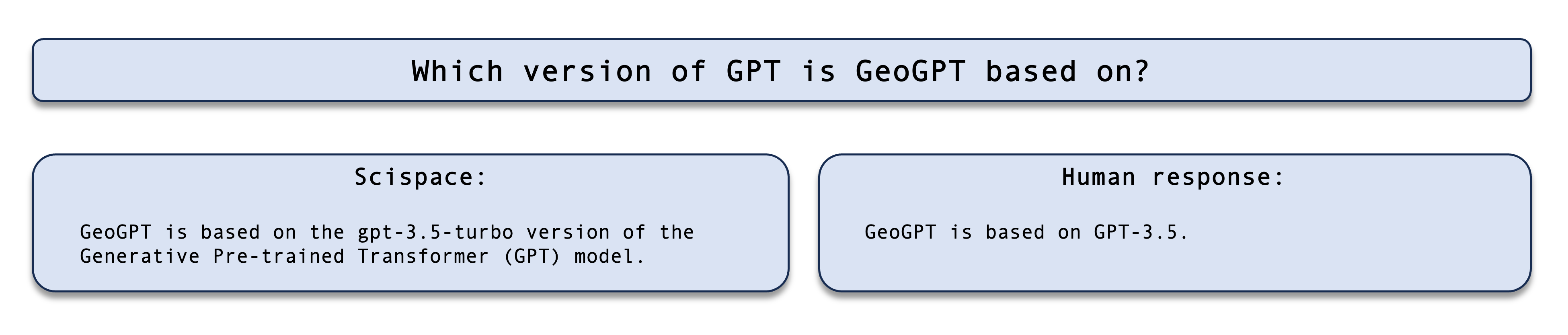}
\end{figure}

\begin{figure}[htbp]
    \centering
    \includegraphics[scale = 0.6]{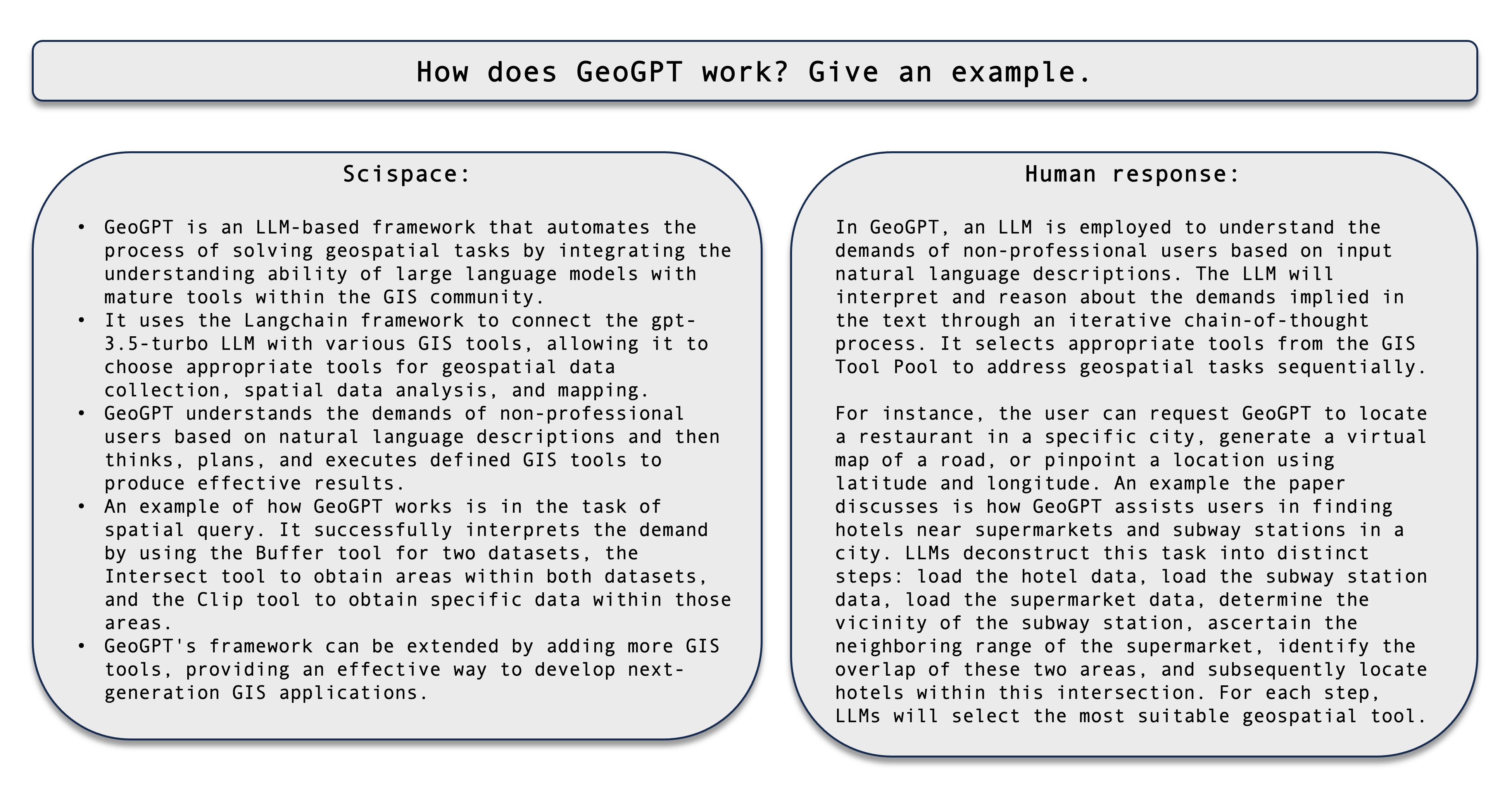}
\end{figure}

\begin{figure}[htbp]
    \centering
    \includegraphics[scale = 0.6]{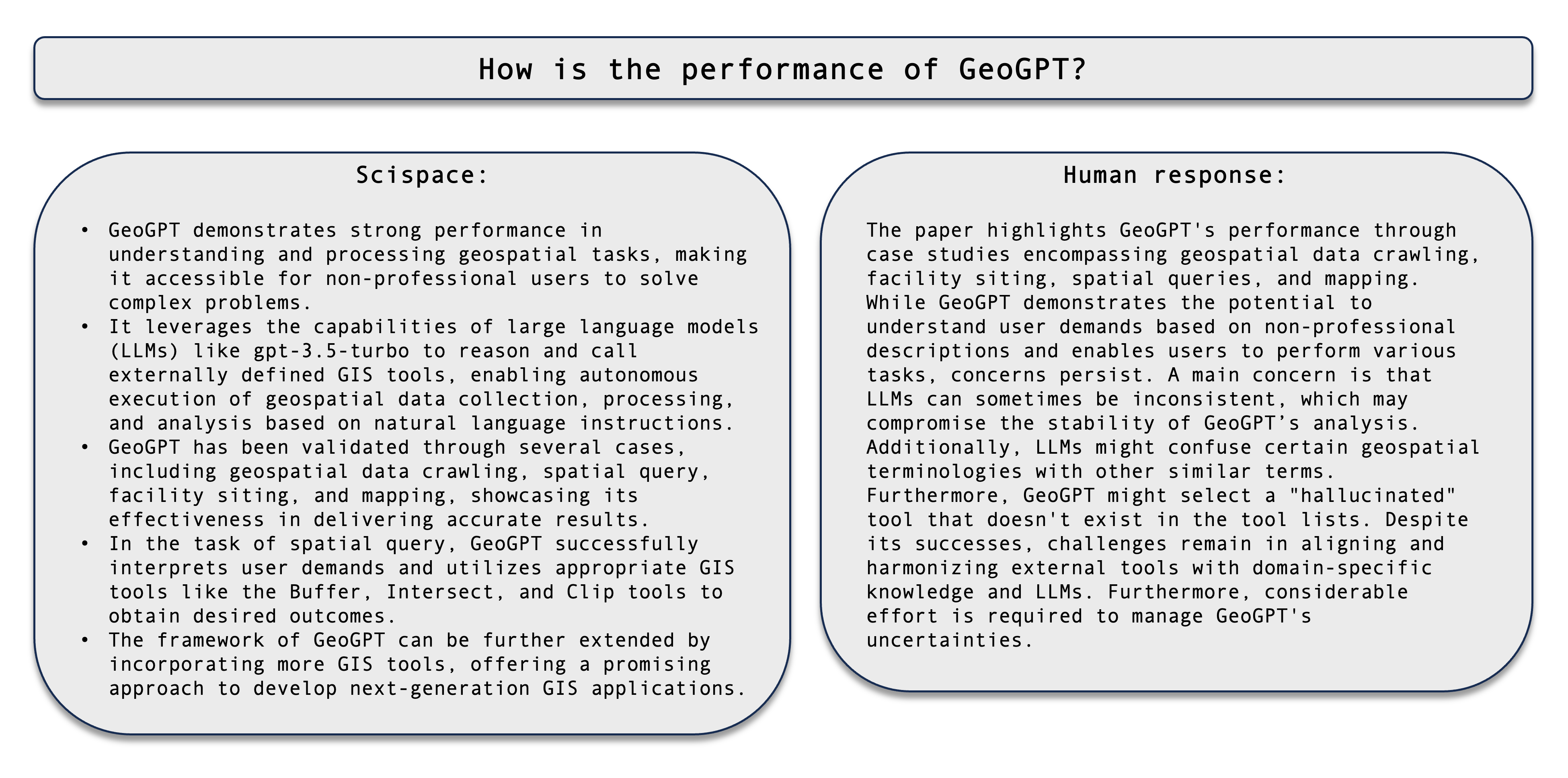}
\end{figure}

\begin{figure}[htbp]
    \centering
    \includegraphics[scale = 0.6]{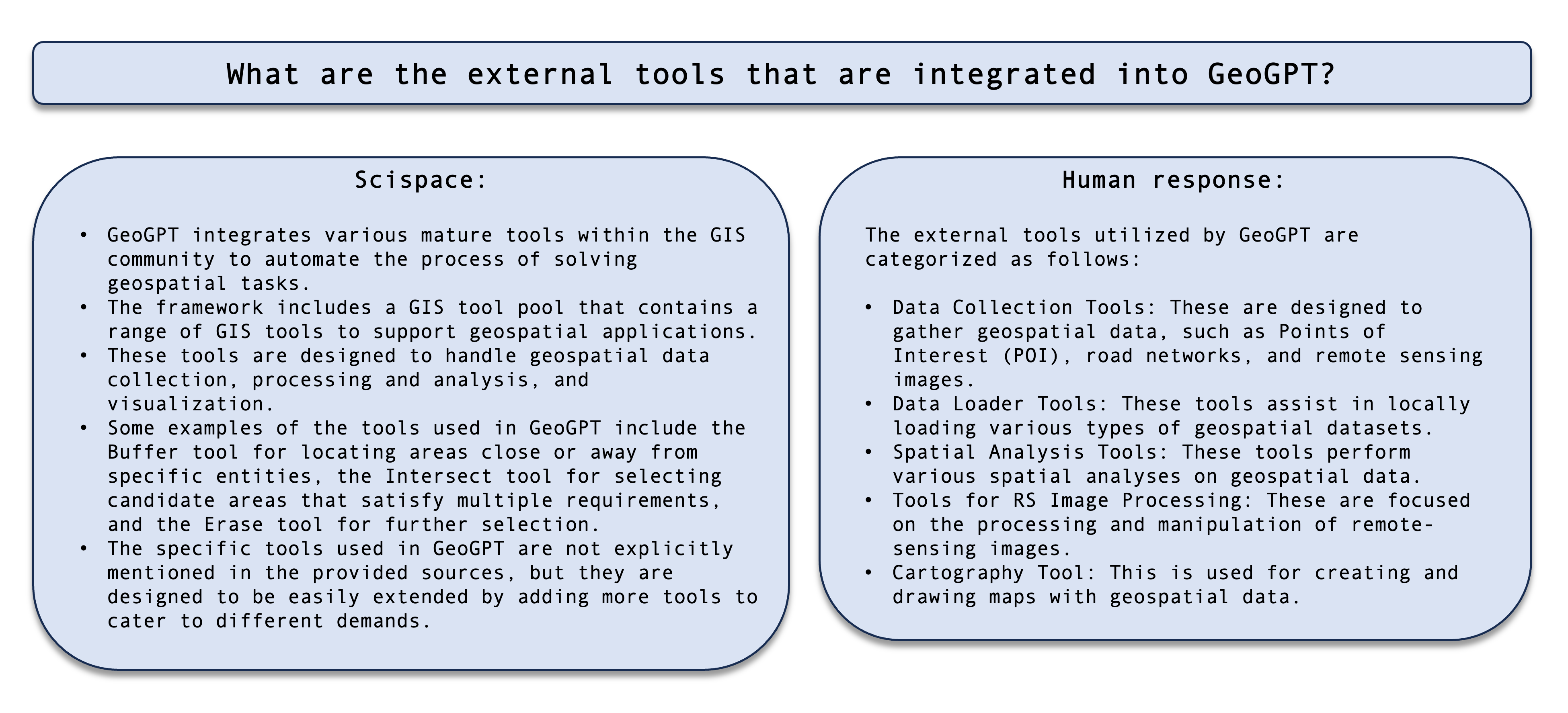}
\end{figure}

\clearpage
Geotechnical Parrot Tales (GPT): Harnessing Large Language Models in geotechnical engineering \cite{geotechgpt}

\begin{figure}[htbp]
    \centering
    \includegraphics[scale = 0.6]{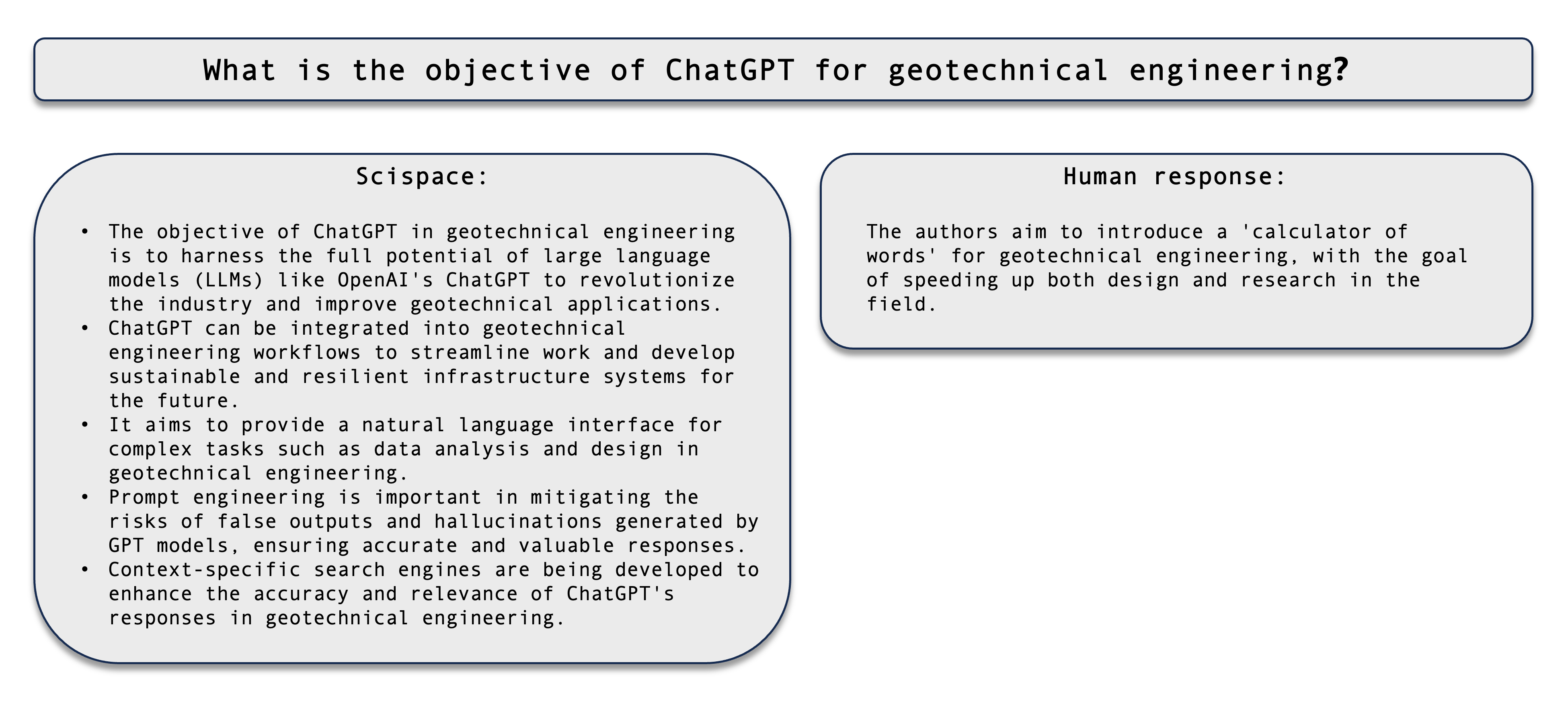}
\end{figure}

\begin{figure}[htbp]
    \centering
    \includegraphics[scale = 0.6]{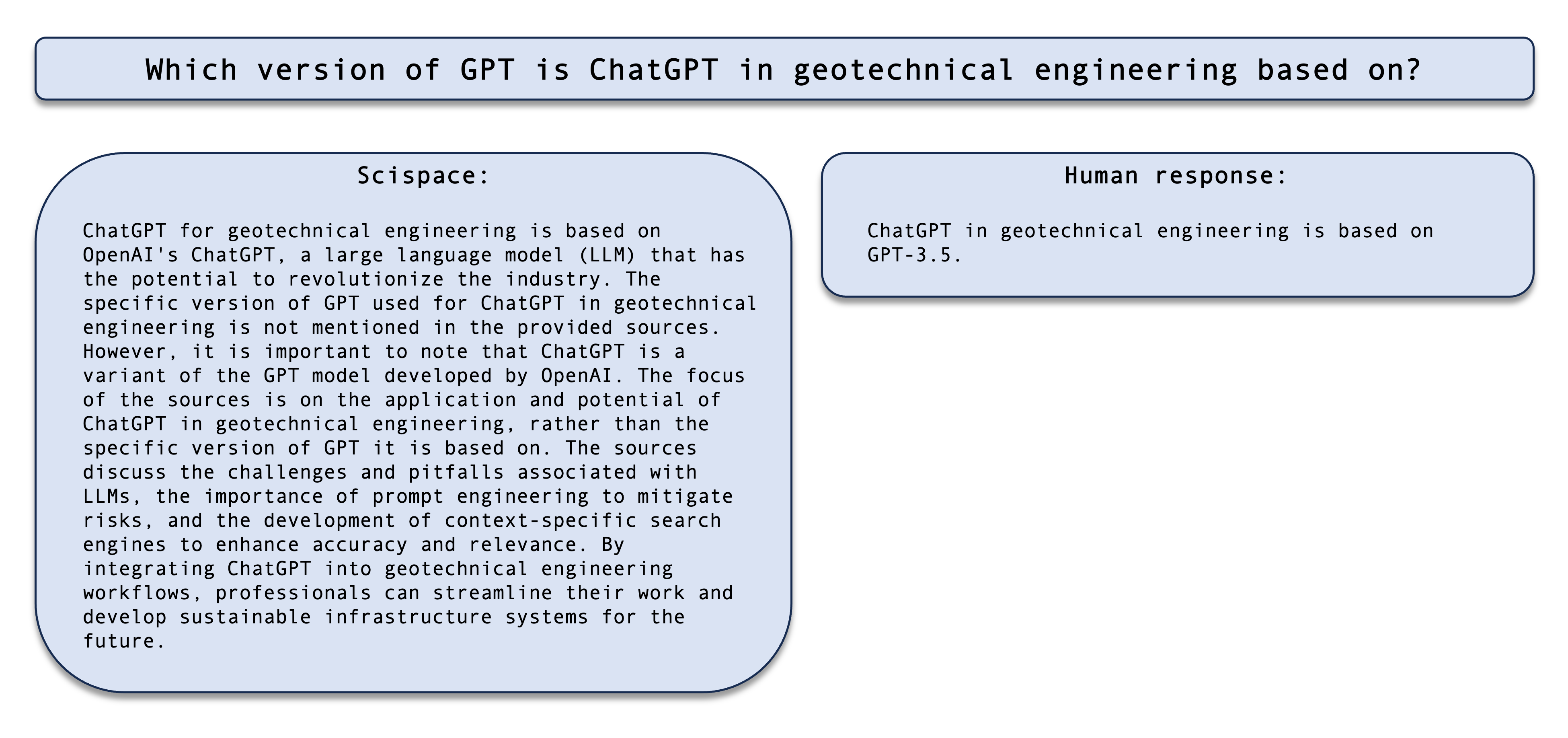}
    
\end{figure}\begin{figure}[htbp]
    \centering
    \includegraphics[scale = 0.6]{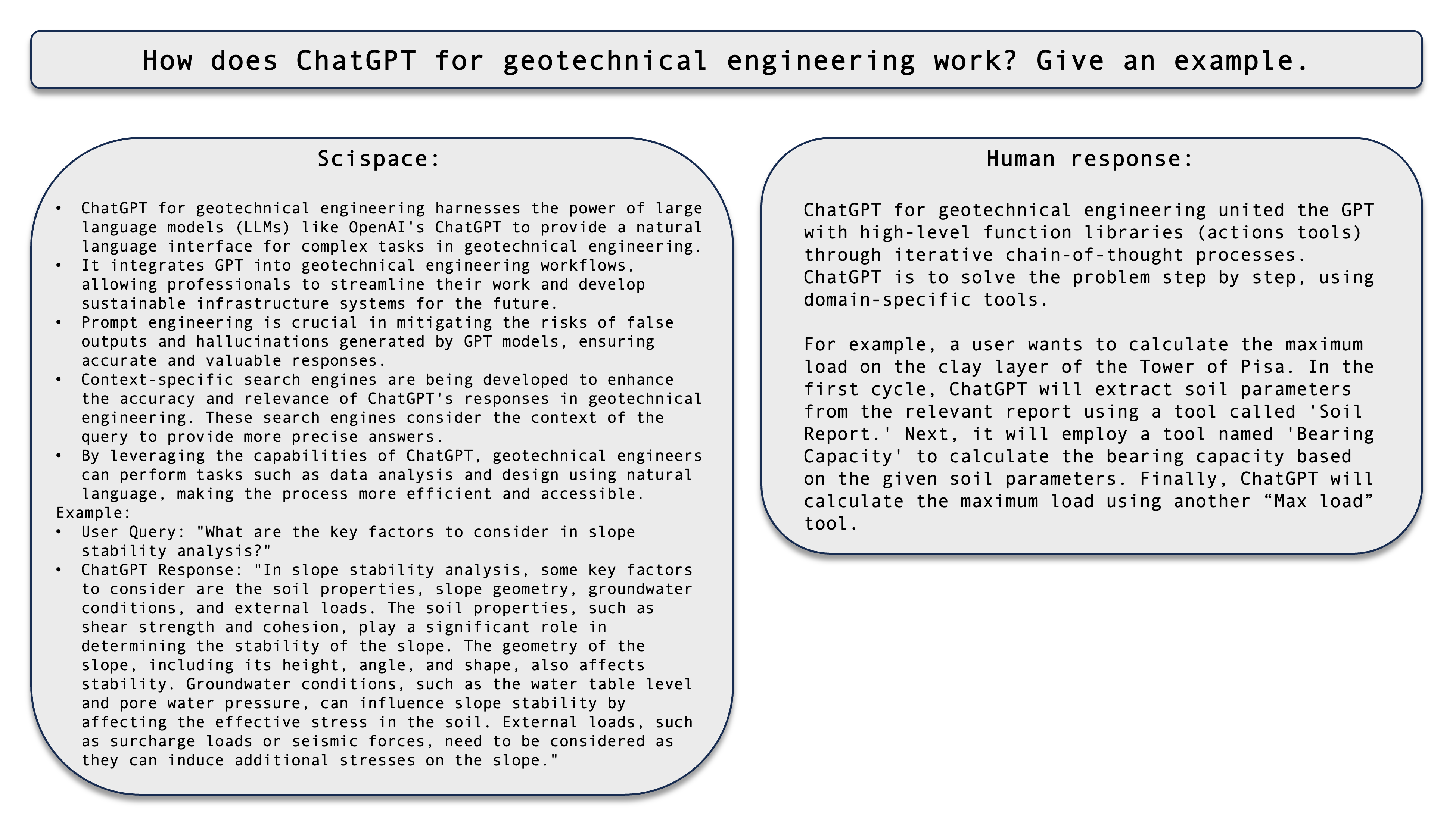}
\end{figure}

\begin{figure}[htbp]
    \centering
    \includegraphics[scale = 0.6]{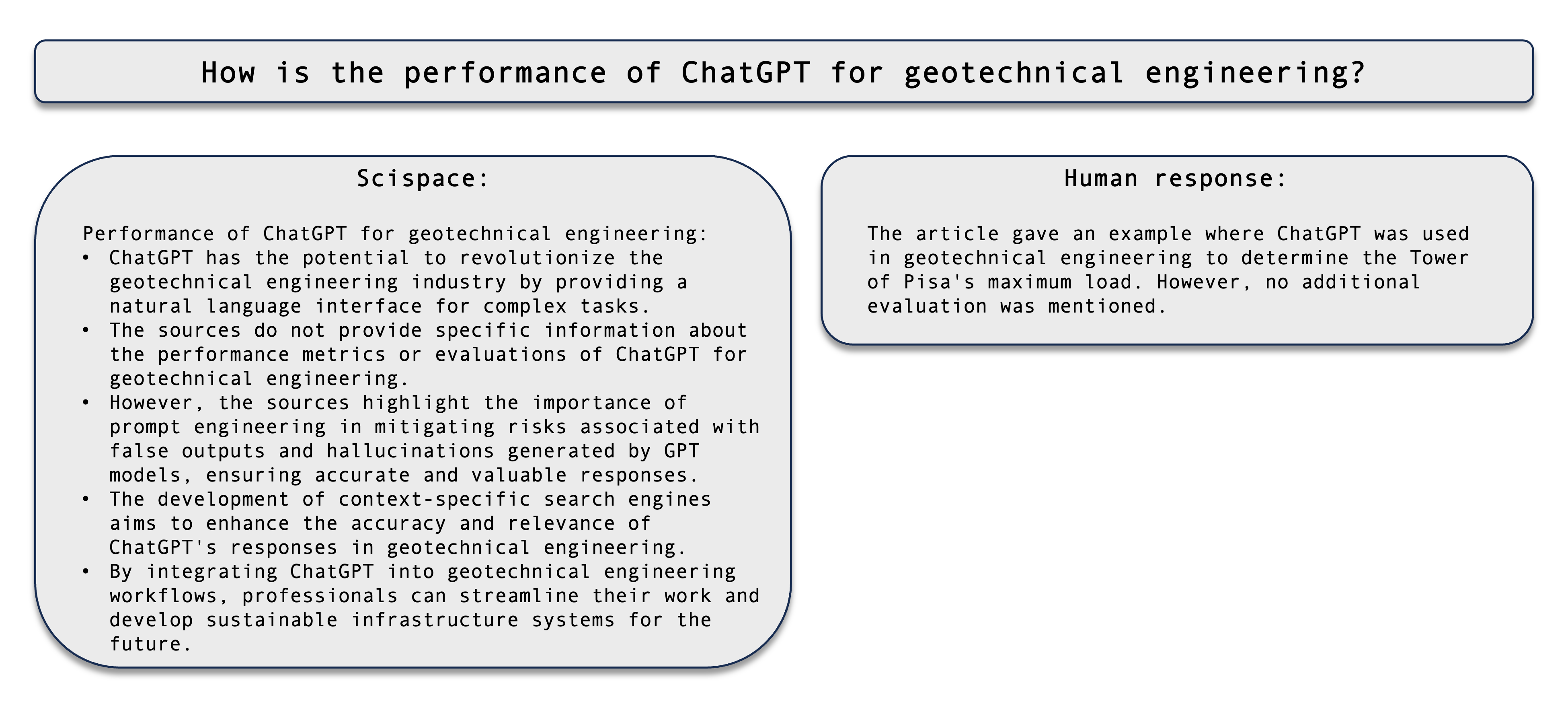}
\end{figure}

\begin{figure}[htbp]
    \centering
    \includegraphics[scale = 0.6]{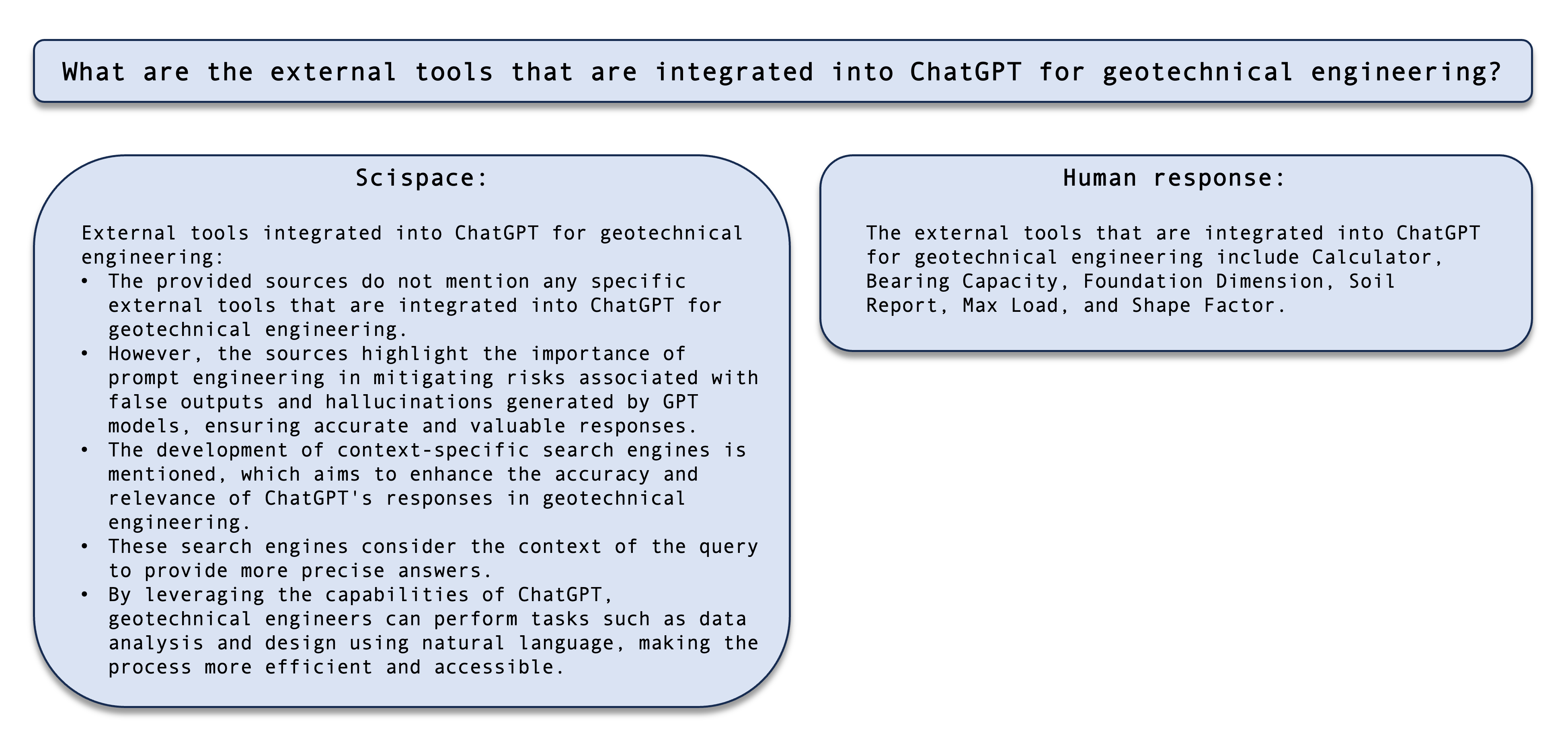}
\end{figure}

\end{document}